\documentclass{article}

     \PassOptionsToPackage{numbers, compress}{natbib}
     \usepackage[final]{neurips_2024}

\usepackage[utf8]{inputenc} % allow utf-8 input
\usepackage[T1]{fontenc}    % use 8-bit T1 fonts
\usepackage{hyperref}       % hyperlinks
\usepackage{url}            % simple URL typesetting
\usepackage{booktabs}       % professional-quality tables
\usepackage{amsfonts}       % blackboard math symbols
\usepackage{physics}
\usepackage{amsmath}
\usepackage{nicefrac}       % compact symbols for 1/2, etc.
\usepackage{microtype}      % microtypography
\usepackage[dvipsnames]{xcolor} % colors
\usepackage{graphicx}
\usepackage{wrapfig}
\usepackage{caption}
\usepackage{subcaption}
\usepackage{longtable}
\usepackage{multirow}

\usepackage[capitalize]{cleveref}
\crefname{section}{Sec.}{Secs.}
\Crefname{section}{Section}{Sections}
\Crefname{table}{Table}{Tables}
\crefname{table}{Tab.}{Tabs.}

\newcommand{\llava}{LLaVA}
\newcommand{\llama}{LLaMA}
\newcommand{\nerf}{NeRF}
\newcommand{\nftovec}{\texttt{nf2vec}}
\newcommand{\inrtovec}{\texttt{inr2vec}}

\newcommand{\model}{LLaNA}

\title{LLaNA: Large Language and NeRF Assistant}
\author{
  Andrea Amaduzzi \\
  \texttt{andrea.amaduzzi4@unibo.it}
  \And
  Pierluigi Zama Ramirez \\
  \texttt{pierluigi.zama@unibo.it}
  \AND
  Giuseppe Lisanti\\
   \texttt{giuseppe.lisanti@unibo.it}
  \And
  Samuele Salti \\
  \texttt{samuele.salti@unibo.it}\\
  \And
  Luigi Di Stefano\\
  \texttt{luigi.distefano@unibo.it} \\
  \\ 
  CVLAB, University of Bologna \\\\
  \url{https://andreamaduzzi.github.io/llana/}
}

\begin{document}

\maketitle

\begin{abstract}
 Multimodal Large Language Models (MLLMs) have demonstrated an excellent understanding of images and 3D data. However, both modalities have shortcomings in holistically capturing the appearance and geometry of objects. Meanwhile, Neural Radiance Fields (NeRFs), which encode information within the weights of a simple Multi-Layer Perceptron (MLP), have emerged as an increasingly widespread modality that simultaneously encodes the geometry and photorealistic appearance of objects. This paper investigates the feasibility and effectiveness of ingesting NeRF into MLLM. We create LLaNA, the first general-purpose NeRF-language assistant capable of performing new tasks such as NeRF captioning and Q\&A. Notably, our method directly processes the weights of the NeRF's MLP to extract information about the represented objects without the need to render images or materialize 3D data structures. Moreover, we build a dataset of NeRFs with text annotations for various NeRF-language tasks with no human intervention. Based on this dataset, we develop a benchmark to evaluate the NeRF understanding capability of our method. Results show that processing NeRF weights performs favourably against extracting 2D or 3D representations from NeRFs.
\end{abstract}
\section{Introduction}
\label{sec:intro}
Large Language Models (LLMs)~\cite{GPT4, llama} have revolutionized the field of Natural Language Processing, demonstrating incredible text comprehension and generation capabilities. These results have fostered the development of Multimodal LLMs (MLLMs)~\cite{Palm-E, LLaMA-Adapter, llava, InstructBLIP, VideoLLM}, which can ingest various modalities such as images, videos, and audio, to generate text describing and reasoning about the content of such modalities. 
Recently, MLLMs have also been extended to 3D data~\cite{gpt4point, pointllm}, primarily represented through colored point clouds, yielding remarkable results even in this scenario.

%Although images and colored point clouds are rich representations of an object, they both miss some aspects of it. Indeed, while images capture the interaction of the object's surface with light and by definition provide photorealistic renderings of it, they do not make explicitly available the 3D structure of an object, which can only be recovered using further processing of multiple views. Vice versa,  point clouds do provide the 3D structure of an object but, even if they store an RGB value for each point, they do not store its photorealistic appearance. 
Beyond images and 3D data, another paradigm is emerging to represent objects and scenes: Neural Radiance Fields (\nerf{}s)~\cite{nerf}. \nerf{}s are coordinate-based neural networks, typically Multi-Layer Perceptrons (MLPs), designed to capture both the geometry and the photorealistic appearance of an object by learning a continuous radiance field at each 3D spatial location. After training, a \nerf{} model can be queried to render realistic images or to reconstruct the 3D surface of the encoded object. 
Therefore, capturing an object as a \nerf{} provides an interesting alternative to create a digital twin with respect to standard representations such as multi-view images or point clouds. For instance, thanks to its continuous formulation, from a single \nerf{}, one can generate an infinite number of photorealistic images at any resolution while storing only the weights of an MLP instead of the entire image set. See \cref{sec:memory_nerf_vs_explicit} for more on the memory advantages of using NeRFs. Due to their advantages, \nerf{}s are effectively becoming a new modality stored and communicated independently, with datasets of \nerf{}s being made publicly available~\cite{hu2023nerf,ramirez2023deep} and companies providing digital twins of objects represented as \nerf{}s (e.g., {\footnotesize \url{https://lumalabs.ai/}}).

\begin{figure}[t]
    \centering
    \includegraphics[width=0.90\linewidth]{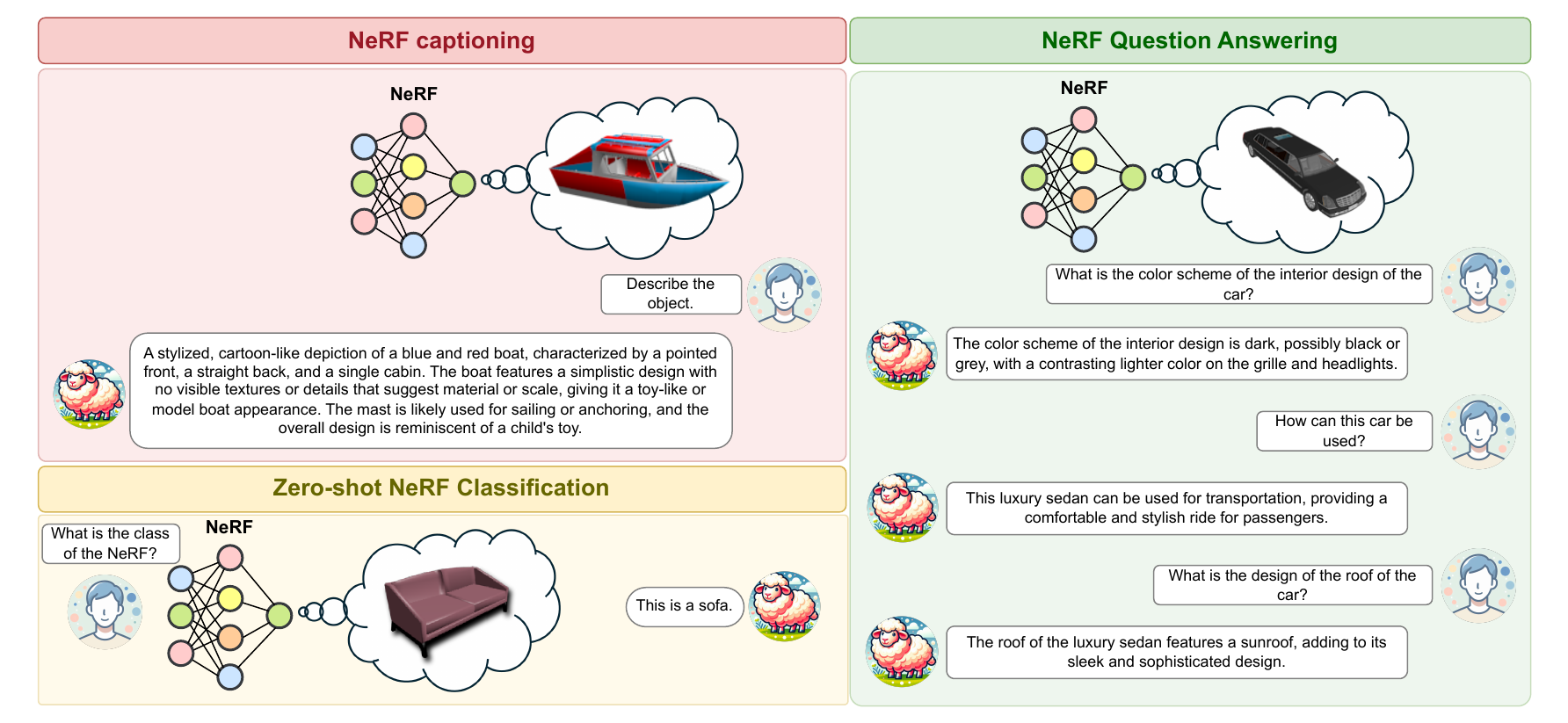}
    \setlength{\belowcaptionskip}{-15pt}
    \caption{\textbf{\model{}.} The first Multimodal Large Language Model that understands and reasons on an input \nerf{}. Our framework  directly processes the \nerf{} weights and performs tasks such as captioning, Q\&A, and zero-shot classification of \nerf{}s.}
    \label{fig:teaser}
\end{figure}

The increasing adoption of NeRFs and their appealing characteristics prompted us to the following research question: is it possible to build an MLLM able to ingest directly \nerf{}s? Inspired by recent studies on meta-networks that can process neural fields~\cite{nf2vec, lim2024graph}, we answer this question in the positive by showing that it is possible to process the weights of a given \nerf{} with a meta-network encoder that projects the NeRF weights into the embedding space of a pre-trained LLM such as \llama{} 2~\cite{llama}. By doing so, we create the first MLLM for NeRFs, dubbed Large Language and \nerf{} Assistant (\model{}), which can solve \nerf-language tasks such as \nerf{} captioning, Q\&A and zero-shot \nerf{} classification (see~\cref{fig:teaser}). 

We also introduce a new \nerf--language dataset, that we will make publicly available, to train \model{} and test the capabilities of our assistant. 
To collect this dataset, we designed an automated annotation framework that leverages MLLMs to produce text annotations for \nerf{}s trained on Shapenet \cite{chang2015shapenet}. Using this dataset alongside an additional split containing manually curated textual descriptions \cite{amaduzzi2023looking}, we establish a benchmark for \nerf{} textual assistants. 

Since a straightforward way to create an assistant for \nerf{}s would be to render images or extract 3D point clouds out of it and provide them as input to existing MLLMs specifically designed to handle such modalities, we thoroughly compare \model{} against these baselines on the proposed benchmark. %We show how their ability to extract information about the input NeRF is more limited than when using the proposed framework. % suffer from two shortcomings. Firstly, extracting 3D information or rendering images out of NeRF is time-consuming, while a forward pass in the meta encoder is extremely fast. 
%Moreover, the effect of the vantage point to render images from or the resolution of the extracted 3D geometry on the quality of the output of the MLLM is not obvious. 
We show how the resolution of the extracted 3D geometry or images, and for images also the vantage point used for rendering, negatively impact the quality of the MLLM's output.
Important details might be lost by rendering from the wrong angle, or the extracted geometry might not be detailed enough. Vice versa, by operating directly on the MLP weights, we extract all the information they hold about the object without any other design decision. Our approach turns out to be the most effective way to create a NeRF assistant as it consistently outperforms MLLMs processing images or 3D geometries extracted by querying \nerf{}s.
%
%Thus, our proposed MLLM approach, namely Large Language and \nerf{} Assistant (\model{}), ingest the weights of a given \nerf{} with a meta-network encoder and then projects the network representation to the token input space of a pre-trained LLM such as Llama \cite{llama}, which yields a textual representation used for \nerf-language tasks such as \nerf{} captioning, Q\&A or Zero-shot \nerf{} classification (see \cref{fig:teaser}).
%
Our contributions can be summarized as follows:

$\bullet$ \model{}, the first MLLM capable of performing tasks such as captioning and Q\&A on \nerf{}s.

$\bullet$ We show that it is possible to build such an assistant by directly processing the \nerf{}s weights with a meta-encoder, which is faster and captures more information than rendering images or extracting  3D data.

$\bullet$ We automatically create a \nerf-language benchmark based on ShapeNet, and we thoroughly evaluate \model{} on it, showing that it performs better than applying popular MLLMs on discrete representations obtained from \nerf{}s.
\section{Related work}
\paragraph{Multimodal Large Language Models (MLLMs).}
Significant advancements have been made by Large Language Models (LLMs) in language understanding, reasoning, and generalization capabilities~\cite{ChatGPT, GPT4, InstructGPT, llama, flan, t5}.
These models have been extended into Multimodal Large Language Models (MLLMs), which broaden their reasoning abilities by including other modalities like images~\cite{Palm-E, LLaMA-Adapter, LLaMA-Adapterv2, image-bind}, audio~\cite{Audiogpt}, and videos~\cite{VideoChatGPT, VideoLLM}. MLLMs generally align target features with textual ones and then integrate them into LLMs for various text inference tasks. 
Some MLLMs are trained entirely from scratch~\cite{Kosmos-1, Kosmos-2}, others utilize pretrained LLMs~\cite{Otter, Qwen-VL, llava, BLIP-2, InstructBLIP}. 
3D MLLMs focus on understanding the 3D world typically represented as colored point clouds~\cite{gpt4point, 3dllm, 3d-vista, Point-bind, pointllm} or multi-view images~\cite{Hong_2023_CVPR}. Some of these models are trained using 2D images~\cite{3dllm, 3d-vista, Hong_2023_CVPR} while others directly align textual phrases with points~\cite{Point-bind, pointllm, gpt4point}.

\paragraph{Neural radiance fields.}
\nerf{}~\cite{nerf} have been applied in several visual tasks such as novel view synthesis~\cite{martin2021nerf}, generative media~\cite{poole2022dreamfusion}, and robotics~\cite{yen2022nerf}.
%, and computational photography~\cite{mildenhall2022nerf}.
The base formulation employs MLPs to convert spatial coordinates into colors and densities. 
Recent advancements substitute or enhance MLPs with explicit data structures~\cite{Chen2022ECCV, sun2022direct, Plenoxels, instant} for faster training and inference. 
%We employ the same base \nerf{} formulation as ~\cite{nf2vec}, i.e., each \nerf{} is a single MLP extracting density and color information for each 3D coordinate. Each MLP represents a different object.   % \nftovec{}

\paragraph{Neural radiance fields and language.}
The interaction between \nerf{} and language has been recently investigated for several practical applications. Many works address the problem of generating geometrically consistent views of objects or scenes described by textual prompts~\cite{seo2023dittonerf, Metzer_2023_CVPR, Jo_2023_WACV, NEURIPS2023_a0303731, li2024instructpixnerf, lee2022understanding, poole2022dreamfusion}. 
Other approaches focus on editing the scene represented by a~\nerf{} from text, e.g., by changing the appearance and shape of objects~\cite{wang2022clip, Hwang_2023_ICCV, Song_2023_ICCV, 10144678, Sun_2024_WACV, Haque_2023_ICCV, 10476703, zhuang2023dreameditor}, or by inserting/removing objects in the scene~\cite{bai2023componerf, Mirzaei_2023_ICCV}.
Some techniques investigate new types of radiance fields that predict language features for each spatial location alongside density and color~\cite{lerf2023, NEURIPS2022_93f25021}. By distilling knowledge from vision-language models into these models, the neural fields can be queried by textual prompts. LERF~\cite{lerf2023} extends the original radiance field formulation, considering functions which model density, color and language features at each spatial coordinate. Such \emph{language fields} are parametrized by a neural network.
%
%\textcolor{red}{\textbf{Unlike all previous methods, \citet{ballerini2024clip2nerf} is the first to utilize \nerf{}s as an input modality.}} 
Unlike all previous methods, \citet{ballerini2024clip2nerf} is the first to utilize the weights of a \nerf{}'s MLP as an input modality. They aim to learn a mapping between the \nerf{} and CLIP~\cite{clip} embedding spaces to perform tasks such as NeRF retrieval from textual or image queries. 
%Similarly, we consider \nerf{}s as an input modality to our framework. 
Differently, our goal is to develop an MLLM capable of reasoning about  \nerf{}s.

\paragraph{Deep learning on neural networks.}
Several studies have explored using meta-networks, i.e. neural networks that analyze other neural networks. Initially, researchers concentrated on predicting network characteristics, such as accuracy and hyperparameters, by processing their weights~\cite{Unterthiner2020PredictingNN,urholt2021selfsupervised, knyazev2021parameter,jaeckle2021generating,Lu2020Neural}.
Several recent works focus on processing networks implicitly representing data (Implicit Neural Representations or Neural Fields). These methods perform tasks such as classifying or segmenting the data by processing solely the weights of the input neural networks.
Among these works, Functa~\cite{functa} trains a shared network on the entire dataset and then learns a compact embedding for each sample for downstream tasks. 
Later works concentrate on processing networks representing individual data samples, e.g., a specific object. 
By leveraging a novel encoder architecture for MLP weights, \inrtovec{}~\cite{deluigi2023inr2vec} extracts compact embeddings from INRs of 3D shapes, which are employed as inputs for downstream tasks. 
\nftovec{}~\cite{ramirez2023deep} extends \inrtovec{} to ingest the \nerf's network weights to classify, segment, or retrieve similar \nerf{}s. 
%Notably, \nftovec{} can process even large MLPs thanks to the efficient design of the encoder. 
\citet{cardace2024neural} develop a strategy to process neural fields represented by a hybrid tri-plane structure.  
Other approaches~\cite{navon2023equivariant, zhou2023neural, zhou2023permutation, zhou2024universal} develop equivariant architectures to handle MLPs by exploiting weight space symmetries~\cite{hecht1990algebraic} as an inductive bias. 
Also, Graph Neural Networks have been investigated to compute a network representation~\cite{kofinas2024graph, lim2024graph}. 
%To improve the generalization of approaches processing neural networks, \cite{shamsian2024improved} explores various strategies for data augmentation directly in weight spaces.
%
Since we aim to process \nerf s directly from the network weights, we employ \nftovec{} as our meta-encoder due to its efficient and scalable architecture.
\section{Methodology}
\label{sec:method}
This section describes the proposed Large Language and \nerf{} Assistant (\model{}). We provide an overview of \nerf{}s and the meta-encoder that maps NeRF weights into a global embedding. Then, we present the overall LLaNA framework and discuss our training protocol.

\begin{figure}
    \centering
    \includegraphics[width=0.9\linewidth]{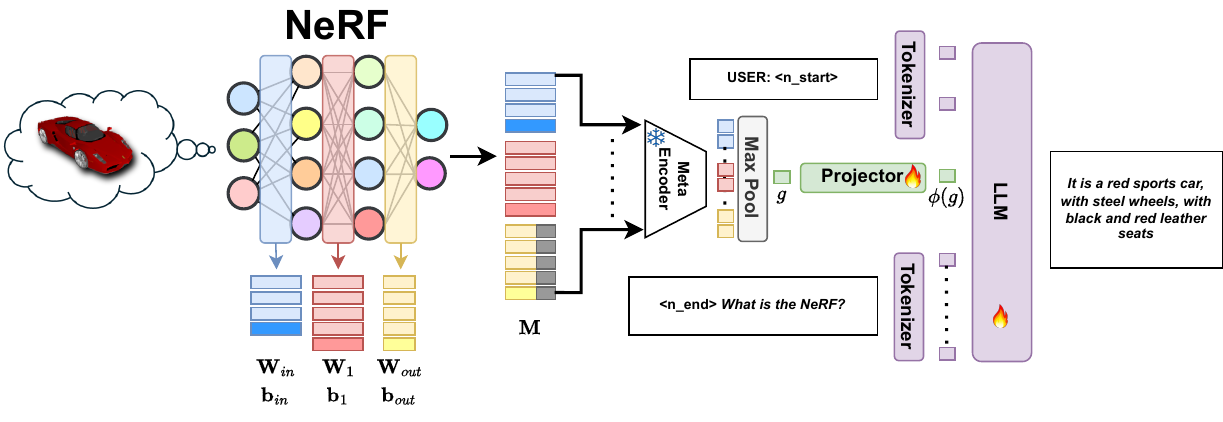}
    \caption{\textbf{Framework overview. Example of \nerf{} captioning.}}
    \label{fig:framework}
\end{figure}

\paragraph{Neural Radiance Fields (\nerf{})}
%\label{sec:nerf}
Neural Radiance Field (NeRF)~\cite{nerf} is a framework that employs coordinate-based neural networks, typically MultiLayer Perceptrons (MLP) and is trained on a collection of images of an object or scene taken from various vantage points. The main application of NeRFs is the task of novel views synthesis, i.e., photorealistic rendering of images from viewpoints unseen at training time.
In its base formulation, the MLP is a function of continuous 3D coordinates $\vb{p} = (x, y, z) \in \mathbb{R}^3$, that yields four-dimensional outputs, $RGB\sigma \in [0,1]^4$. 
This output encodes the $RGB$ color and the volume density $\sigma$ of each 3D location in the scene. The volume density $\sigma$ can be interpreted as the differential probability of a ray terminating at point $\vb{p}$. 
After training, a NeRF can render images from any desired viewpoint at arbitrary resolution by querying it for the values of $RGB$ and $\sigma$ at several points along the ray corresponding to each pixel and applying the volumetric rendering equation~\cite{nerf}.

In this work, we realize NeRFs as MLPs composed of $L$ hidden layers, an input layer, and an output layer. An example of MLP with 1 input, 1 output, and 1 hidden layer is shown in~\cref{fig:framework} (left).
A layer is parameterized by a weight matrix plus a bias vector. More in detail, the hidden layers in our architecture have the same number of input and output neurons, $H$, thus having squared weight matrices $\mathbf{W}_{l} \in \mathbb{R}^{H \times H}$ for $l=1,\dots,L$ and $H$-dimensional biases $\mathbf{b}_l \in \mathbb{R}^{H}$. 
As input $\vb{p}$ goes through a 24-frequency encoding~\cite{nerf}, the first layer has $\mathbf{W}_{in} \in \mathbb{R}^{144 \times H}$ and $\mathbf{b}_{in} \in \mathbb{R}^{H}$. The final one has $\mathbf{W}_{out} \in \mathbb{R}^{H \times 4}$ and $\mathbf{b}_{out} \in \mathbb{R}^{4}$. Refer to \cref{sec:nerf_details} for more details on NeRFs.
%
%Due to their many advantages over other data representations, NeRFs have become a new standard for describing the world. They have become a new input modality, stored and communicated independently.

\paragraph{Meta-encoder}
%\label{sec:meta-encoder}

In this work, we explore how a NeRF assistant can be realized by processing the NeRF weights directly. We expect the NeRF weights to contain comprehensive information about the represented object, such as its geometry and appearance. Thus, an encoder processing them might extract all the necessary information for downstream language tasks such as captioning and Q\&A.

Inspired by the recent development of meta-networks capable of processing neural fields~\cite{lim2024graph, nf2vec}, we employ as our meta-encoder architecture \nftovec{} \cite{nf2vec}. It takes as input the weights of a NeRF and yields a global embedding that distills the content of the input \nerf{}. 
In particular, the weight matrices and biases of the input NeRF are stacked along the row dimension to form a matrix $\mathbf{M} \in \mathbb{R}^{S \times H}$, where the number of rows $S$ depends on the number of hidden layers $L$, the number of units per hidden layer $H$, and the dimension of the input, which is a $144$-dimensional array obtained by frequency encoding of the 3D coordinates. Before stacking, we pad the output layer weights $\mathbf{W}_{out}$ and biases $\mathbf{b}_{out}$ with zeros to obtain $H$ columns (see  \cref{fig:framework}, center).

The meta-encoder is parametrized as an MLP with batch normalization layers~\cite{ioffe2015batch} and ReLU non-linearities. To scale gracefully with the input MLP dimensions, the encoder processes each row of $\mathbf{M}$ independently, extracting a total of $S$ tokens, each of length $G$, from an input NeRF. They are then max-pooled to obtain a global representation $g \in  \mathbb{R}^{G}$ of the \nerf{}, with $G=1024$ in our experiments.
The encoder is pre-trained using the self-training protocol of \nftovec{}~\cite{nf2vec}, i.e., jointly with a decoder architecture that, given as input the \nerf{} global embedding, reconstructs the same images as the input \nerf{} from arbitrary viewpoints. More details in \cref{sec:nftovec_supp}.

\paragraph{Large language and NeRF assistant}
%\label{sec:meta-encoder}
Inspired by recent approaches that created effective Multimodal Large Language Models, we build LLaNA by leveraging on a pre-trained LLM with a transformer backbone~\cite{transformers}, in our experiments \llama{}~2~\cite{llama}, and injecting the NeRF modality into its embedding input space, as proposed for images and 3D data~\cite{llava, pointllm} (see \cref{fig:framework}, right). Thanks to the self-attention mechanism, the transformer can understand the contextual relationships between text and \nerf{} tokens, enabling it to generate responses based on both text and \nerf{} inputs. 

We employ a trainable linear projection layer, $\phi$, to project the embedding of the input \nerf{} computed by the meta-encoder into the \llama{} 2 embedding space. The projection layer has weights $\mathbf{W}_{proj} \in \mathbb{R}^{G \times T}$, where $T$ is the word embedding dimension of the employed \llama{} model. This embedding is encapsulated between two special tokens, whose embeddings are learned end-to-end while training, namely <n\_start> and <n\_end>. 

Then, given an input sequence of mixed \nerf{} and word tokens, $($<n\_start>$,\phi(g), $<n\_end>$, w_1, w_2, ..., w_k)$, where $k$ is the number of word tokens, the large language model returns a sequence of predicted word tokens $(\hat{w}_{k+1}, \hat{w}_{k+2}, \dots, \hat{w}_{eos})$. 
%Each mapped token is transformed into a probability distribution over a word vocabulary, and the prediction is the word with the highest probability.

\paragraph{Training protocol}
%\label{sec:training}
%To train our framework, we hold multi-turn conversations about each \nerf{} available in the ShapeNeRF--Text dataset that we created (see \cref{sec:dataset}). 
Our framework is trained on the ShapeNeRF--Text dataset, described in detail in \cref{sec:dataset}.
This dataset is organized into a set of prompts from the user and expected ground-truth answers that are used to optimize the original auto-regressive objective of the LLM. 
For the meta-encoder, we employ the \nftovec{} encoder pre-trained on ShapeNet released by the authors~\cite{nf2vec}, and we keep it frozen during training. We follow the two-stage training protocol delineated in \citet{llava}:

%$\bullet$ 
\emph{Stage1: projector training.}
In the first stage, we train the projector network $\phi$ to align the \nerf{} and the word embedding spaces while keeping the LLM weights fixed. We train on an instruction dataset of brief descriptions to learn the projection layer efficiently. We also train the embeddings of the special tokens used to encapsulate the \nerf{} one. We optimize the projector weights and the embeddings for $3$ epochs with a learning rate of $0.002$ and batch size of $64$.
        
%$\bullet$ 
\emph{Stage2: instruction tuning.} During the second stage, we train on complex instructions to help the model understand and reason about \nerf{} data. In this phase, we optimize both the projector and the LLM for $3$ epochs on the detailed descriptions, single-round and multi-round Q\&A conversations available in our dataset. For this phase, we employ a learning rate of $0.0002$ and a batch size of $16$.

Our model is implemented in PyTorch and trained on 4 NVIDIA A100 with 64GB of VRAM each. Completing both stages requires $\sim$1 day of training.
%As for the safeguards of the LLMs employed, both LLaVA2-7b and Llama3-8b-chat are equipped with such mechanisms. 
%When fine-tuning \llama{} 2 for our specific applications, we strictly use a dataset that does not come from scraped web data. This approach should ensure that the integrity and effectiveness of the pre-existing safeguards are maintained.

%We train \model{} on a NeRF-Language dataset, ShapeNeRF-Text, described in \cref{sec:dataset}. In Stage 1, we train the model, . Then, in Stage 2, we optimize LLaMA 2 and the projector for 3 epochs on detailed descriptions, single-round and multi-round Q\&A with a learning rate of $0.0002$ and batch size 16. 
\section{Benchmark}
\label{sec:dataset}

\subsection{ShapeNeRF--Text dataset}
\label{sec:shapenerf}
To train and validate our NeRF assistant, we automatically created a dataset of conversations about NeRFs, the ShapeNeRF--Text dataset. 

\begin{wrapfigure}{r}{5.0cm}
    \centering
    \includegraphics[width=0.70\linewidth]{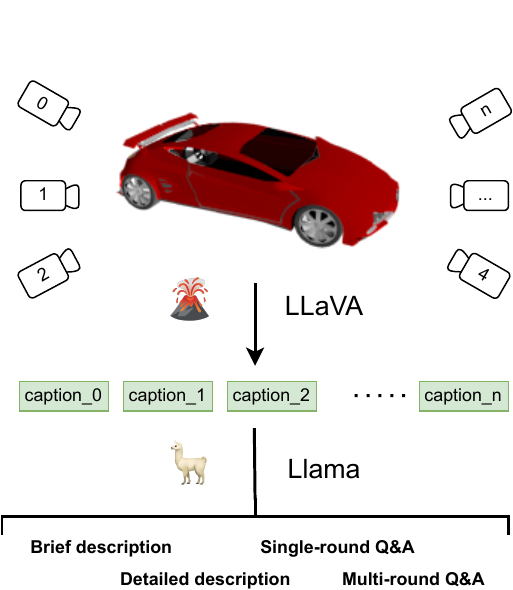}
    \caption{\textbf{Automatic annotation pipeline.} Given a 3D model, $N$ views are rendered and processed by a VLM (LLaVA) to generate view-specific captions. These are aggregated by an LLM (LLaMA) for final descriptions and Q\&A.}
    \label{fig:dataset}
\end{wrapfigure}

It features paired \nerf{}s and language annotations for ShapeNet objects~\cite{chang2015shapenet}, in particular for all the 40K \nerf{}s available in the \nftovec{} dataset~\cite{ramirez2023deep}.
%Regarding the \nerf{}s, we employ the dataset released by \nftovec{}, which provides 30K pre-trained \nerf{}s, one for each object. 
%
%
%
We followed the structure defined in PointLLM~\cite{pointllm} to create the textual annotations. More in detail, for each object, we generated a \emph{brief description}, a \emph{detailed description}, 3 \emph{single-round Q\&As}, and one \emph{multi-round Q\&A}. 
The brief descriptions are concise captions of the object, taking into account its global structure and appearance. 
The detailed descriptions are longer sentences that describe all the details of the object. 
The single-round Q\&As consist of a question about the object and the corresponding ground-truth answer. 
Finally, the multi-round Q\&As are longer conversations formed by 3 questions and the relative answers.
The automatic data annotation pipeline is inspired by Cap3D~\cite{cap3d} and is shown in~\cref{fig:dataset}.  
First, multiple views of each ShapeNet object have been rendered from different perspectives. 
Then, each view has been provided as input to \llava{} (\llava{}2-13b)~\cite{llava} to get a detailed description of the object from that point of view. 
Afterward, starting from the captions generated by \llava{}, \llama{} 3 (\llama{}3-8B-chat) was used to generate the final ground-truth text data (brief and detailed descriptions, single and multi-round Q\&As).
%
%The aggregation via LLMs of multiple captions obtained from different rendered views has been introduced by Cap3D \cite{cap3d}, and it is crucial for generating correct descriptions of 3D data.
%
Both the frozen LLMs employed to create our benchmark (\llava{}2-13b, \llama{}3-8b-chat) are equipped with safeguards.  

When building the ground-truth data, to ensure diversity in the language annotations, each brief and detailed description has been associated with a question randomly sampled from 30 instructions for each kind of description. 
Such instructions, together with the carefully engineered request prompts for \llava{} and \llama{}, are reported in \cref{sec:instruction_prompts_gt_questions}. 
%The main strengths of this annotation pipeline are twofold: first, it is fully automatic, and it does not rely on any human annotation; secondly, the use of \llama{} as LLM for the aggregation of the single captions ensures that this pipeline is completely free to use. The same cannot be said of Pyramid-XL, which, conversely, has a total cost of $\sim$\$9000 as it relies on GPT-4 APIs.
%
%ShapeNeRF--Text comprises 40K objects, with 240K textual annotations: 40K brief captions, 40K detailed descriptions, 120K single-round Q\&A, and 40K three-round Q\&A.

ShapeNeRF--Text provides 30939, 3846 and 3859 objects for the train, validation and test sets, respectively. Overall, the dataset features 13 object classes, and the train, validation and test splits are obtained by randomly sampling objects within each class, i.e., holding out a fixed percentage of objects per class (80\%, 10\%, and 10\% for the sets, respectively).
\cref{sec:dataset_stats} provides more dataset statistics.
As quantitatively proven in \cref{sec:data_quality_analysis} and \cref{sec:language_baseline_supp}, many of the questions belonging to the Q\&A set require a holistic 3D understanding of the object, to be answered correctly.
%We split the dataset into train, validation, and test sets, stratified on the ShapeNet classes, containing distinct objects in proportions of 80\%, 10\%, and 10\% of the total number of shapes, respectively.

\subsection{Language tasks and metrics}
\label{sec:benchmark}
We evaluate NeRF assistants on three different language tasks, given an input NeRF: brief captioning, detailed captioning, and single-round Q\&A. We evaluate all tasks on the objects from the ShapeNeRF--Text test set. For brief captioning, we additionally evaluate the methods on the GPT2Shape Human Shape Text (HST) dataset~\cite{amaduzzi2023looking}, a subset of ShapeNet for which human-curated brief descriptions are publicly available. To generate the dialogues for HST, we randomly pair each of its captions with one of the 30 instructions requesting a brief description, used in ShapeNeRF--Text and reported in \cref{sec:instruction_prompts_gt_questions}. 
We employ standard language similarity metrics to evaluate these methods. We compute the cosine similarity between the global embeddings of the generated and ground-truth sentences provided by the pre-trained encoders Sentence-BERT~\cite{sentencebert} and SimCSE~\cite{simcse}. These metrics based on learned networks are the most effective at measuring the quality of the generated output. We also include standard handcrafted metrics based on n-gram statistics, like BLEU-1~\cite{bleu}, ROUGE-L~\cite{rouge}, and METEOR~\cite{meteor}.
\section{Experiment results}
\label{sec:experiments}
%\textcolor{red}{In this section, we compare \model{} with multiple MLLM baselines, which have been evaluated as off-the-shelf foundation models or after training on ShapeNeRF--Text. Finally, in \cref{sec:generalization_supp} we provide experiments to assess the generalization capability of \model{} on Objaverse~\cite{deitke2023objaverse}.}

\subsection{Foundation models as baselines}
\label{sec:baselines}
\begin{figure}[t]
    \begin{minipage}{0.48\linewidth}
    \captionof{table}{\textbf{Zero-Shot NeRF Classification. Frozen baselines.}\\Best results are in \textbf{bold}, runner-up is \underline{underlined}. \\(FV: front-view, BV: back-view, MV: multi-view)}
    \begin{center}
        \resizebox{0.6\linewidth}{!}{%
        \begin{tabular}{lcc}
        \midrule
        \textbf{Model} & \textbf{Modality} & \textbf{Accuracy (\%)} \\
        \cmidrule(lr){1-1} \cmidrule(lr){2-2} \cmidrule(lr){3-3}
        \llava{}-vicuna-13b & Image (FV) & 66.13 \\
        \llava{}-vicuna-13b & Image (BV) & 63.90 \\
        \llava{}-vicuna-13b & Image (MV) & \textbf{73.45} \\
        \llava{}-vicuna-7b & Image (FV) & 60.25 \\
        \llava{}-vicuna-7b & Image (BV) & 57.00 \\
        BLIP-2 FlanT5-xxl & Image (FV) & 63.67 \\
        BLIP-2 FlanT5-xxl & Image (BV) & 61.47 \\
        \cmidrule(lr){1-1} \cmidrule(lr){2-2} \cmidrule(lr){3-3}
        PointLLM-7b & Point cloud & 50.14 \\
        GPT4Point-Opt-2.7b & Point cloud & 41.93 \\
        \cmidrule(lr){1-1} \cmidrule(lr){2-2} \cmidrule(lr){3-3}
        3D-LLM & Mesh + MV & 60.55 \\
        \cmidrule(lr){1-1} \cmidrule(lr){2-2} \cmidrule(lr){3-3}
        \model{}-7b & \nerf{} & \underline{67.14} \\
        \bottomrule
        \end{tabular}}
    \label{tab:classification_frozen}
    \end{center}
    \end{minipage}
    \hfill
    \begin{minipage}{0.48\linewidth}
    \begin{center}
        \includegraphics[width=0.7\linewidth]{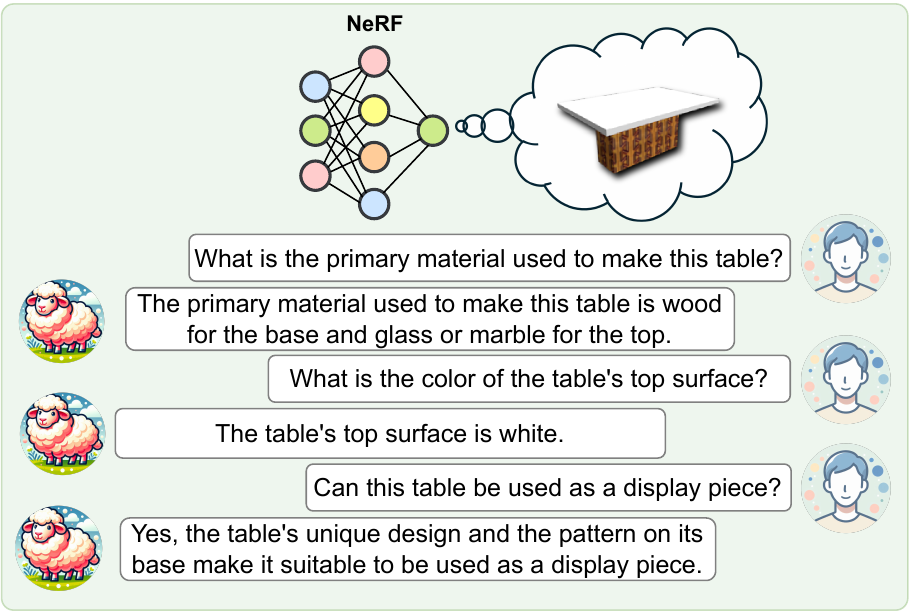}
        \captionof{figure}{\textbf{NeRF multi-round Q\&A example.}}
        \label{fig:multi-round}
    \end{center}
    \end{minipage}
\end{figure}

As our method is the first to investigate language tasks on \nerf{}, there are no baselines in the literature. However, given a NeRF, a straightforward way to create an assistant for it could be to render an image and use an MLLM capable of ingesting images. Alternatively, we could extract the 3D shape from the NeRF and use one of the recent 3D MLLMs. Hence, in a first set of experiments, we use MLLMs as off-the-shelf foundation models, trained on hundreds of thousands of shapes or millions of images, without performing any fine-tuning on the training set of ShapeNeRF--Text, and consider such pipelines as natural baselines. Specifically, we use \llava{} (v1.6)~\cite{llava} and BLIP-2~\cite{blip2} for images, as well as PointLLM~\cite{pointllm} and GPT4Point \cite{gpt4point} for colored point clouds. 
Since NeRFs can render arbitrary viewpoints after training, we also include the evaluation of LLaVA~\cite{llava} in a multi-view scenario. 
More in detail, we render images from $N$ viewpoints randomly sampled between the set of camera poses used to train each NeRF; then, we concatenate tokens from these N images and fed them into LLaVA alongside text instructions. We set $N$=$3$ because the model cannot process a higher number of images correctly.
%following the procedure detailed in \cref{sec:multiview_llava}.
%We have realized a multi-view baseline for LLaVA by rendering images from $N$ viewpoints randomly sampled between the set of camera poses used to train each NeRF. Then, we concatenated tokens from the N images and fed them into LLaVA alongside text instructions. We set $N$=$3$ because the model cannot process a higher number of images correctly. Results are reported in \cref{tab:brief_frozen}, \cref{tab:brief_frozen_hst}, \cref{tab:detailed_frozen}, \cref{tab:qa_frozen} and \cref{tab:classification_frozen}.
In addition, we test 3D-LLM~\cite{3dllm} to compare its performance to LLaNA. We employ the official code and pre-trained models released by the respective authors for such evaluations~\footnote{
\llava{}: \url{https://github.com/haotian-liu/LLaVA}
BLIP-2: \url{https://github.com/salesforce/LAVIS/tree/main/projects/blip2}
PointLLM: \url{https://github.com/OpenRobotLab/PointLLM}
GPT4Point: \url{https://github.com/Pointcept/GPT4Point}
3D-LLM: \url{https://github.com/UMass-Foundation-Model/3D-LLM}}. 
We note that the only official GPT4Point weights available at submission time were those obtained from fine-tuning OPT-2.7B on Cap3D~\cite{cap3d}.
%We highlight that all competitors have been trained on much larger datasets than our method, i.e., $\sim$660K shapes of Objaverse-1.0~\cite{deitke2023objaverse} for PointLLM and GPT4Point, or over one million paired visual-language data for LLaVA.
 % without and do not experiment on how they could have performed had they undergone fine-tuning on the training set of.
%
In \cref{tab:brief_frozen,tab:brief_frozen_hst,tab:detailed_frozen,tab:qa_frozen,tab:qa_frozen,tab:classification_frozen}, we present the performance of all methods under the more realistic scenario where \nerf{}s are treated as the only input data to the assistant. Hence, images and point clouds can only be extracted from \nerf{}s. Details on the extraction procedure are provided in \cref{sec:extract_data_from_nerf}. 
As for 3D-LLM, we extract colored 3D meshes from the NeRFs of ShapeNeRF--Text and process such data with the official 3D-LLM code to render images from multiple views and compute both the 2D and 3D features required by the model at inference time.  Moreover, in \cref{sec:gt_mesh_supp},  we report the results dealing with the images used to train the NeRF or the original 3D point cloud from ShapeNet, which confirms the methods' ranking. When rendering an image, a non-obvious design decision for the pipeline is from which vantage point to render it. ShapeNet artificially simplifies this task since all objects have been canonically aligned to a common reference frame, but this may not be the case in a general setting. To show the vantage point's effect on the assistant's results, we report results processing a frontal or back view. 

%%% TABLE OF FROZEN BASELINES WITH NEW FORMAT
%%% AGGREGATED FROZEN AND TRAINED TABLES

\begin{table}[t]
    \centering
    \begin{minipage}{0.49\linewidth}
    \caption{\textbf{NeRF brief captioning on ShapeNeRF-Text. Frozen baselines.} \\Best results are in \textbf{bold}, runner-up is \underline{underlined}. \\(FV: front-view, BV: back-view, MV: multi-view)}
    \resizebox{\linewidth}{!}{%
        \begin{tabular}{lcccccc}
        \midrule
        \textbf{Model} & \textbf{Modality} & \textbf{S-BERT} & \textbf{SimCSE} & \textbf{BLEU-1} & \textbf{ROUGE-L} & \textbf{METEOR} \\
        \cmidrule(lr){1-1} \cmidrule(lr){2-2} \cmidrule(lr){3-7}
        \llava{}-vicuna-13b & Image (FV) & \underline{61.00} & 61.16 & 14.30 & 20.00 & 23.31 \\
        \llava{}-vicuna-13b & Image (BV) & 54.35 & 56.09 & 21.94 & 21.67 & 22.09 \\
        \llava{}-vicuna-13b & Image (MV) & 59.64 & 61.01 & \textbf{22.84} & 22.17 & 23.08 \\
        \llava{}-vicuna-7b & Image (FV) & 59.85 & \underline{62.35} & \underline{22.67} & \underline{23.24} & \underline{23.35} \\
        \llava{}-vicuna-7b & Image (BV) & 55.68 & 58.46 & 21.97 & 22.46 & 22.50 \\
        BLIP-2 FlanT5-xxl & Image (FV) & 56.13 & 58.21 & 5.46  & 18.69 & 9.67 \\
        BLIP-2 FlanT5-xxl & Image (BV) & 52.48 & 54.05 & 5.67  & 18.20 & 9.50 \\
        \cmidrule(lr){1-1} \cmidrule(lr){2-2} \cmidrule(lr){3-7}
        PointLLM-7b & Point cloud & 49.59 & 48.84 & 16.74 & 17.92 & 14.56 \\
        GPT4Point-Opt-2.7b & Point cloud & 41.85 & 40.22 & 11.76 & 16.54 & 11.63 \\
        \cmidrule(lr){1-1} \cmidrule(lr){2-2} \cmidrule(lr){3-7}
        3D-LLM & Mesh + MV &  59.46 & 56.42 & 12.69 & 21.49 & 14.32 \\
        \cmidrule(lr){1-1} \cmidrule(lr){2-2} \cmidrule(lr){3-7}
        \model{}-7b & \nerf{} & \textbf{68.63} & \textbf{70.54} & 20.64 & \textbf{28.33} & \textbf{31.76} \\
        \bottomrule
        \end{tabular}}
        \label{tab:brief_frozen}
    \end{minipage}
    \hfill
    \begin{minipage}{0.49\linewidth}
        \caption{\textbf{NeRF brief captioning on the HST dataset. Frozen baselines.} \\Best results are in \textbf{bold}, runner-up is \underline{underlined}. \\(FV: front-view, BV: back-view, MV: multi-view)}
        \resizebox{\linewidth}{!}{%
        \begin{tabular}{lcccccc}
        \midrule
        \textbf{Model} & \textbf{Modality} & \textbf{S-BERT} & \textbf{SimCSE} & \textbf{BLEU-1} & \textbf{ROUGE-L} & \textbf{METEOR} \\
        \cmidrule(lr){1-1} \cmidrule(lr){2-2} \cmidrule(lr){3-7}
        \llava{}-vicuna-13b& Image (FV) & 55.62 & 55.56 &  6.56 & 11.81 & 14.52 \\
        \llava{}-vicuna-13b& Image (BV)    & 50.00 & 50.79 & 9.39 & 12.76 & 14.46 \\
        \llava{}-vicuna-13b & Image (MV) & 54.25 & 55.56 & 9.78 & 14.13 & 14.99 \\
        \llava{}-vicuna-7b& Image (FV)  & 54.31 & 56.28 & 10.08 & 14.71 & 14.53 \\
        \llava{}-vicuna-7b& Image (BV)     & 51.75 & 52.29 & 8.13 & 13.96 & 14.18 \\
        BLIP-2 FlanT5-xxl& Image (FV)& \underline{57.11} & \underline{59.43} &  8.21 & 18.02 & 12.14 \\
        BLIP-2 FlanT5-xxl& Image (BV)   & 54.11 & 56.37 &  9.09 & 17.38 & 11.79 \\
        \cmidrule(lr){1-1} \cmidrule(lr){2-2} \cmidrule(lr){3-7}
        PointLLM-7b & Point cloud                      & 43.40 & 44.50 &  8.53 & 11.64 &  9.97 \\
        GPT4Point-Opt-2.7B & Point cloud                        & 43.15     & 42.22     & \underline{12.02}     & \underline{18.73}    & 13.69     \\
        \cmidrule(lr){1-1} \cmidrule(lr){2-2} \cmidrule(lr){3-7}
        3D-LLM & Mesh + MV &  56.07 & 52.13 & \textbf{15.94} & \textbf{20.71} & \underline{15.22} \\
        \cmidrule(lr){1-1} \cmidrule(lr){2-2} \cmidrule(lr){3-7}
        \model{}-7b & \nerf{}                      & \textbf{59.20} & \textbf{61.66} &  9.47 & 14.94 & \textbf{17.06} \\
        \bottomrule
        \end{tabular}}
        \label{tab:brief_frozen_hst}
    \end{minipage}
\end{table}
            
%%% AGGREGATED FROZEN AND TRAINED TABLES

\begin{table}[t]
    \caption{\textbf{NeRF detailed captioning on ShapeNeRF-Text. Frozen baselines.}\\Best results are in \textbf{bold}, runner-up is \underline{underlined}. (FV: front-view, BV: back-view, MV: multi-view)}
    \centering
    \resizebox{0.7\linewidth}{!}{
    \begin{tabular}{lcccccc}
        \midrule
        \textbf{Model} & \textbf{Modality} & \textbf{S-BERT} & \textbf{SimCSE} & \textbf{BLEU-1} & \textbf{ROUGE-L} & \textbf{METEOR} \\
        \cmidrule(lr){1-1} \cmidrule(lr){2-2} \cmidrule(lr){3-7}
        \llava{}-vicuna-13b& Image (FV) & 59.08 & 58.87 & \underline{23.63} & 23.55 & \underline{22.55} \\
        \llava{}-vicuna-13b& Image (BV)    & 50.09 & 50.33 & 13.77 & 21.36 & 13.18 \\
        \llava{}-vicuna-13b & Image (MV) & \underline{60.21} & \underline{59.51} & 15.07 & \underline{32.16} & 14.64\\
        \llava{}-vicuna-7b& Image (FV)  & 57.55 & 57.68 & 14.99 & 22.82 & 14.36 \\
        \llava{}-vicuna-7b& Image (BV) & 53.11 & 54.46 & 14.73 &	22.47 &	14.05 \\
        BLIP-2 FlanT5-xxl& Image (FV)& 41.27 & 40.69 &  0.18 &  7.83 &  2.60 \\
        BLIP-2 FlanT5-xxl& Image (BV)   & 38.49 & 37.89 &  0.19 &  7.72 &  2.58 \\
        \cmidrule(lr){1-1} \cmidrule(lr){2-2} \cmidrule(lr){3-7}
        PointLLM-7b & Point cloud                      & 59.02 & 58.30 & 10.28 & 19.26 & 10.55 \\
        GPT4Point-Opt-2.7b & Point cloud  & 42.44     & 38.33  & 3.72     & 9.21 & 5.13    \\
        \cmidrule(lr){1-1} \cmidrule(lr){2-2} \cmidrule(lr){3-7}
        3D-LLM & Mesh + MV &  60.00 & 53.91 & 1.58 & 14.40 & 5.28 \\
        
        \cmidrule(lr){1-1} \cmidrule(lr){2-2} \cmidrule(lr){3-7}
        \model{}-7b & \nerf{}                      & \textbf{77.43} & \textbf{79.81} & \textbf{41.32} & \textbf{36.18} & \textbf{32.39} \\
        \bottomrule
    \end{tabular}}
    \label{tab:detailed_frozen}
\end{table}
%%% AGGREGATED FROZEN AND TRAINED TABLES

\begin{table}[t]
    \centering
    \caption{\textbf{NeRF single-round Q\&A on ShapeNeRF-Text. Frozen baselines.}\\Best results are in \textbf{bold}, runner-up is \underline{underlined}. (FV: front-view, BV: back-view, MV: multi-view)}
        \resizebox{0.7\linewidth}{!}{%
        \begin{tabular}{lcccccc}
        \midrule
        \textbf{Model} & \textbf{Modality} & \textbf{S-BERT} & \textbf{SimCSE} & \textbf{BLEU-1} & \textbf{ROUGE-L} & \textbf{METEOR} \\
        \cmidrule(lr){1-1} \cmidrule(lr){2-2} \cmidrule(lr){3-7}
        \llava{}-vicuna-13b& Image (FV) & 71.61 & 70.98 & 20.19 & 30.42 & 32.53 \\
        \llava{}-vicuna-13b& Image (BV)    & 68.25 & 69.06 & 20.03 & 29.84 & 32.27 \\
        \llava{}-vicuna-13b & Image (MV) & 71.84 & 71.16 & 20.04 & 30.20 & 33.46 \\
        \llava{}-vicuna-7b& Image (FV)  & 71.79 & 71.96 & 25.79 & 34.04 & 34.86 \\
        \llava{}-vicuna-7b& Image (BV)     & 70.88 & 70.93 & 25.17 & 33.30 & 34.22 \\
        BLIP-2 FlanT5-xxl& Image (FV) & 45.20 & 47.92 & 11.50 & 20.16 & 13.49 \\
        BLIP-2 FlanT5-xxl& Image (BV)   & 45.06 & 47.66 & 11.50 & 19.98 & 13.44 \\
        \cmidrule(lr){1-1} \cmidrule(lr){2-2} \cmidrule(lr){3-7}
        PointLLM-7b & Point cloud                      & \underline{74.70} & \underline{74.40} & \underline{36.81} & \underline{44.41} & \underline{39.76} \\
        GPT4Point-Opt-2.7b & Point cloud                        & 27.62     & 31.41     & 6.26     & 9.38     & 5.41     \\
        \cmidrule(lr){1-1} \cmidrule(lr){2-2} \cmidrule(lr){3-7}
        3D-LLM & Mesh + MV & 69.62 & 67.55 & 32.19 & 40.95 & 35.83\\
        \cmidrule(lr){1-1} \cmidrule(lr){2-2} \cmidrule(lr){3-7}
        
        \model{}-7b & \nerf{}                      & \textbf{81.03} & \textbf{81.56} & \textbf{46.16} & \textbf{53.17} & \textbf{50.15} \\
        \bottomrule
        \end{tabular}}
    \label{tab:qa_frozen}
\end{table}

\begin{figure}[t]
    \centering
    % \begin{tabular}{c}
    %     \includegraphics[width=\linewidth]{figures/qualitative_captioning.pdf} \\
    %     \includegraphics[width=\linewidth]{figures/single_round_qa.pdf} \\
    % \end{tabular}
    \includegraphics[width=\linewidth]{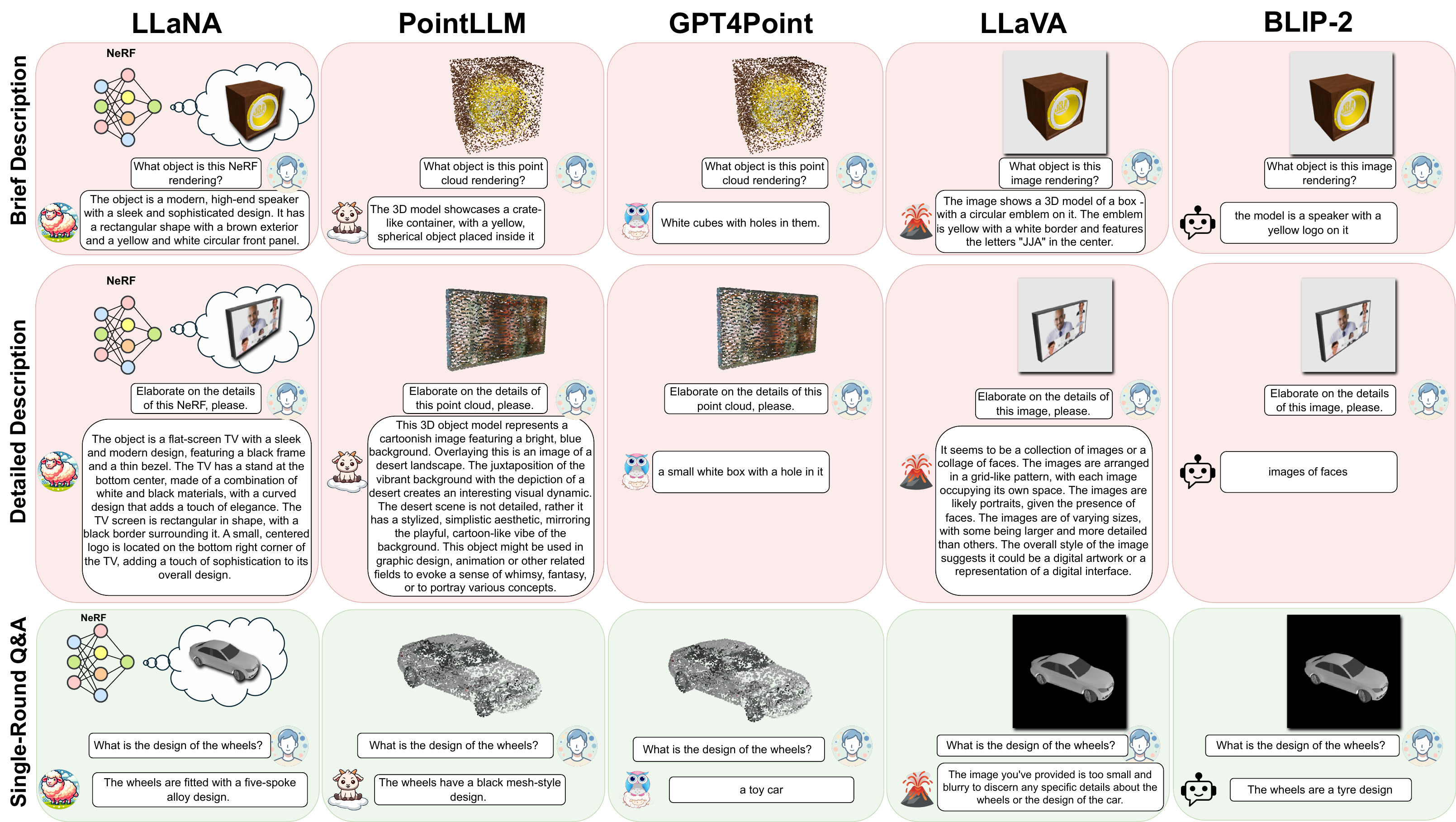}
    \caption{\textbf{Qualitative results of \nerf{} captioning and  Q\&A.} Results on ShapeNeRF--Text. From top to bottom: brief and detailed descriptions, single-round Q\&A}
    \label{fig:qualitative_results}
\end{figure}
%%% AGGREGATED FROZEN AND TRAINED TABLES

\subsection{NeRF captioning}
\label{sec:nerf_captioning}
We test the assistants' ability to describe the NeRF content in the captioning experiments. We prompt them with the NeRF, or the image/cloud extracted from it, followed by the question which has been paired with its ground-truth description, as detailed in Section~\ref{sec:benchmark}, 
e.g. \emph{"What's the content of this NeRF/image/cloud?"}. We then collect the answers generated by the models and compare them with the ground-truth description according to the selected metrics.   

\paragraph{Brief description.}

We report results for the brief description tasks on ShapeNeRF--Text and the HST dataset in \cref{tab:brief_frozen} and \cref{tab:brief_frozen_hst}, respectively.
Comparing \model{} with the baselines described in \cref{sec:baselines}, we appreciate how \model{} achieves the best performance in most metrics, often by large margins against runner-ups. For instance, for the Sentence-BERT similarity on the ShapeNeRF--Text dataset, \model{} achieves $68.63$, $7.63$ points more than  \llava{}-vicuna13b, even if \model{} uses a smaller LLM. Results on the HST dataset, which provides ground-truth descriptions validated by humans, are generally lower for all methods. Yet, \model{} provides again the best performance according to most metrics. The difference in the quality of the brief description provided by \model{} compared to the baselines is showcased by the qualitative result reported in the first row of \cref{fig:qualitative_results}, where the description provided by \model{} is the most accurate. 
%Interestingly, the runner-up in this test is often BLIP-2, which was not so effective when compared to the machine-generated ground-truth. This suggests that the performance of LLaVA models can be slightly overestimated because we used the same model in our automatic annotation pipeline.

A clear trend in both tables and qualitative results is that image-based models tend to perform better than models processing point clouds. This is likely due to the larger amount of data used during training of the modality encoder, i.e. millions of images versus hundreds of thousands of shapes, which enhances their generalization ability, as well as the capability of images to capture more details than point clouds at the input resolutions required by image-based MLLMs versus 3D MLLMs. Nonetheless, our method, which operates on NeRFs, benefits from a holistic view of the object and provides the most accurate descriptions. Remarkably, in \model{}, all the necessary information for this language task can be extracted from a single global embedding obtained by directly processing the \nerf{} weights. 
It is also worth pointing out that,  while \model{} directly processes weights and thus is independent by design from spatial resolution, the baselines face a computational overhead growing with the desired resolution due to the necessity of extracting spatial data from NeRF (\cref{sec:extract_data_from_nerf}).  % especially in the case of point clouds 
% Additionally, given a NeRF, the non-trivial computational and memory overhead of rendering views or extracting point clouds from a NeRF. In contrast, our approach, which processes directly the weights, does not encounter this issue, as shown in the sup \textcolor{red}{TODOTODO}.
% 
Results show that 3D-LLM performs better than the point-based models and comparably to image-based models. Comparing the results of image-based MLLMs when processing front versus back views, we can see that the vantage point has a non-negligible effect on the performance of such baselines, with SentenceBERT and SimCSE metrics diminishing by about 4 points in all baselines. In a dataset without canonical poses for objects, this would be a relevant limitation that processing  NeRF weights seamlessly sidesteps. Finally, we observe that the multi-view setup of \llava{} provides similar performance to the single-view counterpart. In \cref{sec:qualitatives_supp}, additional qualitative examples are provided.

\paragraph{Detailed description.}
We evaluate the performance for the detailed description tasks on the proposed ShapeNeRF--Text, reporting the results in \cref{tab:detailed_frozen}.
For this task, the point-based model PointLLM~\cite{pointllm} performs similarly to the image-based one, \llava{}~\cite{llava}. However, we appreciate that \model{} achieves the best performance in all metrics by large margins.  For instance, for the Sentence-BERT metric, \model{} achieves $77.43$,  notably $18.35$ points more than \llava{}-vicuna-13b single-view and $17.22$ for the \llava{}-vicuna-13b multi-view setup. 
These large improvements indicate that, while individual or aggregated images may be sufficient for brief descriptions, they may lack all the details needed to provide a comprehensive description. Moreover, the dependency of the output quality on the selected vantage points remains strong. Contrarily, the NeRF weights contain detailed and complete information about the object, which is fundamental for more granular description tasks, with the additional advantage of not requiring tuning such hyperparameters. 
The ability of NeRF to capture holistic information about the object is also shown in the second row of \cref{fig:qualitative_results}, where only the direct processing of NeRF weights lets LLaNA understand that the object is a TV. PointLLM and \llava{} provide detailed but wrong descriptions, likely because of the need to extract the intermediate discrete representation as a point cloud or an image, losing information. Indeed, in both cases, it is hard even for a human observer to provide the right description from the intermediate modalities shown in the figure. More qualitative examples of this task are shown in \cref{sec:qualitatives_supp}.

\subsection{NeRF single-round Q\&A}

In the single-round Q\&A experiment, we test the ability of the assistants to provide precise answers to specific questions about the object instead of open-ended general descriptions.
% \begin{wrapfigure}{r}{5cm}
%     \includegraphics[width=\linewidth]{figures/qualitativi_multi_round.pdf}
%     \caption{\textbf{Example of multi-round NeRF Q\&A.}}
%     \label{fig:multi-round}
% \end{wrapfigure}
We prompt the models with the NeRF, or the image/cloud extracted from it, followed by one of the questions in the single-round Q\&A annotations associated with the NeRF. We then collect the answer generated by the model and compare it against the ground-truth answer with the selected metrics. Results are reported in \cref{tab:qa_frozen}.
Interestingly, PointLLM~\cite{pointllm} performs better than \llava{}~\cite{llava} in this task, likely because it has been specifically trained to answer detailed questions about objects represented as point clouds. %\textcolor{red}{DECIDERE SE TOGLIERE QUESTA CONSIDERAZIONE}
Nevertheless, similarly to the detailed description results, \model{} is the top-performing method across all metrics, again by large margins. This result suggests that the meta-encoder and the projector can extract fine-grained information from the NeRF, even if they are processing directly NeRF weights. Remarkably, the amount of information they can extract lets \model{} answer more precisely than when images or point clouds are extracted from the NeRF. Indeed, as shown in the third row of \cref{fig:qualitative_results} which reports a qualitative example, the only assistant able to answer correctly to a precise question about the appearance of the tyres of the car is \model{}. In \cref{sec:qualitatives_supp}, additional qualitative examples of this task are provided. Finally, another qualitative result confirming the ability of \model{} to provide high-quality answers to specific questions, in this case in a multi-round Q\&A experiment, is reported in \cref{fig:multi-round}.

\begin{table}
    \centering
    \begin{minipage}{0.48\linewidth}
    \caption{\textbf{NeRF brief captioning on ShapeNeRF-Text. Trained baselines.}\\Best results are in \textbf{bold}, runner-up is \underline{underlined}. \\(FV: front-view)}
    \resizebox{\linewidth}{!}{%
        \begin{tabular}{lcccccc}
        \midrule
        \textbf{Model} & \textbf{Modality} & \textbf{S-BERT} & \textbf{SimCSE} & \textbf{BLEU-1} & \textbf{ROUGE-L} & \textbf{METEOR} \\
        \cmidrule(lr){1-1} \cmidrule(lr){2-2} \cmidrule(lr){3-7}
        \llava{}-vicuna-13b & Image (FV) & 42.86 & 43.22 & 15.56 & 13.74 & 15.27 \\
        \cmidrule(lr){1-1} \cmidrule(lr){2-2} \cmidrule(lr){3-7}
        PointLLM-7b & Point cloud & \underline{55.48} & \underline{57.28} & \textbf{21.67} &   \underline{25.84} & 24.54 \\
        GPT4Point-Opt-2.7b & Point cloud & 37.96 & 39.00 & \underline{21.33} & 22.29 & \underline{24.88} \\
        \cmidrule(lr){1-1} \cmidrule(lr){2-2} \cmidrule(lr){3-7}
        \model{}-7b & \nerf{} & \textbf{68.63} & \textbf{70.54} & 20.64 & \textbf{28.33} & \textbf{31.76} \\
        \bottomrule
        \end{tabular}}
    \label{tab:brief_trained}
    \end{minipage}
    \hfill
    \begin{minipage}{0.48\linewidth}
    \caption{\textbf{NeRF brief captioning on the HST dataset. Trained baselines.}\\Best results are in \textbf{bold}, runner-up is \underline{underlined}. \\(FV: front-view)}
    \resizebox{\linewidth}{!}{%
        \begin{tabular}{lcccccc}
        \midrule
        \textbf{Model} & \textbf{Modality} & \textbf{S-BERT} & \textbf{SimCSE} & \textbf{BLEU-1} & \textbf{ROUGE-L} & \textbf{METEOR} \\
        \cmidrule(lr){1-1} \cmidrule(lr){2-2} \cmidrule(lr){3-7}
        \llava{}-vicuna-13b & Image (FV) & 33.79 & 42.66 & \textbf{10.28} & \textbf{13.22} & 12.19 \\
        \cmidrule(lr){1-1} \cmidrule(lr){2-2} \cmidrule(lr){3-7}
        PointLLM-7b & Point cloud & \underline{44.65} & \underline{44.68} & \underline{8.91} & 12.33 &  \underline{12.64} \\
        GPT4Point-Opt-2.7B & Point cloud & 30.50 & 31.08 & 8.12 & \underline{12.35} & 11.62 \\
        \cmidrule(lr){1-1} \cmidrule(lr){2-2} \cmidrule(lr){3-7}
        \model{}-7b & \nerf{} & \textbf{55.62} & \textbf{55.56} & 6.56 & 11.81 & \textbf{14.52}  \\
        \bottomrule
        \end{tabular}}
    \label{tab:brief_trained_hst}
    \end{minipage}
\end{table}
%%% AGGREGATED FROZEN AND TRAINED TABLES

\begin{table}[t]

    \centering
    \begin{minipage}{0.48\linewidth}
    \caption{\textbf{NeRF detailed captioning on ShapeNeRF-Text. Trained baselines.}\\Best results are in \textbf{bold}, runner-up is \underline{underlined}. \\(FV: front-view)}\textbf{}
    \resizebox{\linewidth}{!}{
        \begin{tabular}{lcccccc}
        \midrule
        \textbf{Model} & \textbf{Modality} & \textbf{S-BERT} & \textbf{SimCSE} & \textbf{BLEU-1} & \textbf{ROUGE-L} & \textbf{METEOR} \\
        \cmidrule(lr){1-1} \cmidrule(lr){2-2} \cmidrule(lr){3-7}
        \llava{}-vicuna-13b & Image (FV) & 44.69 & 42.31 & 10.08 & \underline{23.46} & 12.70 \\
        \cmidrule(lr){1-1} \cmidrule(lr){2-2} \cmidrule(lr){3-7}
        PointLLM-7b & Point cloud & \underline{67.30} & \underline{59.56} & \underline{15.39} & 21.42 & 11.37 \\
        GPT4Point-Opt-2.7b & Point cloud & 41.33 & 40.52 & 14.48 & 19.15 & \underline{13.80} \\
        \cmidrule(lr){1-1} \cmidrule(lr){2-2} \cmidrule(lr){3-7}
        \model{}-7b & \nerf{} & \textbf{77.43} & \textbf{79.81} & \textbf{41.32} & \textbf{36.18} & \textbf{32.39} \\
        \bottomrule
        \end{tabular}}
    \label{tab:detailed_trained}
    \end{minipage}
    \hfill
    \begin{minipage}{0.48\linewidth}
    \caption{\textbf{NeRF single-round Q\&A on ShapeNeRF-Text. Trained baselines.}\\Best results are in \textbf{bold}, runner-up is \underline{underlined}. \\(FV: front-view)}
    \resizebox{\linewidth}{!}{%
        \begin{tabular}{lcccccc}
        \midrule
        \textbf{Model} & \textbf{Modality} & \textbf{S-BERT} & \textbf{SimCSE} & \textbf{BLEU-1} & \textbf{ROUGE-L} & \textbf{METEOR} \\
        \cmidrule(lr){1-1} \cmidrule(lr){2-2} \cmidrule(lr){3-7}
        \llava{}-vicuna-13b & Image (FV) & 56.29 & 62.36 & 26.87 & 29.55 & 30.49 \\
        \cmidrule(lr){1-1} \cmidrule(lr){2-2} \cmidrule(lr){3-7}
        PointLLM-7b & Point cloud & \underline{79.24} & \underline{80.38} & \underline{46.00} & \underline{52.60} & \underline{42.36} \\
        GPT4Point-Opt-2.7b & Point cloud & 22.22 & 28.66 & 8.76 & 13.46 & 14.19 \\
        \cmidrule(lr){1-1} \cmidrule(lr){2-2} \cmidrule(lr){3-7}
        \model{}-7b & \nerf{} & \textbf{81.03} & \textbf{81.56} & \textbf{46.16} & \textbf{53.17} & \textbf{50.15} \\
        \bottomrule
        \end{tabular}}
    \label{tab:qa_trained}
    \end{minipage}
\end{table}

% \begin{wraptable}{r}{5.5cm}
% \centering
% \resizebox{\linewidth}{!}{%
% \begin{tabular}{lcc}
% \toprule
% \textbf{Model} & \textbf{Modality} & \textbf{Accuracy (\%)} \\
% \cmidrule(lr){1-1} \cmidrule(lr){2-2} \cmidrule(lr){3-3}
% LLaVA-vicuna-13b& Image (Front View) & 66.13 \\
% LLaVA-vicuna-13b& Image (Hard View)    & 63.90 \\
% LLaVA-vicuna-7b& Image (Front View)  & 60.25 \\
% LLaVA-vicuna-7b& Image (Hard View)  & 57.00 \\
% BLIP-2 FlanT5-xxl& Image (Front View) & 63.67 \\
% BLIP-2 FlanT5-xxl& Image (Hard View)   & 61.47 \\
% \cmidrule(lr){1-1} \cmidrule(lr){2-2} \cmidrule(lr){3-3}
% PointLLM-7b & Point cloud                      & 50.14 \\
% GPT4Point-Opt-2.7b & Point cloud                        & 41.93     \\
% \cmidrule(lr){1-1} \cmidrule(lr){2-2} \cmidrule(lr){3-3}
% \model{}-7b & \nerf{}                      & \textbf{67.14} \\
% \bottomrule
% \end{tabular}
% }
% \caption{Zero-shot classification accuracy of different models.}
% \label{tab:zero_shot_classification_accuracy}
% \end{wraptable}
\subsection{Zero-shot NeRF classification}
Finally, we compare assistants on the task of zero-shot classification.  We query the models with the sentence \emph{"What is the class of the NeRF/image/cloud? Choose among these: <Shapenet\_classes>"} where \emph{<Shapenet\_classes>} are the 10 ShapeNet classes available in our dataset. We consider the answer correct only if the ground truth class appears in the response.
We report results in \cref{tab:classification_frozen} on the ShapeNeRF--Text dataset.
%
%Notably, \model{} achieves the best performance, demonstrating the effectiveness of the proposed architecture in realizing a NeRF assistant.
Using multiple views boosts the zero-shot classification performance of \llava{}, which turns out to be the best model for this task, followed by \model{}.

\subsection{Training baselines on ShapeNeRF--Text}
\cref{tab:brief_trained,tab:brief_trained_hst,tab:detailed_trained,tab:qa_trained} report results on language tasks of several baselines trained on ShapeNeRF--Text, while  \cref{tab:classification_trained} of the appendix, shows zero-shot NeRF classification performance of such models. 
We employed those baselines on ShapeNeRF--Text, for which we were able to run the official training code. Accordingly, we followed their protocol, which, for all of them, keeps the modality-specific encoder frozen and trains an adaptor and the LLM in two steps. We notice that the trained baselines exhibit different behaviors to their frozen counterparts, with \llava{} performing significantly worse and PointLLM showing clear improvements.  As for  GPT4Point, we observe greater variability across metrics; however, overall, it shows no significant benefit from training on ShapeNeRF–Text. \model{} yields the best performance compared to all baselines, either frozen or trained on ShapeNeRF--Text.
Finally, \cref{sec:generalize_supp} shows the generalization performance on Objaverse of \model{} and the trained baselines.
\section{Limitations and future directions}
\label{sec:limitations}
Despite the promising results of our framework, it is the first study in this direction and several limitations are yet to be addressed. First, the pre-trained \nftovec{} encoder, having been trained exclusively on synthetic data from ShapeNet, may not generalize well to real-world objects. To address this, future work should create a NeRF--Text dataset including a more diverse set of objects, like the ones provided by Objaverse~\cite{deitke2023objaverse} and OmniObject3D~\cite{wu2023omniobject3d}.
Another limitation is that \nftovec{} currently processes only MLPs, restricting our model to MLP-only \nerf{}s. However, with the rapid advancements in meta-networks, it may become very soon possible to extend \model{} to more complex \nerf{} architectures, such as InstantNGP~\cite{instant}. For instance, the approach by \citet{lim2024graph} suggests the feasibility of processing various input architectures, although it is currently limited to small networks. Finally, our framework has been tested solely on object-centric \nerf{}s. Expanding its application to \nerf{}s representing entire scenes would be a compelling direction for future research.

\section{Concluding remarks}
\label{sec:remarks}
This paper addressed the novel task of creating a language assistant for NeRF. We have tackled this problem by leveraging recent advances in MLLMs and meta-networks processing neural fields.  We have shown that it is feasible and effective to directly process the weights of a NeRF to project it into the input embedding space of an LLM. We have built and made publicly available a dataset of textual annotations of NeRFs and have shown that our approach compares favourably with respect to several MLLMs used as baselines for the novel tasks of brief and detailed captioning, question answering, and zero-shot classification of NeRFs.

%\paragraph{Broader impacts}
%Bla bla bla
%\todo{completa questa parte}

% \section{Acknowledgement}
% We acknowledge the CINECA award under the ISCRA initiative, for the availability of high-performance computing resources and support.
\section*{Acknowledgements}
We acknowledge the CINECA award under the ISCRA initiative, for the availability of high-performance computing resources and support.

{
\small
\bibliographystyle{ieeenat_fullname}
\bibliography{main}

\begin{thebibliography}{87}
\providecommand{\natexlab}[1]{#1}
\providecommand{\url}[1]{\texttt{#1}}
\expandafter\ifx\csname urlstyle\endcsname\relax
  \providecommand{\doi}[1]{doi: #1}\else
  \providecommand{\doi}{doi: \begingroup \urlstyle{rm}\Url}\fi

\bibitem[Achiam et~al.(2023)Achiam, Adler, Agarwal, Ahmad, Akkaya, Aleman, Almeida, Altenschmidt, Altman, Anadkat, et~al.]{GPT4}
Josh Achiam, Steven Adler, Sandhini Agarwal, Lama Ahmad, Ilge Akkaya, Florencia~Leoni Aleman, Diogo Almeida, Janko Altenschmidt, Sam Altman, Shyamal Anadkat, et~al.
\newblock Gpt-4 technical report.
\newblock \emph{arXiv preprint arXiv:2303.08774}, 2023.

\bibitem[Amaduzzi et~al.(2023)Amaduzzi, Lisanti, Salti, and Di~Stefano]{amaduzzi2023looking}
Andrea Amaduzzi, Giuseppe Lisanti, Samuele Salti, and Luigi Di~Stefano.
\newblock Looking at words and points with attention: a benchmark for text-to-shape coherence.
\newblock In \emph{2023 IEEE/CVF International Conference on Computer Vision Workshops (ICCVW)}, pages 2860--2869. IEEE Computer Society, 2023.

\bibitem[Bai et~al.(2023{\natexlab{a}})Bai, Lyu, Jiang, Li, Lu, Lin, and Wang]{bai2023componerf}
Haotian Bai, Yuanhuiyi Lyu, Lutao Jiang, Sijia Li, Haonan Lu, Xiaodong Lin, and Lin Wang.
\newblock Componerf: Text-guided multi-object compositional nerf with editable 3d scene layout.
\newblock \emph{arXiv preprint arXiv:2303.13843}, 2023{\natexlab{a}}.

\bibitem[Bai et~al.(2023{\natexlab{b}})Bai, Bai, Yang, Wang, Tan, Wang, Lin, Zhou, and Zhou]{Qwen-VL}
Jinze Bai, Shuai Bai, Shusheng Yang, Shijie Wang, Sinan Tan, Peng Wang, Junyang Lin, Chang Zhou, and Jingren Zhou.
\newblock Qwen-vl: A frontier large vision-language model with versatile abilities.
\newblock \emph{arXiv preprint arXiv:2308.12966}, 2023{\natexlab{b}}.

\bibitem[Ballerini et~al.(2024)Ballerini, Zama~Ramirez, Mirabella, Salti, and Di~Stefano]{ballerini2024clip2nerf}
Francesco Ballerini, Pierluigi Zama~Ramirez, Roberto Mirabella, Samuele Salti, and Luigi Di~Stefano.
\newblock Connecting nerfs, images, and text.
\newblock In \emph{Proceedings of the IEEE/CVF Conference on Computer Vision and Pattern Recognition (CVPR) Workshops}, 2024.

\bibitem[Banerjee and Lavie(2005)]{meteor}
Satanjeev Banerjee and Alon Lavie.
\newblock Meteor: An automatic metric for mt evaluation with improved correlation with human judgments.
\newblock In \emph{Proceedings of the acl workshop on intrinsic and extrinsic evaluation measures for machine translation and/or summarization}, pages 65--72, 2005.

\bibitem[Cardace et~al.(2024)Cardace, Ramirez, Ballerini, Zhou, Salti, and di~Stefano]{cardace2024neural}
Adriano Cardace, Pierluigi~Zama Ramirez, Francesco Ballerini, Allan Zhou, Samuele Salti, and Luigi di Stefano.
\newblock Neural processing of tri-plane hybrid neural fields.
\newblock In \emph{The Twelfth International Conference on Learning Representations}, 2024.

\bibitem[Chang et~al.(2015)Chang, Funkhouser, Guibas, Hanrahan, Huang, Li, Savarese, Savva, Song, Su, et~al.]{chang2015shapenet}
Angel~X Chang, Thomas Funkhouser, Leonidas Guibas, Pat Hanrahan, Qixing Huang, Zimo Li, Silvio Savarese, Manolis Savva, Shuran Song, Hao Su, et~al.
\newblock Shapenet: An information-rich 3d model repository.
\newblock \emph{arXiv preprint arXiv:1512.03012}, 2015.

\bibitem[Chen et~al.(2022)Chen, Xu, Geiger, Yu, and Su]{Chen2022ECCV}
Anpei Chen, Zexiang Xu, Andreas Geiger, Jingyi Yu, and Hao Su.
\newblock Tensorf: Tensorial radiance fields.
\newblock In \emph{European Conference on Computer Vision (ECCV)}, 2022.

\bibitem[Chen et~al.(2023)Chen, Zheng, Wang, Xu, Huang, Pan, Wang, Wang, Qiao, Lu, et~al.]{VideoLLM}
Guo Chen, Yin-Dong Zheng, Jiahao Wang, Jilan Xu, Yifei Huang, Junting Pan, Yi Wang, Yali Wang, Yu Qiao, Tong Lu, et~al.
\newblock Videollm: Modeling video sequence with large language models.
\newblock \emph{arXiv preprint arXiv:2305.13292}, 2023.

\bibitem[Dai et~al.(2024)Dai, Li, Li, Tiong, Zhao, Wang, Li, Fung, and Hoi]{InstructBLIP}
Wenliang Dai, Junnan Li, Dongxu Li, Anthony Meng~Huat Tiong, Junqi Zhao, Weisheng Wang, Boyang Li, Pascale~N Fung, and Steven Hoi.
\newblock Instructblip: Towards general-purpose vision-language models with instruction tuning.
\newblock \emph{Advances in Neural Information Processing Systems}, 36, 2024.

\bibitem[De~Luigi et~al.(2023)De~Luigi, Cardace, Spezialetti, Zama~Ramirez, Salti, and Di~Stefano]{deluigi2023inr2vec}
Luca De~Luigi, Adriano Cardace, Riccardo Spezialetti, Pierluigi Zama~Ramirez, Samuele Salti, and Luigi Di~Stefano.
\newblock Deep learning on implicit neural representations of shapes.
\newblock In \emph{International Conference on Learning Representations (ICLR)}, 2023.

\bibitem[Deitke et~al.(2023)Deitke, Schwenk, Salvador, Weihs, Michel, VanderBilt, Schmidt, Ehsani, Kembhavi, and Farhadi]{deitke2023objaverse}
Matt Deitke, Dustin Schwenk, Jordi Salvador, Luca Weihs, Oscar Michel, Eli VanderBilt, Ludwig Schmidt, Kiana Ehsani, Aniruddha Kembhavi, and Ali Farhadi.
\newblock Objaverse: A universe of annotated 3d objects.
\newblock In \emph{Proceedings of the IEEE/CVF Conference on Computer Vision and Pattern Recognition}, pages 13142--13153, 2023.

\bibitem[Driess et~al.(2023)Driess, Xia, Sajjadi, Lynch, Chowdhery, Ichter, Wahid, Tompson, Vuong, Yu, et~al.]{Palm-E}
Danny Driess, Fei Xia, Mehdi~SM Sajjadi, Corey Lynch, Aakanksha Chowdhery, Brian Ichter, Ayzaan Wahid, Jonathan Tompson, Quan Vuong, Tianhe Yu, et~al.
\newblock Palm-e: An embodied multimodal language model.
\newblock In \emph{International Conference on Machine Learning}, pages 8469--8488. PMLR, 2023.

\bibitem[Dupont et~al.(2022)Dupont, Kim, Eslami, Rezende, and Rosenbaum]{functa}
Emilien Dupont, Hyunjik Kim, SM~Ali Eslami, Danilo~Jimenez Rezende, and Dan Rosenbaum.
\newblock From data to functa: Your data point is a function and you can treat it like one.
\newblock In \emph{International Conference on Machine Learning}, pages 5694--5725. PMLR, 2022.

\bibitem[Fridovich-Keil et~al.(2022)Fridovich-Keil, Yu, Tancik, Chen, Recht, and Kanazawa]{Plenoxels}
Sara Fridovich-Keil, Alex Yu, Matthew Tancik, Qinhong Chen, Benjamin Recht, and Angjoo Kanazawa.
\newblock Plenoxels: Radiance fields without neural networks.
\newblock \emph{2022 IEEE/CVF Conference on Computer Vision and Pattern Recognition (CVPR)}, 2022.

\bibitem[Gao et~al.(2023)Gao, Han, Zhang, Lin, Geng, Zhou, Zhang, Lu, He, Yue, et~al.]{LLaMA-Adapterv2}
Peng Gao, Jiaming Han, Renrui Zhang, Ziyi Lin, Shijie Geng, Aojun Zhou, Wei Zhang, Pan Lu, Conghui He, Xiangyu Yue, et~al.
\newblock Llama-adapter v2: Parameter-efficient visual instruction model.
\newblock \emph{arXiv preprint arXiv:2304.15010}, 2023.

\bibitem[Gao et~al.(2021)Gao, Yao, and Chen]{simcse}
Tianyu Gao, Xingcheng Yao, and Danqi Chen.
\newblock Simcse: Simple contrastive learning of sentence embeddings.
\newblock In \emph{2021 Conference on Empirical Methods in Natural Language Processing, EMNLP 2021}, pages 6894--6910. Association for Computational Linguistics (ACL), 2021.

\bibitem[Girdhar et~al.(2023)Girdhar, El-Nouby, Liu, Singh, Alwala, Joulin, and Misra]{image-bind}
Rohit Girdhar, Alaaeldin El-Nouby, Zhuang Liu, Mannat Singh, Kalyan~Vasudev Alwala, Armand Joulin, and Ishan Misra.
\newblock Imagebind: One embedding space to bind them all.
\newblock In \emph{Proceedings of the IEEE/CVF Conference on Computer Vision and Pattern Recognition}, pages 15180--15190, 2023.

\bibitem[Guo et~al.(2023)Guo, Zhang, Zhu, Tang, Ma, Han, Chen, Gao, Li, Li, et~al.]{Point-bind}
Ziyu Guo, Renrui Zhang, Xiangyang Zhu, Yiwen Tang, Xianzheng Ma, Jiaming Han, Kexin Chen, Peng Gao, Xianzhi Li, Hongsheng Li, et~al.
\newblock Point-bind \& point-llm: Aligning point cloud with multi-modality for 3d understanding, generation, and instruction following.
\newblock \emph{arXiv preprint arXiv:2309.00615}, 2023.

\bibitem[Haque et~al.(2023)Haque, Tancik, Efros, Holynski, and Kanazawa]{Haque_2023_ICCV}
Ayaan Haque, Matthew Tancik, Alexei~A. Efros, Aleksander Holynski, and Angjoo Kanazawa.
\newblock Instruct-nerf2nerf: Editing 3d scenes with instructions.
\newblock In \emph{Proceedings of the IEEE/CVF International Conference on Computer Vision (ICCV)}, pages 19740--19750, 2023.

\bibitem[Hecht-Nielsen(1990)]{hecht1990algebraic}
Robert Hecht-Nielsen.
\newblock On the algebraic structure of feedforward network weight spaces.
\newblock In \emph{Advanced Neural Computers}, pages 129--135. Elsevier, 1990.

\bibitem[Hong et~al.(2023{\natexlab{a}})Hong, Lin, Du, Chen, Tenenbaum, and Gan]{Hong_2023_CVPR}
Yining Hong, Chunru Lin, Yilun Du, Zhenfang Chen, Joshua~B. Tenenbaum, and Chuang Gan.
\newblock 3d concept learning and reasoning from multi-view images.
\newblock In \emph{Proceedings of the IEEE/CVF Conference on Computer Vision and Pattern Recognition (CVPR)}, pages 9202--9212, 2023{\natexlab{a}}.

\bibitem[Hong et~al.(2023{\natexlab{b}})Hong, Zhen, Chen, Zheng, Du, Chen, and Gan]{3dllm}
Yining Hong, Haoyu Zhen, Peihao Chen, Shuhong Zheng, Yilun Du, Zhenfang Chen, and Chuang Gan.
\newblock 3d-{LLM}: Injecting the 3d world into large language models.
\newblock In \emph{Thirty-seventh Conference on Neural Information Processing Systems}, 2023{\natexlab{b}}.

\bibitem[Hu et~al.(2023)Hu, Huang, Liu, Tai, and Tang]{hu2023nerf}
Benran Hu, Junkai Huang, Yichen Liu, Yu-Wing Tai, and Chi-Keung Tang.
\newblock Nerf-rpn: A general framework for object detection in nerfs.
\newblock In \emph{Proceedings of the IEEE/CVF Conference on Computer Vision and Pattern Recognition}, pages 23528--23538, 2023.

\bibitem[Huang et~al.(2024{\natexlab{a}})Huang, Li, Yang, Shi, Chang, Ye, Wu, Hong, Huang, Liu, et~al.]{Audiogpt}
Rongjie Huang, Mingze Li, Dongchao Yang, Jiatong Shi, Xuankai Chang, Zhenhui Ye, Yuning Wu, Zhiqing Hong, Jiawei Huang, Jinglin Liu, et~al.
\newblock Audiogpt: Understanding and generating speech, music, sound, and talking head.
\newblock In \emph{Proceedings of the AAAI Conference on Artificial Intelligence}, pages 23802--23804, 2024{\natexlab{a}}.

\bibitem[Huang et~al.(2024{\natexlab{b}})Huang, Dong, Wang, Hao, Singhal, Ma, Lv, Cui, Mohammed, Patra, et~al.]{Kosmos-1}
Shaohan Huang, Li Dong, Wenhui Wang, Yaru Hao, Saksham Singhal, Shuming Ma, Tengchao Lv, Lei Cui, Owais~Khan Mohammed, Barun Patra, et~al.
\newblock Language is not all you need: Aligning perception with language models.
\newblock \emph{Advances in Neural Information Processing Systems}, 36, 2024{\natexlab{b}}.

\bibitem[Hwang et~al.(2023)Hwang, Hyung, Kim, Kim, and Choo]{Hwang_2023_ICCV}
Sungwon Hwang, Junha Hyung, Daejin Kim, Min-Jung Kim, and Jaegul Choo.
\newblock Faceclipnerf: Text-driven 3d face manipulation using deformable neural radiance fields.
\newblock In \emph{Proceedings of the IEEE/CVF International Conference on Computer Vision (ICCV)}, pages 3469--3479, 2023.

\bibitem[Ioffe and Szegedy(2015)]{ioffe2015batch}
Sergey Ioffe and Christian Szegedy.
\newblock Batch normalization: Accelerating deep network training by reducing internal covariate shift.
\newblock In \emph{International conference on machine learning}, pages 448--456. pmlr, 2015.

\bibitem[Jaeckle and Kumar(2021)]{jaeckle2021generating}
Florian Jaeckle and M~Pawan Kumar.
\newblock Generating adversarial examples with graph neural networks.
\newblock In \emph{Uncertainty in Artificial Intelligence}, pages 1556--1564. PMLR, 2021.

\bibitem[Jo et~al.(2023)Jo, Shim, Jung, Yang, and Choo]{Jo_2023_WACV}
Kyungmin Jo, Gyumin Shim, Sanghun Jung, Soyoung Yang, and Jaegul Choo.
\newblock Cg-nerf: Conditional generative neural radiance fields for 3d-aware image synthesis.
\newblock In \emph{Proceedings of the IEEE/CVF Winter Conference on Applications of Computer Vision (WACV)}, pages 724--733, 2023.

\bibitem[Kerr et~al.(2023)Kerr, Kim, Goldberg, Kanazawa, and Tancik]{lerf2023}
Justin* Kerr, Chung~Min* Kim, Ken Goldberg, Angjoo Kanazawa, and Matthew Tancik.
\newblock Lerf: Language embedded radiance fields.
\newblock In \emph{International Conference on Computer Vision (ICCV)}, 2023.

\bibitem[Knyazev et~al.(2021)Knyazev, Drozdzal, Taylor, and Romero]{knyazev2021parameter}
Boris Knyazev, Michal Drozdzal, Graham~W. Taylor, and Adriana Romero.
\newblock Parameter prediction for unseen deep architectures.
\newblock In \emph{Advances in Neural Information Processing Systems}, 2021.

\bibitem[Kobayashi et~al.(2022)Kobayashi, Matsumoto, and Sitzmann]{NEURIPS2022_93f25021}
Sosuke Kobayashi, Eiichi Matsumoto, and Vincent Sitzmann.
\newblock Decomposing nerf for editing via feature field distillation.
\newblock In \emph{Advances in Neural Information Processing Systems}, pages 23311--23330. Curran Associates, Inc., 2022.

\bibitem[Kofinas et~al.(2024)Kofinas, Knyazev, Zhang, Chen, Burghouts, Gavves, Snoek, and Zhang]{kofinas2024graph}
Miltiadis Kofinas, Boris Knyazev, Yan Zhang, Yunlu Chen, Gertjan~J Burghouts, Efstratios Gavves, Cees~GM Snoek, and David~W Zhang.
\newblock Graph neural networks for learning equivariant representations of neural networks.
\newblock In \emph{The Twelfth International Conference on Learning Representations}, 2024.

\bibitem[Lee and Chang(2022)]{lee2022understanding}
Han-Hung Lee and Angel~X. Chang.
\newblock Understanding pure clip guidance for voxel grid nerf models, 2022.

\bibitem[Li et~al.(2023{\natexlab{a}})Li, Zhang, Chen, Wang, Pu, Yang, Li, and Liu]{Otter}
Bo Li, Yuanhan Zhang, Liangyu Chen, Jinghao Wang, Fanyi Pu, Jingkang Yang, Chunyuan Li, and Ziwei Liu.
\newblock Mimic-it: Multi-modal in-context instruction tuning.
\newblock \emph{arXiv preprint arXiv:2306.05425}, 2023{\natexlab{a}}.

\bibitem[Li et~al.(2023{\natexlab{b}})Li, Li, Savarese, and Hoi]{BLIP-2}
Junnan Li, Dongxu Li, Silvio Savarese, and Steven Hoi.
\newblock Blip-2: Bootstrapping language-image pre-training with frozen image encoders and large language models.
\newblock In \emph{International conference on machine learning}, pages 19730--19742. PMLR, 2023{\natexlab{b}}.

\bibitem[Li et~al.(2023{\natexlab{c}})Li, Li, Savarese, and Hoi]{blip2}
Junnan Li, Dongxu Li, Silvio Savarese, and Steven Hoi.
\newblock {BLIP-2}: Bootstrapping language-image pre-training with frozen image encoders and large language models.
\newblock \emph{arXiv preprint arXiv:2301.12597}, 2023{\natexlab{c}}.

\bibitem[Li et~al.(2024)Li, Liu, Liu, Wang, Zheng, Xu, Li, and Zhu]{li2024instructpixnerf}
Jianhui Li, Shilong Liu, Zidong Liu, Yikai Wang, Kaiwen Zheng, Jinghui Xu, Jianmin Li, and Jun Zhu.
\newblock Instructpix2ne{RF}: Instructed 3d portrait editing from a single image.
\newblock In \emph{The Twelfth International Conference on Learning Representations}, 2024.

\bibitem[Li et~al.(2023{\natexlab{d}})Li, Gao, Tancik, and Kanazawa]{nerfacc}
Ruilong Li, Hang Gao, Matthew Tancik, and Angjoo Kanazawa.
\newblock Nerfacc: Efficient sampling accelerates nerfs.
\newblock \emph{arXiv preprint arXiv:2305.04966}, 2023{\natexlab{d}}.

\bibitem[Lim et~al.(2024)Lim, Maron, Law, Lorraine, and Lucas]{lim2024graph}
Derek Lim, Haggai Maron, Marc~T. Law, Jonathan Lorraine, and James Lucas.
\newblock Graph metanetworks for processing diverse neural architectures.
\newblock In \emph{The Twelfth International Conference on Learning Representations}, 2024.

\bibitem[Lin(2004)]{rouge}
Chin-Yew Lin.
\newblock Rouge: A package for automatic evaluation of summaries.
\newblock In \emph{Text summarization branches out}, pages 74--81, 2004.

\bibitem[Liu et~al.(2024)Liu, Li, Wu, and Lee]{llava}
Haotian Liu, Chunyuan Li, Qingyang Wu, and Yong~Jae Lee.
\newblock Visual instruction tuning.
\newblock \emph{Advances in neural information processing systems}, 36, 2024.

\bibitem[Lu and Kumar(2020)]{Lu2020Neural}
Jingyue Lu and M.~Pawan Kumar.
\newblock Neural network branching for neural network verification.
\newblock In \emph{International Conference on Learning Representations}, 2020.

\bibitem[Luo et~al.(2023)Luo, Rockwell, Lee, and Johnson]{cap3d}
Tiange Luo, Chris Rockwell, Honglak Lee, and Justin Johnson.
\newblock Scalable 3d captioning with pretrained models.
\newblock \emph{arXiv preprint arXiv:2306.07279}, 2023.

\bibitem[Maaz et~al.(2023)Maaz, Rasheed, Khan, and Khan]{VideoChatGPT}
Muhammad Maaz, Hanoona Rasheed, Salman Khan, and Fahad~Shahbaz Khan.
\newblock Video-chatgpt: Towards detailed video understanding via large vision and language models.
\newblock \emph{arXiv preprint arXiv:2306.05424}, 2023.

\bibitem[Martin-Brualla et~al.(2021)Martin-Brualla, Radwan, Sajjadi, Barron, Dosovitskiy, and Duckworth]{martin2021nerf}
Ricardo Martin-Brualla, Noha Radwan, Mehdi~SM Sajjadi, Jonathan~T Barron, Alexey Dosovitskiy, and Daniel Duckworth.
\newblock Nerf in the wild: Neural radiance fields for unconstrained photo collections.
\newblock In \emph{Proceedings of the IEEE/CVF Conference on Computer Vision and Pattern Recognition}, pages 7210--7219, 2021.

\bibitem[Metzer et~al.(2023)Metzer, Richardson, Patashnik, Giryes, and Cohen-Or]{Metzer_2023_CVPR}
Gal Metzer, Elad Richardson, Or Patashnik, Raja Giryes, and Daniel Cohen-Or.
\newblock Latent-nerf for shape-guided generation of 3d shapes and textures.
\newblock In \emph{Proceedings of the IEEE/CVF Conference on Computer Vision and Pattern Recognition (CVPR)}, pages 12663--12673, 2023.

\bibitem[Mildenhall et~al.(2020)Mildenhall, Srinivasan, Tancik, Barron, Ramamoorthi, and Ng]{nerf}
Ben Mildenhall, Pratul~P Srinivasan, Matthew Tancik, Jonathan~T Barron, Ravi Ramamoorthi, and Ren Ng.
\newblock Nerf: Representing scenes as neural radiance fields for view synthesis.
\newblock In \emph{European conference on computer vision}, pages 405--421. Springer, 2020.

\bibitem[Mirzaei et~al.(2023)Mirzaei, Aumentado-Armstrong, Brubaker, Kelly, Levinshtein, Derpanis, and Gilitschenski]{Mirzaei_2023_ICCV}
Ashkan Mirzaei, Tristan Aumentado-Armstrong, Marcus~A. Brubaker, Jonathan Kelly, Alex Levinshtein, Konstantinos~G. Derpanis, and Igor Gilitschenski.
\newblock Reference-guided controllable inpainting of neural radiance fields.
\newblock In \emph{Proceedings of the IEEE/CVF International Conference on Computer Vision (ICCV)}, pages 17815--17825, 2023.

\bibitem[M\"uller et~al.(2022)M\"uller, Evans, Schied, and Keller]{instant}
Thomas M\"uller, Alex Evans, Christoph Schied, and Alexander Keller.
\newblock Instant neural graphics primitives with a multiresolution hash encoding.
\newblock \emph{ACM Trans. Graph.}, 41\penalty0 (4):\penalty0 102:1--102:15, 2022.

\bibitem[Navon et~al.(2023)Navon, Shamsian, Achituve, Fetaya, Chechik, and Maron]{navon2023equivariant}
Aviv Navon, Aviv Shamsian, Idan Achituve, Ethan Fetaya, Gal Chechik, and Haggai Maron.
\newblock Equivariant architectures for learning in deep weight spaces.
\newblock In \emph{International Conference on Machine Learning}, 2023.

\bibitem[Ouyang et~al.(2022)Ouyang, Wu, Jiang, Almeida, Wainwright, Mishkin, Zhang, Agarwal, Slama, Ray, et~al.]{InstructGPT}
Long Ouyang, Jeffrey Wu, Xu Jiang, Diogo Almeida, Carroll Wainwright, Pamela Mishkin, Chong Zhang, Sandhini Agarwal, Katarina Slama, Alex Ray, et~al.
\newblock Training language models to follow instructions with human feedback.
\newblock \emph{Advances in neural information processing systems}, 35:\penalty0 27730--27744, 2022.

\bibitem[Papineni et~al.(2002)Papineni, Roukos, Ward, and Zhu]{bleu}
Kishore Papineni, Salim Roukos, Todd Ward, and Wei-Jing Zhu.
\newblock Bleu: a method for automatic evaluation of machine translation.
\newblock In \emph{Proceedings of the 40th annual meeting of the Association for Computational Linguistics}, pages 311--318, 2002.

\bibitem[Peng et~al.(2023)Peng, Wang, Dong, Hao, Huang, Ma, and Wei]{Kosmos-2}
Zhiliang Peng, Wenhui Wang, Li Dong, Yaru Hao, Shaohan Huang, Shuming Ma, and Furu Wei.
\newblock Kosmos-2: Grounding multimodal large language models to the world.
\newblock \emph{arXiv preprint arXiv:2306.14824}, 2023.

\bibitem[Poole et~al.(2022)Poole, Jain, Barron, and Mildenhall]{poole2022dreamfusion}
Ben Poole, Ajay Jain, Jonathan~T Barron, and Ben Mildenhall.
\newblock Dreamfusion: Text-to-3d using 2d diffusion.
\newblock In \emph{The Eleventh International Conference on Learning Representations}, 2022.

\bibitem[Qi et~al.(2024)Qi, Fang, Sun, Wu, Wu, Wang, Lin, and Zhao]{gpt4point}
Zhangyang Qi, Ye Fang, Zeyi Sun, Xiaoyang Wu, Tong Wu, Jiaqi Wang, Dahua Lin, and Hengshuang Zhao.
\newblock Gpt4point: A unified framework for point-language understanding and generation.
\newblock In \emph{CVPR}, 2024.

\bibitem[Radford et~al.(2021)Radford, Kim, Hallacy, Ramesh, Goh, Agarwal, Sastry, Askell, Mishkin, Clark, et~al.]{clip}
Alec Radford, Jong~Wook Kim, Chris Hallacy, Aditya Ramesh, Gabriel Goh, Sandhini Agarwal, Girish Sastry, Amanda Askell, Pamela Mishkin, Jack Clark, et~al.
\newblock Learning transferable visual models from natural language supervision.
\newblock In \emph{International Conference on Machine Learning}, pages 8748--8763. PMLR, 2021.

\bibitem[Raffel et~al.(2020)Raffel, Shazeer, Roberts, Lee, Narang, Matena, Zhou, Li, and Liu]{t5}
Colin Raffel, Noam Shazeer, Adam Roberts, Katherine Lee, Sharan Narang, Michael Matena, Yanqi Zhou, Wei Li, and Peter~J Liu.
\newblock Exploring the limits of transfer learning with a unified text-to-text transformer.
\newblock \emph{Journal of machine learning research}, 21\penalty0 (140):\penalty0 1--67, 2020.

\bibitem[Ramirez et~al.(2023)Ramirez, De~Luigi, Sirocchi, Cardace, Spezialetti, Ballerini, Salti, and Di~Stefano]{ramirez2023deep}
Pierluigi~Zama Ramirez, Luca De~Luigi, Daniele Sirocchi, Adriano Cardace, Riccardo Spezialetti, Francesco Ballerini, Samuele Salti, and Luigi Di~Stefano.
\newblock Deep learning on 3d neural fields.
\newblock \emph{arXiv preprint arXiv:2312.13277}, 2023.

\bibitem[Ray(2023)]{ChatGPT}
Partha~Pratim Ray.
\newblock Chatgpt: A comprehensive review on background, applications, key challenges, bias, ethics, limitations and future scope.
\newblock \emph{Internet of Things and Cyber-Physical Systems}, 3:\penalty0 121--154, 2023.

\bibitem[Reimers and Gurevych(2019)]{sentencebert}
Nils Reimers and Iryna Gurevych.
\newblock Sentence-bert: Sentence embeddings using siamese bert-networks.
\newblock \emph{arXiv preprint arXiv:1908.10084}, 2019.

\bibitem[Sch{\"u}rholt et~al.(2021)Sch{\"u}rholt, Kostadinov, and Borth]{urholt2021selfsupervised}
Konstantin Sch{\"u}rholt, Dimche Kostadinov, and Damian Borth.
\newblock Self-supervised representation learning on neural network weights for model characteristic prediction.
\newblock In \emph{Advances in Neural Information Processing Systems}, 2021.

\bibitem[Sen et~al.(2023)Sen, Singh, Agarwal, Agaram, Krishna, and Sridhar]{NEURIPS2023_a0303731}
Bipasha Sen, Gaurav Singh, Aditya Agarwal, Rohith Agaram, Madhava Krishna, and Srinath Sridhar.
\newblock Hyp-nerf: Learning improved nerf priors using a hypernetwork.
\newblock In \emph{Advances in Neural Information Processing Systems}, pages 51050--51064. Curran Associates, Inc., 2023.

\bibitem[Seo et~al.(2023)Seo, Kim, Kim, and Chun]{seo2023dittonerf}
Hoigi Seo, Hayeon Kim, Gwanghyun Kim, and Se~Young Chun.
\newblock Ditto-nerf: Diffusion-based iterative text to omni-directional 3d model, 2023.

\bibitem[Song et~al.(2023)Song, Choi, Do, Lee, and Kim]{Song_2023_ICCV}
Hyeonseop Song, Seokhun Choi, Hoseok Do, Chul Lee, and Taehyeong Kim.
\newblock Blending-nerf: Text-driven localized editing in neural radiance fields.
\newblock In \emph{Proceedings of the IEEE/CVF International Conference on Computer Vision (ICCV)}, pages 14383--14393, 2023.

\bibitem[Sun et~al.(2022)Sun, Sun, and Chen]{sun2022direct}
Cheng Sun, Min Sun, and Hwann-Tzong Chen.
\newblock Direct voxel grid optimization: Super-fast convergence for radiance fields reconstruction.
\newblock In \emph{Proceedings of the IEEE/CVF Conference on Computer Vision and Pattern Recognition}, pages 5459--5469, 2022.

\bibitem[Sun et~al.(2024)Sun, Liu, Han, and Gould]{Sun_2024_WACV}
Chunyi Sun, Yanbin Liu, Junlin Han, and Stephen Gould.
\newblock Nerfeditor: Differentiable style decomposition for 3d scene editing.
\newblock In \emph{Proceedings of the IEEE/CVF Winter Conference on Applications of Computer Vision (WACV)}, pages 7306--7315, 2024.

\bibitem[Touvron et~al.(2023)Touvron, Lavril, Izacard, Martinet, Lachaux, Lacroix, Rozi{\`e}re, Goyal, Hambro, Azhar, et~al.]{llama}
Hugo Touvron, Thibaut Lavril, Gautier Izacard, Xavier Martinet, Marie-Anne Lachaux, Timoth{\'e}e Lacroix, Baptiste Rozi{\`e}re, Naman Goyal, Eric Hambro, Faisal Azhar, et~al.
\newblock Llama: Open and efficient foundation language models.
\newblock \emph{arXiv preprint arXiv:2302.13971}, 2023.

\bibitem[Unterthiner et~al.(2020)Unterthiner, Keysers, Gelly, Bousquet, and Tolstikhin]{Unterthiner2020PredictingNN}
Thomas Unterthiner, Daniel Keysers, Sylvain Gelly, Olivier Bousquet, and Ilya~O. Tolstikhin.
\newblock Predicting neural network accuracy from weights.
\newblock \emph{arXiv}, abs/2002.11448, 2020.

\bibitem[Vaswani et~al.(2017)Vaswani, Shazeer, Parmar, Uszkoreit, Jones, Gomez, Kaiser, and Polosukhin]{transformers}
Ashish Vaswani, Noam Shazeer, Niki Parmar, Jakob Uszkoreit, Llion Jones, Aidan~N Gomez, {\L}ukasz Kaiser, and Illia Polosukhin.
\newblock Attention is all you need.
\newblock In \emph{Advances in Neural Information Processing Systems}. Curran Associates, Inc., 2017.

\bibitem[Wang et~al.(2022)Wang, Chai, He, Chen, and Liao]{wang2022clip}
Can Wang, Menglei Chai, Mingming He, Dongdong Chen, and Jing Liao.
\newblock Clip-nerf: Text-and-image driven manipulation of neural radiance fields.
\newblock In \emph{Proceedings of the IEEE/CVF Conference on Computer Vision and Pattern Recognition}, pages 3835--3844, 2022.

\bibitem[Wang et~al.(2023)Wang, Jiang, Chai, He, Chen, and Liao]{10144678}
Can Wang, Ruixiang Jiang, Menglei Chai, Mingming He, Dongdong Chen, and Jing Liao.
\newblock Nerf-art: Text-driven neural radiance fields stylization.
\newblock \emph{IEEE Transactions on Visualization and Computer Graphics}, pages 1--15, 2023.

\bibitem[Wei et~al.(2021)Wei, Bosma, Zhao, Guu, Yu, Lester, Du, Dai, and Le]{flan}
Jason Wei, Maarten Bosma, Vincent Zhao, Kelvin Guu, Adams~Wei Yu, Brian Lester, Nan Du, Andrew~M Dai, and Quoc~V Le.
\newblock Finetuned language models are zero-shot learners.
\newblock In \emph{International Conference on Learning Representations}, 2021.

\bibitem[Wu et~al.(2023)Wu, Zhang, Fu, Wang, Jiawei~Ren, Wu, Yang, Wang, Qian, Lin, and Liu]{wu2023omniobject3d}
Tong Wu, Jiarui Zhang, Xiao Fu, Yuxin Wang, Liang~Pan Jiawei~Ren, Wayne Wu, Lei Yang, Jiaqi Wang, Chen Qian, Dahua Lin, and Ziwei Liu.
\newblock Omniobject3d: Large-vocabulary 3d object dataset for realistic perception, reconstruction and generation.
\newblock In \emph{IEEE/CVF Conference on Computer Vision and Pattern Recognition (CVPR)}, 2023.

\bibitem[Xu et~al.(2019)Xu, Wang, Ceylan, Mech, and Neumann]{shapenet_render}
Qiangeng Xu, Weiyue Wang, Duygu Ceylan, Radomir Mech, and Ulrich Neumann.
\newblock Disn: Deep implicit surface network for high-quality single-view 3d reconstruction.
\newblock In \emph{Advances in Neural Information Processing Systems}. Curran Associates, Inc., 2019.

\bibitem[Xu et~al.(2023)Xu, Wang, Wang, Chen, Pang, and Lin]{pointllm}
Runsen Xu, Xiaolong Wang, Tai Wang, Yilun Chen, Jiangmiao Pang, and Dahua Lin.
\newblock Pointllm: Empowering large language models to understand point clouds.
\newblock \emph{arXiv preprint arXiv:2308.16911}, 2023.

\bibitem[Yen-Chen et~al.(2022)Yen-Chen, Florence, Barron, Lin, Rodriguez, and Isola]{yen2022nerf}
Lin Yen-Chen, Pete Florence, Jonathan~T Barron, Tsung-Yi Lin, Alberto Rodriguez, and Phillip Isola.
\newblock Nerf-supervision: Learning dense object descriptors from neural radiance fields.
\newblock In \emph{2022 international conference on robotics and automation (ICRA)}, pages 6496--6503. IEEE, 2022.

\bibitem[Yu et~al.(2024)Yu, Wu, Men, Lu, Cui, Xie, and Miao]{10476703}
Yingchen Yu, Rongliang Wu, Yifang Men, Shijian Lu, Miaomiao Cui, Xuansong Xie, and Chunyan Miao.
\newblock Morphnerf: Text-guided 3d-aware editing via morphing generative neural radiance fields.
\newblock \emph{IEEE Transactions on Multimedia}, pages 1--13, 2024.

\bibitem[Zama~Ramirez et~al.(2023)Zama~Ramirez, De~Luigi, Sirocchi, Cardace, Spezialetti, Ballerini, Salti, and Di~Stefano]{nf2vec}
Pierluigi Zama~Ramirez, Luca De~Luigi, Daniele Sirocchi, Adriano Cardace, Riccardo Spezialetti, Francesco Ballerini, Samuele Salti, and Luigi Di~Stefano.
\newblock Deep learning on {3D} neural fields.
\newblock \emph{arXiv preprint arXiv:2312.13277}, 2023.

\bibitem[Zhang et~al.(2023)Zhang, Han, Liu, Gao, Zhou, Hu, Yan, Lu, Li, and Qiao]{LLaMA-Adapter}
Renrui Zhang, Jiaming Han, Chris Liu, Peng Gao, Aojun Zhou, Xiangfei Hu, Shilin Yan, Pan Lu, Hongsheng Li, and Yu Qiao.
\newblock Llama-adapter: Efficient fine-tuning of language models with zero-init attention.
\newblock \emph{arXiv preprint arXiv:2303.16199}, 2023.

\bibitem[Zhou et~al.(2023{\natexlab{a}})Zhou, Yang, Burns, Cardace, Jiang, Sokota, Kolter, and Finn]{zhou2023permutation}
Allan Zhou, Kaien Yang, Kaylee Burns, Adriano Cardace, Yiding Jiang, Samuel Sokota, J~Zico Kolter, and Chelsea Finn.
\newblock Permutation equivariant neural functionals.
\newblock \emph{Advances in neural information processing systems}, 37, 2023{\natexlab{a}}.

\bibitem[Zhou et~al.(2023{\natexlab{b}})Zhou, Yang, Jiang, Burns, Xu, Sokota, Kolter, and Finn]{zhou2023neural}
Allan Zhou, Kaien Yang, Yiding Jiang, Kaylee Burns, Winnie Xu, Samuel Sokota, J~Zico Kolter, and Chelsea Finn.
\newblock Neural functional transformers.
\newblock \emph{Advances in neural information processing systems}, 37, 2023{\natexlab{b}}.

\bibitem[Zhou et~al.(2024)Zhou, Finn, and Harrison]{zhou2024universal}
Allan Zhou, Chelsea Finn, and James Harrison.
\newblock Universal neural functionals.
\newblock \emph{arXiv preprint arXiv:2402.05232}, 2024.

\bibitem[Zhu et~al.(2023)Zhu, Ma, Chen, Deng, Huang, and Li]{3d-vista}
Ziyu Zhu, Xiaojian Ma, Yixin Chen, Zhidong Deng, Siyuan Huang, and Qing Li.
\newblock 3d-vista: Pre-trained transformer for 3d vision and text alignment.
\newblock \emph{ICCV}, 2023.

\bibitem[Zhuang et~al.(2023)Zhuang, Wang, Lin, Liu, and Li]{zhuang2023dreameditor}
Jingyu Zhuang, Chen Wang, Liang Lin, Lingjie Liu, and Guanbin Li.
\newblock Dreameditor: Text-driven 3d scene editing with neural fields.
\newblock In \emph{SIGGRAPH Asia 2023 Conference Papers}, pages 1--10, 2023.

\end{thebibliography}
}

\newpage
\appendix
% \section{Appendix / Supplemental Material}
% This section provides additional details, experiments, and qualitative results.
\section{Details on NeRFs}
\label{sec:nerf_details}
We report here some details regarding the \nerf{}s of the ShapeNeRF--Text dataset, which were trained by \citet{nf2vec}. The \nerf{} code is implemented leveraging the {\tt NerfAcc} library \citep{nerfacc}.

\subsection{Architecture}
\label{sec:architecture}
An instance of the employed NeRFs consists of a multi-layer perceptron (MLP) that contains three hidden layers, each with $64$ neurons. The ReLU activation function is applied between all layers except for the last one, which calculates the density and RGB values directly without any activation function. A frequency encoding~\cite{nerf} is applied to the input 3D coordinates, in order to improve the \nerf{} reconstruction quality. NeRFs do not take as input the view direction.
The MLP processes an input coordinate $\vb{p} \in \mathbb{R}^3$, to produce a 4-dimensional vector containing $RGB\sigma$.

\subsection{Training}
\label{sec:nerf_training}
Training a \nerf{} consists of minimizing the error between the rendered images from the NeRF and the ground truth images. Our \nerf{}s were trained using an $L_1$ loss between the predicted and ground truth RGB pixel intensities, weighting background pixels less than foreground pixels ($0.8$ foregrounds vs. $0.2$ background). 
Image rendering involves querying the neural network by feeding it 3D coordinates to obtain RGB color values and density estimates. By integrating these outputs along camera rays using volumetric rendering techniques~\cite{nerf}, colors and opacities are accumulated to produce the final rendered image.
Each \nerf{} is trained until it reaches a good reconstruction quality, approximately for $2000$ steps.

\subsection{Generating images and point clouds from NeRFs}
\label{sec:extract_data_from_nerf}
\begin{figure}[b]
    \centering
    \setlength{\tabcolsep}{1pt}
    \begin{tabular}{cccccc}
    GT Front & Rendered Front & GT Back & Rendered Back & GT Points & Extracted Points \\
         \includegraphics[width=0.15\linewidth]{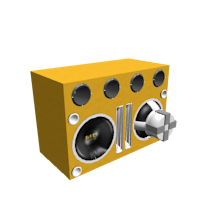} &
         \includegraphics[width=0.15\linewidth]{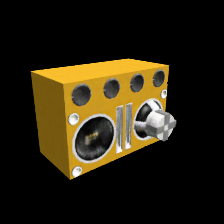} &
         \includegraphics[width=0.15\linewidth]{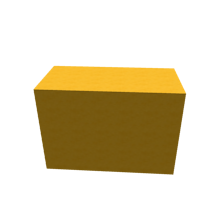} &
         \includegraphics[width=0.15\linewidth]{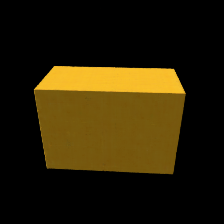} &
         \includegraphics[width=0.15\linewidth]{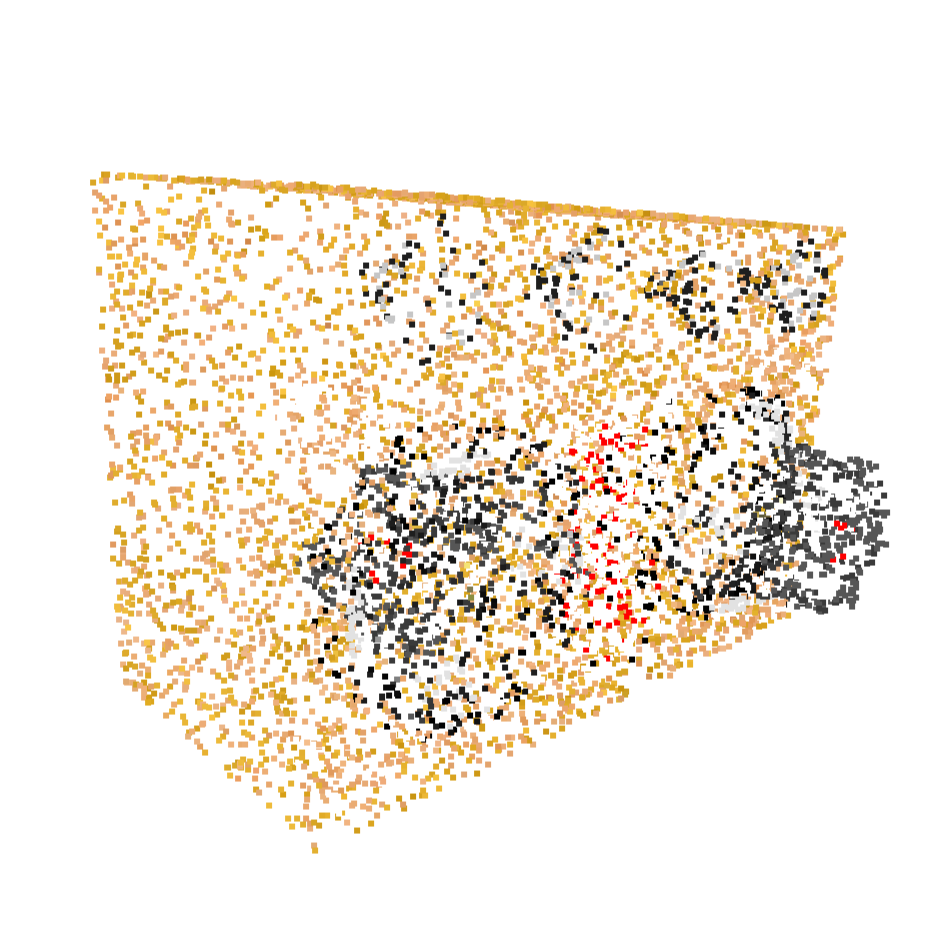} &
         \includegraphics[width=0.15\linewidth]{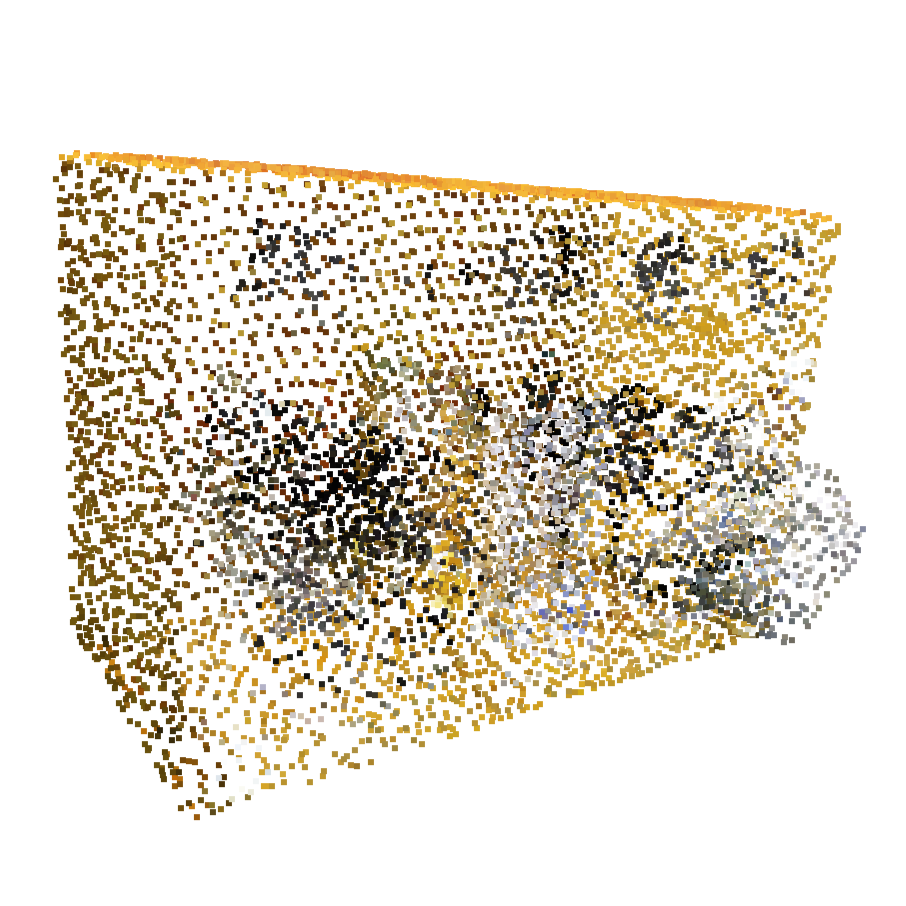}
    \end{tabular}
    
    \caption{\textbf{Example of data extracted from NeRF.} From left to right: GT front view, rendered front view, GT back view, rendered back view, GT point cloud, extracted point cloud.}
    \label{fig:nerf_sample}
\end{figure}

To compare with 2D and 3D MLLMs on the new tasks of NeRF captioning and NeRF Q\&A, we need to render images or reconstruct point clouds from the NeRF. To render images, we employ the same volumetric rendering procedure used during the NeRF's training. In order to extract a point cloud, the marching cubes algorithm is first applied to the volumetric density field derived from the NeRF. This process generates a mesh by identifying isosurfaces within the density field. The mesh is then converted into a point cloud by considering only the mesh vertices, uniformly distributed in the 3D space. We sample RGB values from NeRF for each point coordinate to approximate point cloud colors. 
An example of data extracted from NeRF is depicted in \cref{fig:nerf_sample}.

Generating images and point clouds requires the user to make some decisions, the effects of which on the assistant's performance are not easy to anticipate.
When dealing with images, it is difficult to select the rendering viewpoint.
It might happen that the object is not clearly visible from the chosen viewpoint or that important elements are missing. 
Another decision is the resolution of the generated image, which, if too coarse, may prevent the identification of fine-grained details. 
The same concerns regarding the resolution also apply to point clouds. Yet, the modality encoder may not handle large resolutions or may greatly increase the processing time.
Another important point is the additional computational time required to extract data from NeRF. For instance, extracting point clouds from NeRF with only $8192$ points requires approximately $620$\texttt{ms}. Moreover, the time for sampling the MLP and running a marching cube algorithm scales cubically with the desired spatial resolution.
On the other hand, the time required to process the MLP weights is independent of the spatial resolution.

\subsection{NeRF memory occupation compared to images or point clouds}
\label{sec:memory_nerf_vs_explicit}
\begin{figure}[ht]
    \centering
    \setlength{\tabcolsep}{1pt}
    \scalebox{1}{
    \begin{tabular}{cc}
    \includegraphics[width=0.44\linewidth]{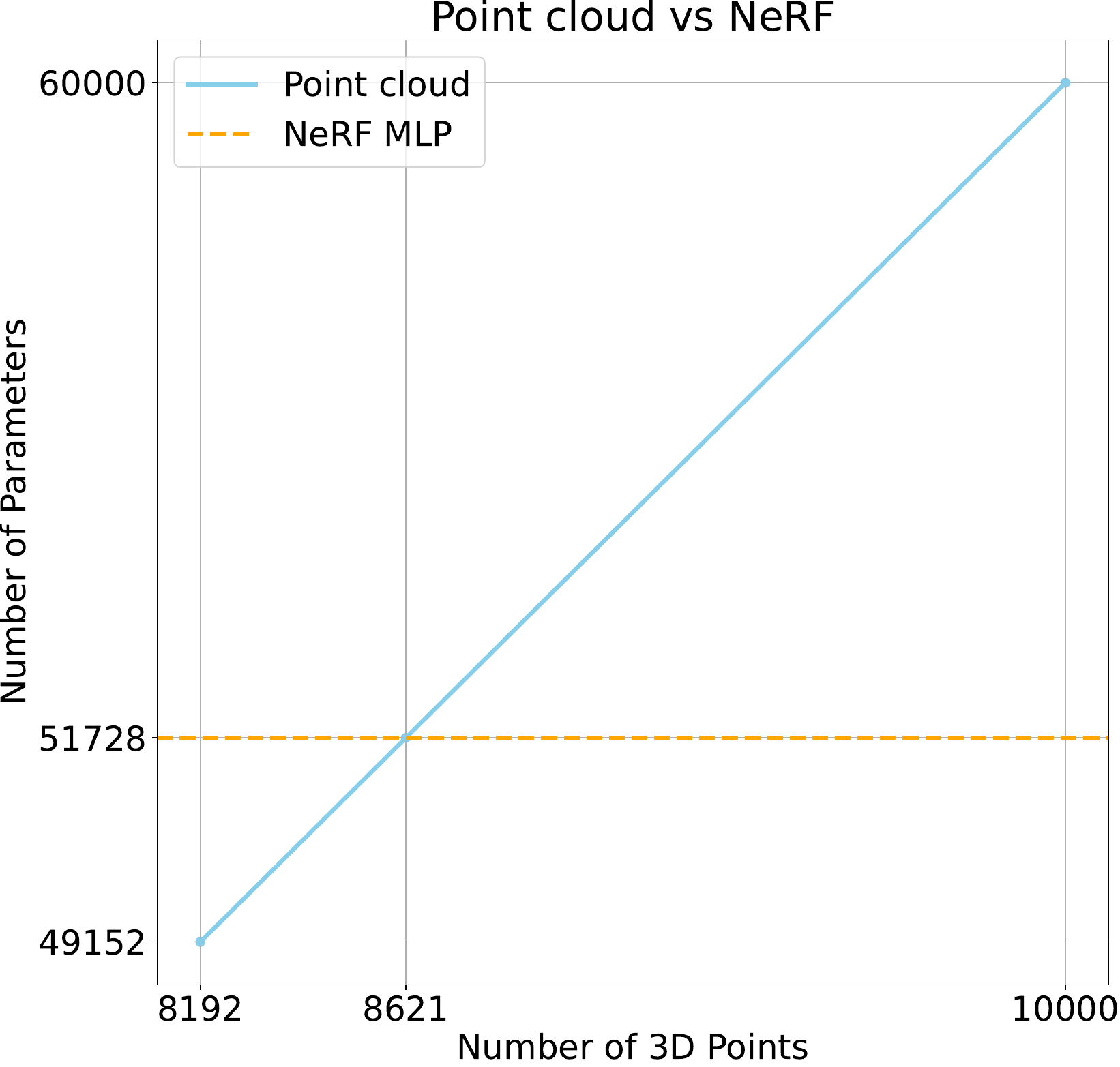}
    \includegraphics[width=0.49\linewidth]{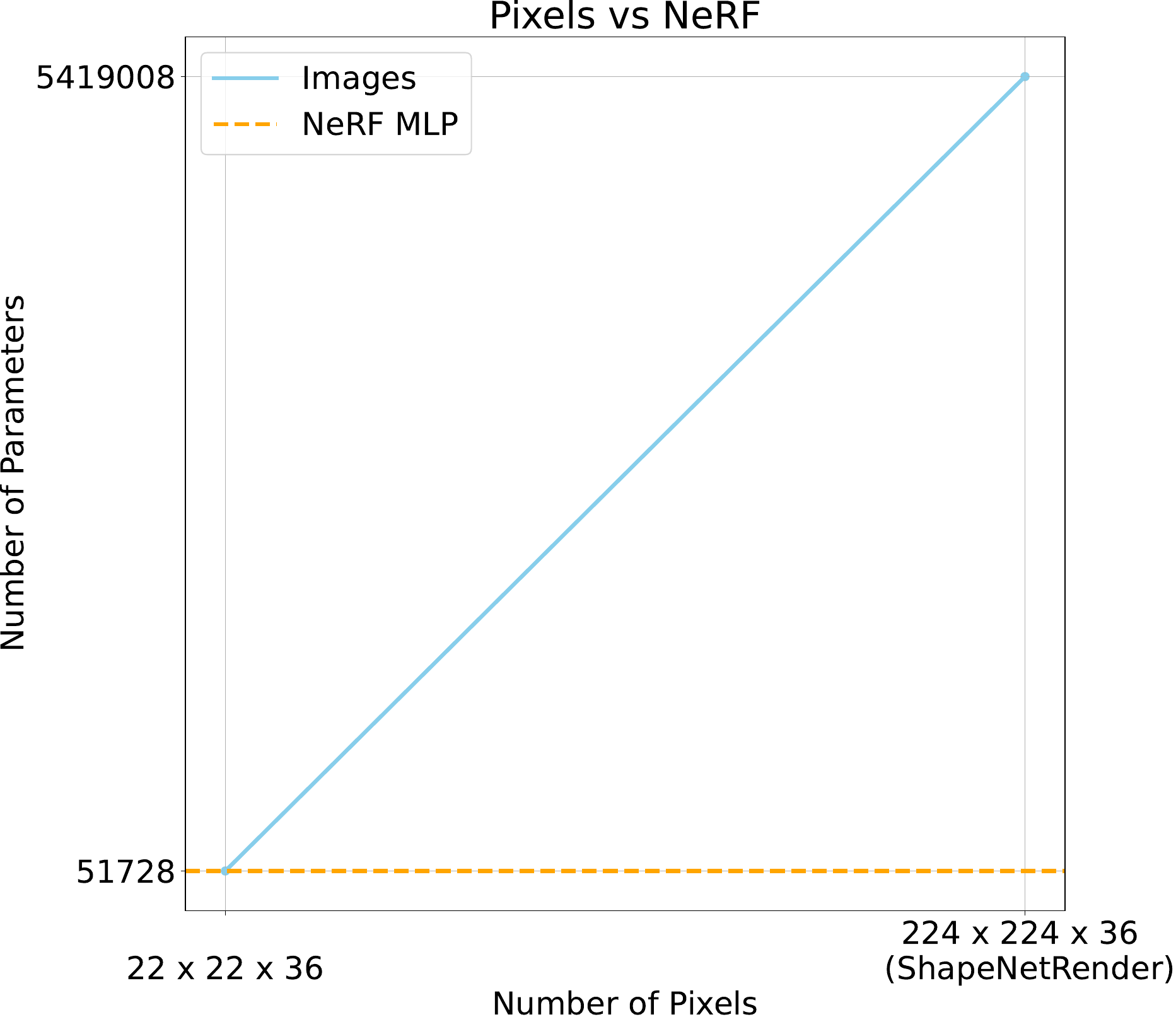}
    \end{tabular}}
    \caption{\textbf{Memory usage of NeRF compared to images or point clouds.} Left: NeRF vs point clouds. Right: NeRF vs pixels.}
    \label{fig:memory}
\end{figure}

An important benefit of using NeRF to represent objects is that memory consumption is decoupled from spatial resolution.
In \cref{fig:memory}, we analyze the number of parameters needed for point clouds and images compared to neural fields by altering the spatial resolution of the data. We account for all variables required by an explicit representation in their parameter count. For instance, each point in a point cloud has six parameters corresponding to its coordinates ($x$, $y$, $z$) and color ($R$, $G$, $B$), while each pixel has only three channels ($R$, $G$, $B$). The orange line represents the parameters of the NeRF MLP, while the blue lines indicate the parameters for 3D points (\cref{fig:memory}-left) and image pixels (\cref{fig:memory}-right).

We observe that the space occupied by the NeRF MLP is comparable to that used by point clouds in our experiments (i.e., $8192$ points, the data size used in GPT4Point~\cite{gpt4point} and PointLLM~\cite{pointllm}). However, NeRF becomes advantageous for representing data as soon as the point cloud size is greater than $8621$ points. This is crucial, considering real datasets may contain point clouds or meshes with significantly more points or faces; for example, Objaverse~\citep{deitke2023objaverse} features meshes with over $10^7$ polygons.

The advantages are even more pronounced for images, where a single NeRF MLP corresponds to $36$ images at a resolution of $22\times22$. Storing the $36$ pictures from ShapeNetRender at $256\times256$ resolution, used to train our NeRF on a single object, requires substantially more memory.

\section{Details on the Meta-Encoder}
\label{sec:nftovec_supp}
We employ \nftovec{}~\cite{nf2vec} as the meta-encoder of \model{}. Thus, in the following paragraphs, we describe the details of the \nftovec{} architecture and training protocol.

\subsection{Architecture}
The \nftovec{} encoder consists of 4 linear layers with $512$, $512$, $512$, and $1024$ output neurons each. The final embedding extracted for a given MLP is a $1024$-dimensional array. Each linear layer is followed by a batch normalization and a ReLU activation function. The weight matrices and biases of the input NeRF are stacked along the row dimension to form a matrix $\mathbf{M} \in \mathbb{R}^{S \times H}$, where $S= 144 + 1 + L*(H+1) + H + 1 = L*H + L + H + 146$ and $L$ is the number of units per hidden layer, while $H$ is the dimension of the input.

\subsection{Training}
We employ the official code and weights of \nftovec{}\footnote{\url{https://cvlab-unibo.github.io/nf2vec/}} pre-trained on an augmented version of ShapeNetRender~\cite{shapenet_render}. The encoder was trained in an end-to-end manner together with an implicit decoder. The decoder takes in input 3D coordinates after a frequency encoding and the global $1024$-dimensional output of the encoder. It consists of $5$ linear layers with $1024$ neurons each, followed by ReLU activations except for the last layer. It yields a $4$-dimensional output $RGB\sigma$, similar to the NeRF given in input to the encoder. The framework supervision comes from the pixel-wise rendering $L_1$ error computed between the ground truth RGB image and the predicted image, which is obtained through volumetric rendering after encoding and decoding the NeRF. 

\begin{figure}[t]
    \centering
    \includegraphics[width=\linewidth]{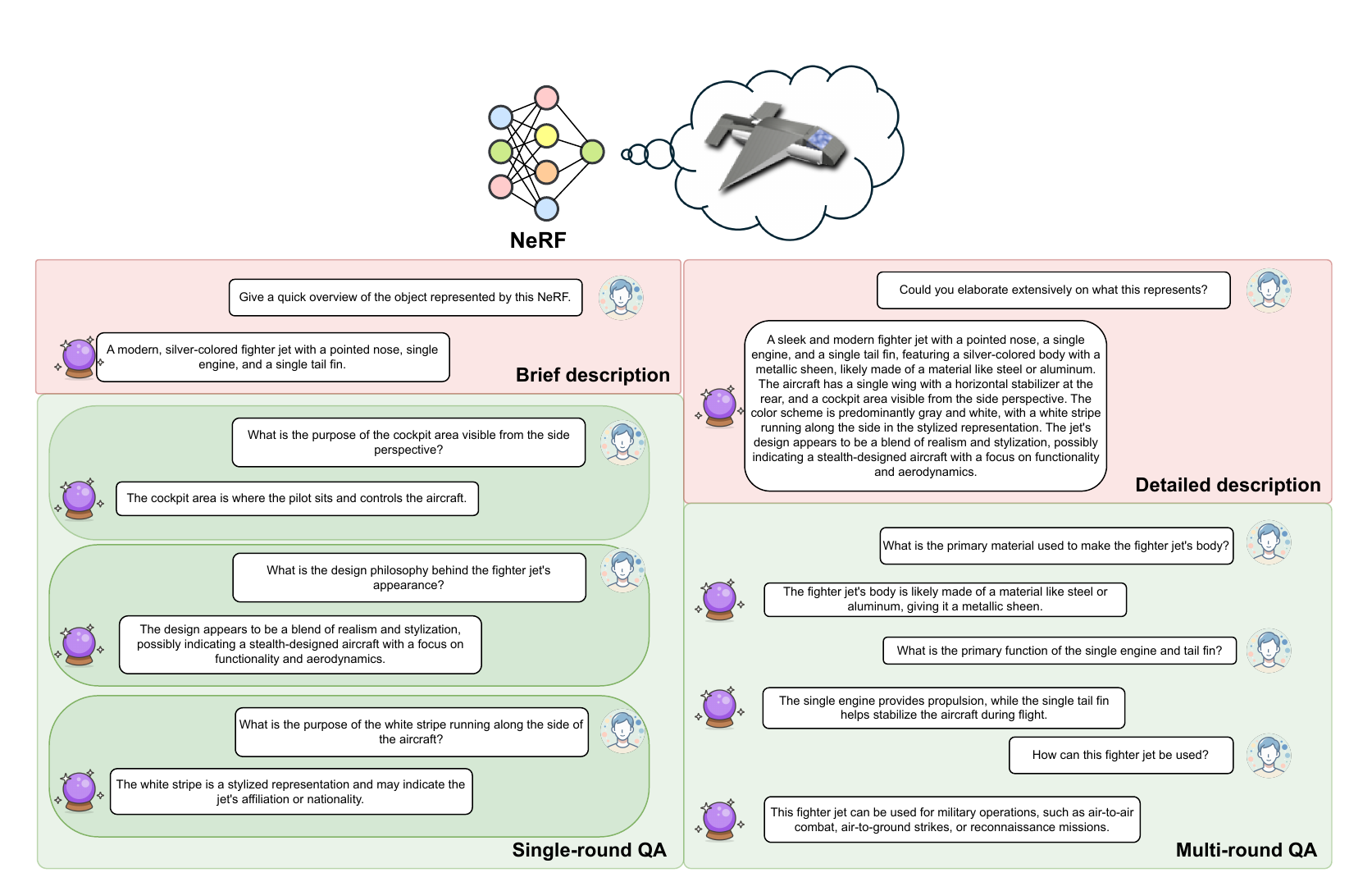}
    \caption{\textbf{Example of data sample from ShapeNeRF--Text dataset.}}
    \label{fig:ex_dataset}
\end{figure}
\section{Details on ShapeNeRF--Text dataset}
\label{sec:dataset_supp}
The proposed ShapeNeRF--Text dataset consists of 40K paired \nerf{}s and language annotations for ShapeNet objects~\cite{chang2015shapenet}.
In particular, for every 3D model, multiple annotations have been provided: a brief description, a detailed description, 3 single-round Q\&As, and one multi-round Q\&A. Figure~\ref{fig:ex_dataset} shows an example of such annotations. These annotations have been obtained by exploiting \llava{} 2 and \llama{} 3 as described in section~\ref{sec:dataset} of the main paper.

\subsection{Instruction prompts and ground-truth questions}
\label{sec:instruction_prompts_gt_questions}
In this section, we provide the instruction prompts used to generate the ground-truth answers of ShapeNeRF--Text and the list of questions used to build the ground-truth questions for the brief and detailed descriptions.

\paragraph{Instruction prompts for \llava{} and \llama{} to generate the dataset}
For constructing ShapeNerf--Text, first, descriptive captions for multiple views of each object have been obtained using the following input request to \llava{}: 
\begin{itemize}
\item
\emph{``USER:<image>\textbackslash{}nYou will be provided the image of an object, seen from the <view\_point>. Describe the object in detail. Include as much information as possible, but do not infer anything that is not in the image. Avoid describing the background. Generate an answer with a maximum length of 30 words.\textbackslash{}nASSISTANT:''}
\end{itemize}
The placeholder <view\_point> was replaced with ``back'', ``side'', or ``front'' according to the viewpoint of the image provided as input. To expedite computation and leverage the high symmetry of ShapeNet objects, $7$ views have been employed for this process.

After obtaining the captions for each view, \llama{} was queried to aggregate these single-view captions into 
 comprehensive descriptions and Q\&A rounds. The input provided to \llama{} was:
\begin{itemize}
\item \emph{You will be shown 7 different descriptions of an object, obtained from different points of view. Please provide two descriptions, which aggregates all these ones. The first description must be concise, the second one will be more descriptive. Both these description must refer to the same subject. Avoid repetitions.
Important: The output descriptions must be followed by the string "Final concise description:" and "Final more detailed description:".
Notice: There are errors in some descriptions, due to occlusion and improper angle. You need to combine all the descriptions and eliminate possible wrong details (please fix the errors directly, do not tell me).
Input descriptions:} [list of the single-view captions generated by \llava{}]
\end{itemize}
The detailed description was then used to generate multiple Q\&A rounds, through the following request: 
\begin{itemize}
\item \emph{Given this description of an object, generate 6 short Q\&A dialogues regarding diverse aspects of the object described, ensuring logical relevance between the questions and answers. Include always a question about how this object can be used. Question begins with 'Q'. Answer begins with 'A'. 
IMPORTANT: Do not mention size, background. Do not mention "how many". Do not add text after the last answer."}. 
\end{itemize}
From the 6 generated Q\&A pairs, 3 were randomly sampled to build the sequence of multi-round Q\&A, while the remaining pairs were used as single-round Q\&A.

\paragraph{Ground-truth questions for the brief and detailed descriptions}
\cref{table:brief_prompt} and \cref{table:det_prompt} provide the list of questions used to build the ground-truth data of ShapeNeRF--Text, as explained in \cref{sec:shapenerf}.

\begin{longtable}[t]{p{0.95\textwidth}}
\caption{List of questions to prompt the model to produce brief descriptions. An instruction from the list is randomly selected and coupled with a ShapeNeRF--Text brief caption to form a ground-truth data sample.} 
\label{table:brief_prompt} \\
\endfirsthead
\endhead
\textbullet\ Summarize the 3D object briefly. \\
\textbullet\ What kind of object is depicted by this NeRF? \\
\textbullet\ Provide a short explanation of this object. \\
\textbullet\ What does this NeRF represent? \\
\textbullet\ Can you give a brief overview of this object? \\
\textbullet\ Characterize the object this NeRF is illustrating. \\
\textbullet\ Share a brief interpretation of this NeRF. \\
\textbullet\ Provide an outline of this 3D shape's characteristics. \\
\textbullet\ What object is this NeRF rendering? \\
\textbullet\ Deliver a quick description of the object represented here. \\
\textbullet\ How would you describe the 3D form shown in this NeRF? \\
\textbullet\ What is the nature of the object this NeRF is representing? \\
\textbullet\ Present a compact account of this 3D object's key features. \\
\textbullet\ What can you infer about the object from this NeRF? \\
\textbullet\ Offer a clear and concise description of this object. \\
\textbullet\ How would you summarize this 3D data? \\
\textbullet\ Give a brief explanation of the object that this NeRF represents. \\
\textbullet\ What kind of structure does this NeRF depict? \\
\textbullet\ Could you delineate the object indicated by this NeRF? \\
\textbullet\ Express in brief, what this NeRF is representing. \\
\textbullet\ Give a quick overview of the object represented by this NeRF. \\
\textbullet\ Convey a summary of the 3D structure represented in this NeRF. \\
\textbullet\ What kind of object is illustrated by this NeRF? \\
\textbullet\ Describe the object that this NeRF forms. \\
\textbullet\ How would you interpret this NeRF? \\
\textbullet\ Can you briefly outline the shape represented by this NeRF? \\
\textbullet\ Give a concise interpretation of the 3D data presented here. \\
\textbullet\ Explain the object this NeRF depicts succinctly. \\
\textbullet\ Offer a summary of the 3D object illustrated by this NeRF. \\
\end{longtable}

\begin{longtable}[t]{p{0.95\textwidth}}
\caption{List of questions to prompt the model to produce detailed descriptions. An instruction from the list is randomly selected and paired with a ShapeNeRF--Text detailed caption to form a ground-truth data sample.}
\label{table:det_prompt} \\
\endfirsthead
\endhead
\textbullet\ Can you tell me more about this? \\
\textbullet\ What does this represent? \\
\textbullet\ Can you describe this in more detail? \\
\textbullet\ I'm interested in this. Can you explain? \\
\textbullet\ Could you provide more information about this? \\
\textbullet\ What exactly am I looking at here? \\
\textbullet\ What is this? \\
\textbullet\ Could you describe the detailed structure of this? \\
\textbullet\ This looks interesting. Can you expand on it? \\
\textbullet\ Can you explain more about this form? \\
\textbullet\ What can you tell me about the shape of this object? \\
\textbullet\ Could you delve deeper into this? \\
\textbullet\ I want to know more about this. Can you help? \\
\textbullet\ Can you walk me through the details of this object? \\
\textbullet\ Can you provide a comprehensive account of this object? \\
\textbullet\ Offer a detailed interpretation of this NeRF. \\
\textbullet\ Please elucidate on the characteristics of this form. \\
\textbullet\ Could you provide an in-depth description of this structure? \\
\textbullet\ What does this NeRF represent in its entirety? \\
\textbullet\ Elaborate on the details of this NeRF, please. \\
\textbullet\ Kindly furnish me with more information about this object. \\
\textbullet\ Please expand on the intricate structure of this form. \\
\textbullet\ Provide a meticulous explanation of what this NeRF represents. \\
\textbullet\ Provide a detailed explanation of what this NeRF represents. \\
\textbullet\ I request a detailed breakdown of this structure. \\
\textbullet\ Give a thorough rundown of this NeRF. \\
\textbullet\ Can you offer a complete analysis of this object? \\
\textbullet\ I would like a comprehensive explanation of this form. \\
\textbullet\ Please detail the specific features of this NeRF. \\
\textbullet\ Could you elaborate extensively on what this represents? \\
\end{longtable}

\subsection{ShapeNeRF--Text statistics}
\label{sec:dataset_stats}
\begin{figure}[t!]
    \centering
    \caption{\textbf{ShapeNeRF-Text statistics for ground-truth brief and detailed descriptions.}}

    \setlength{\tabcolsep}{1pt}
    \begin{tabular}{c}
    \begin{tabular}{cc}
    \multicolumn{2}{c}{\small Brief Descriptions - Word clouds} \\
    \includegraphics[width=0.5\linewidth]{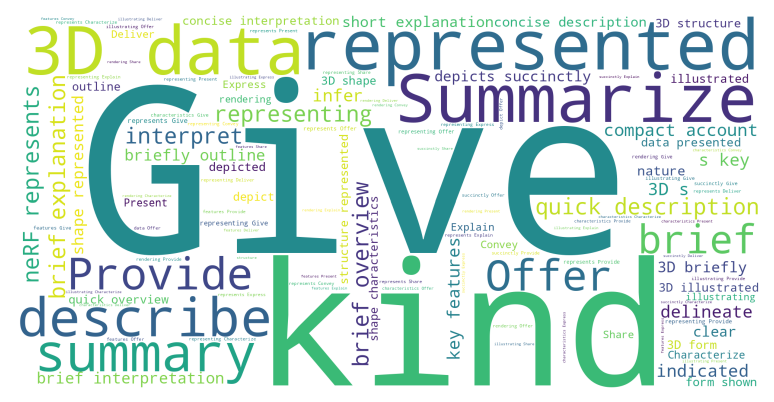} & \includegraphics[width=0.5\linewidth]{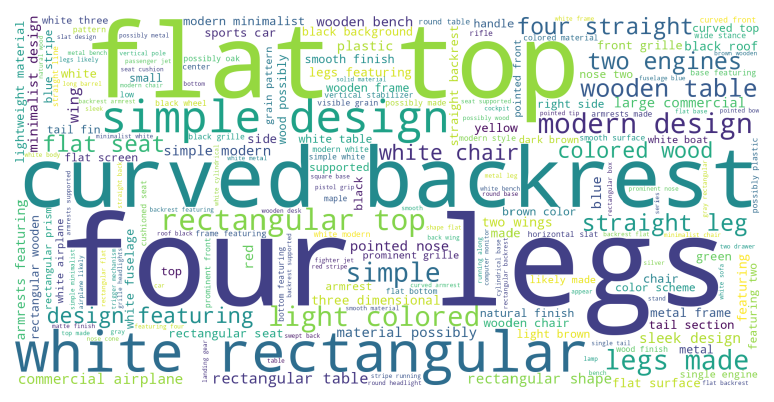} \\
    {\small Instructions} & {\small Responses} \\
    
    \multicolumn{2}{c}{\small Brief Descriptions - Lengths (Words)} \\
    \includegraphics[width=0.5\linewidth]{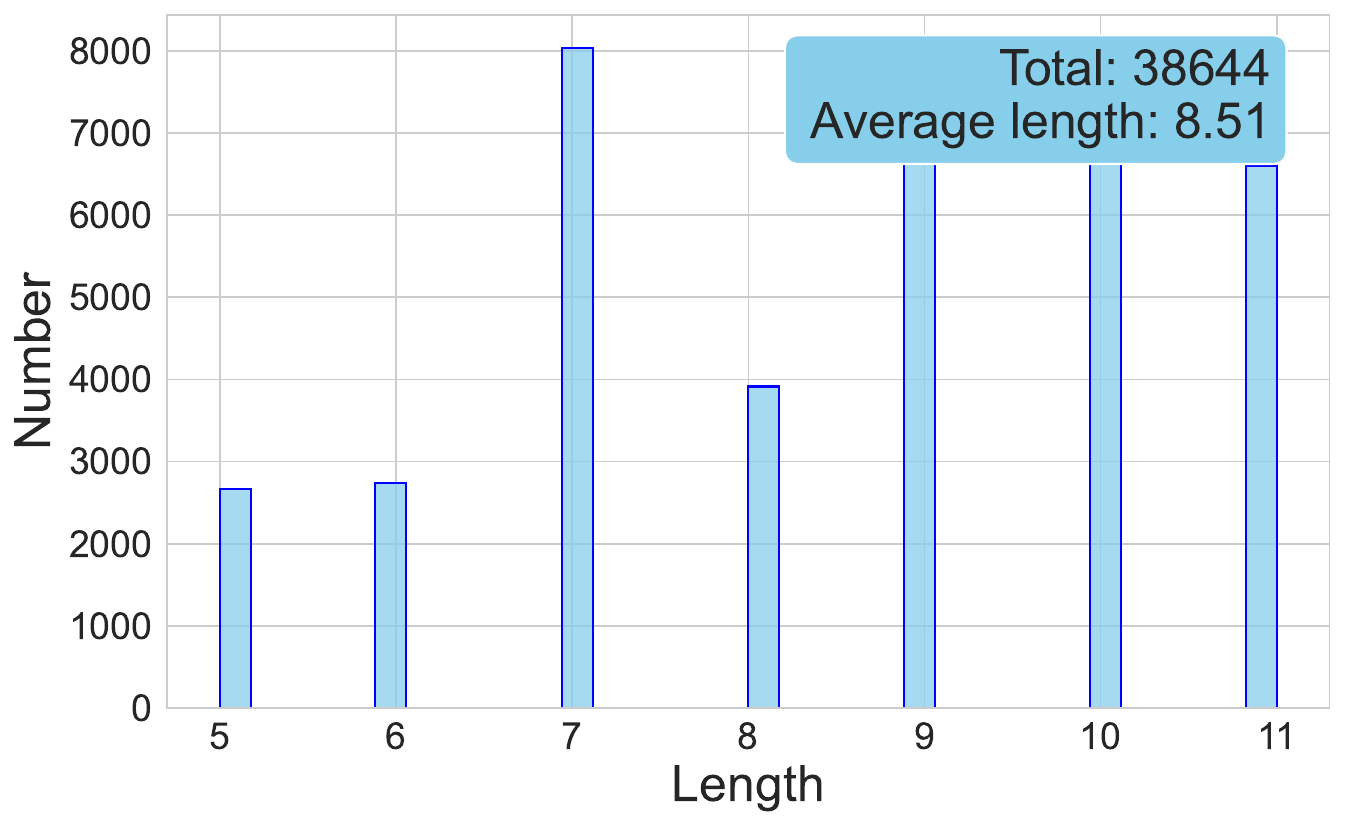} & \includegraphics[width=0.5\linewidth]{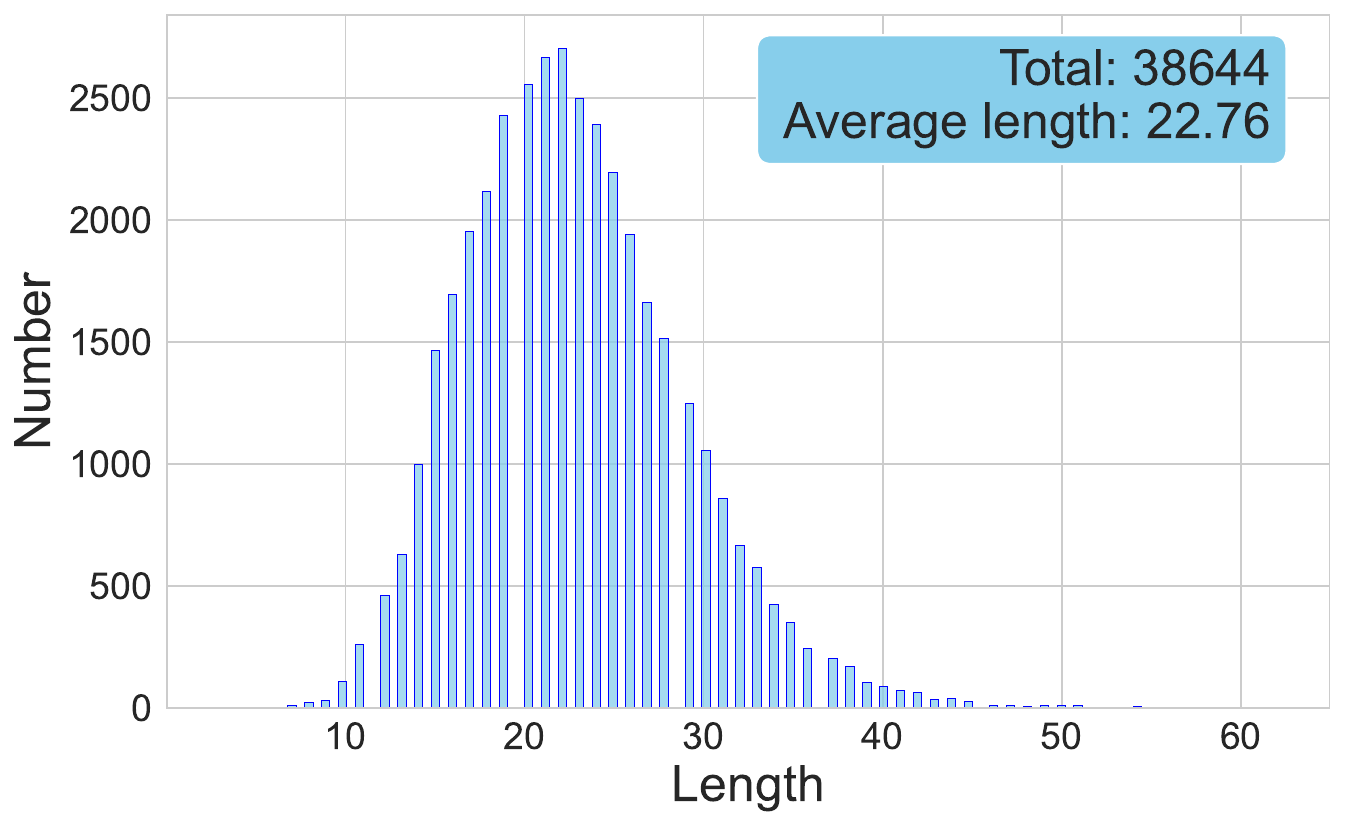} \\
    {\small Instructions} & {\small Responses} \\

    \multicolumn{2}{c}{\small Detailed Descriptions - Word clouds} \\
    \includegraphics[width=0.5\linewidth]{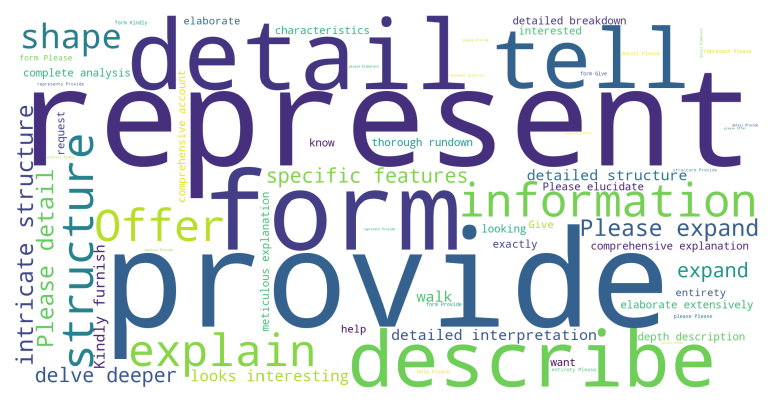} & \includegraphics[width=0.5\linewidth]{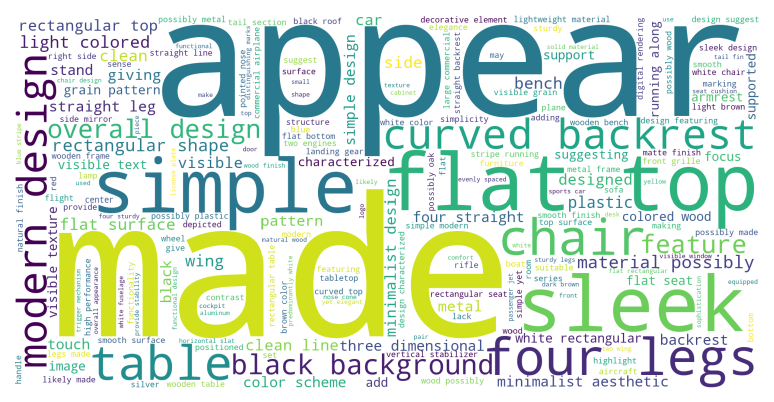} \\
    {\small Instructions} & {\small Responses} \\
    
    \multicolumn{2}{c}{\small Detailed Descriptions - Lengths (Words)} \\
    \includegraphics[width=0.5\linewidth]{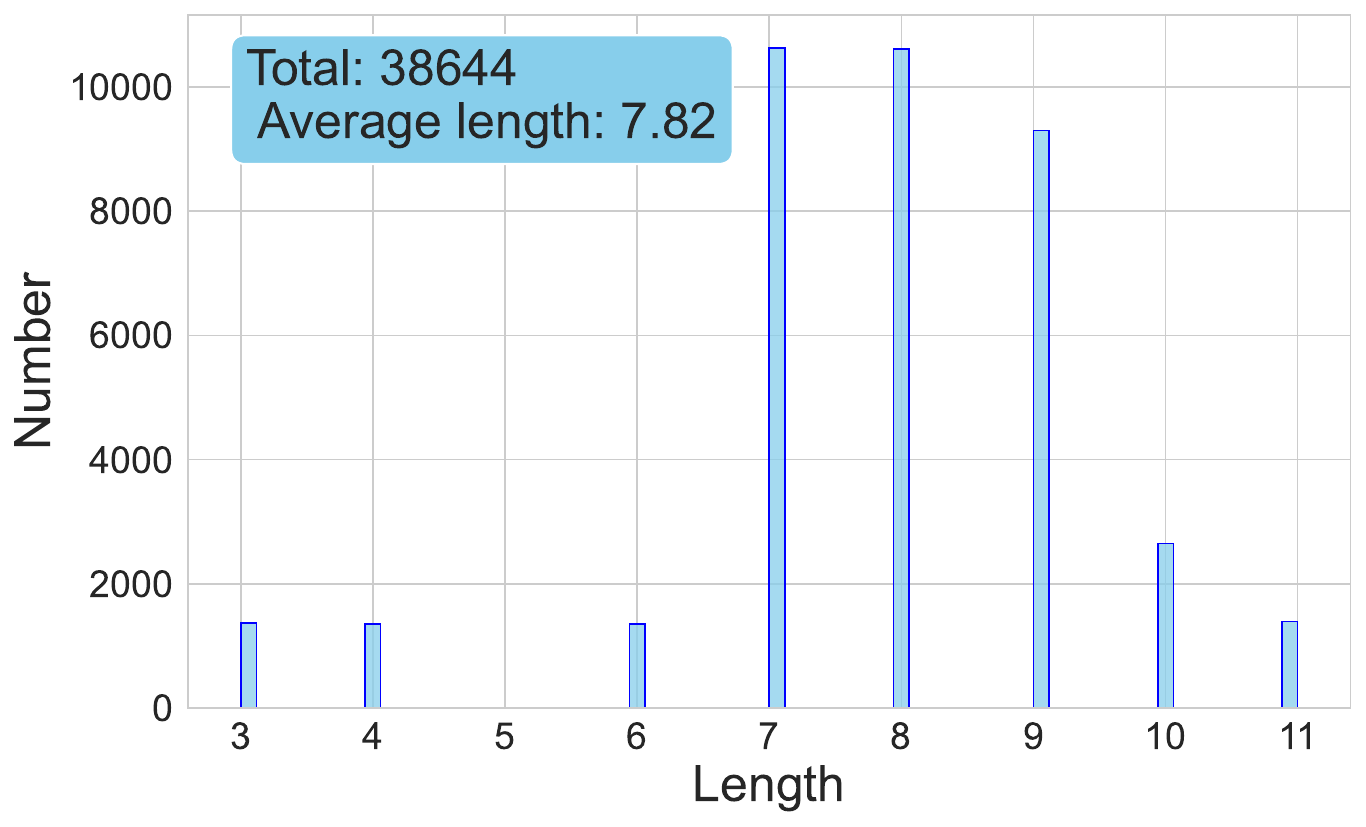} & \includegraphics[width=0.5\linewidth]{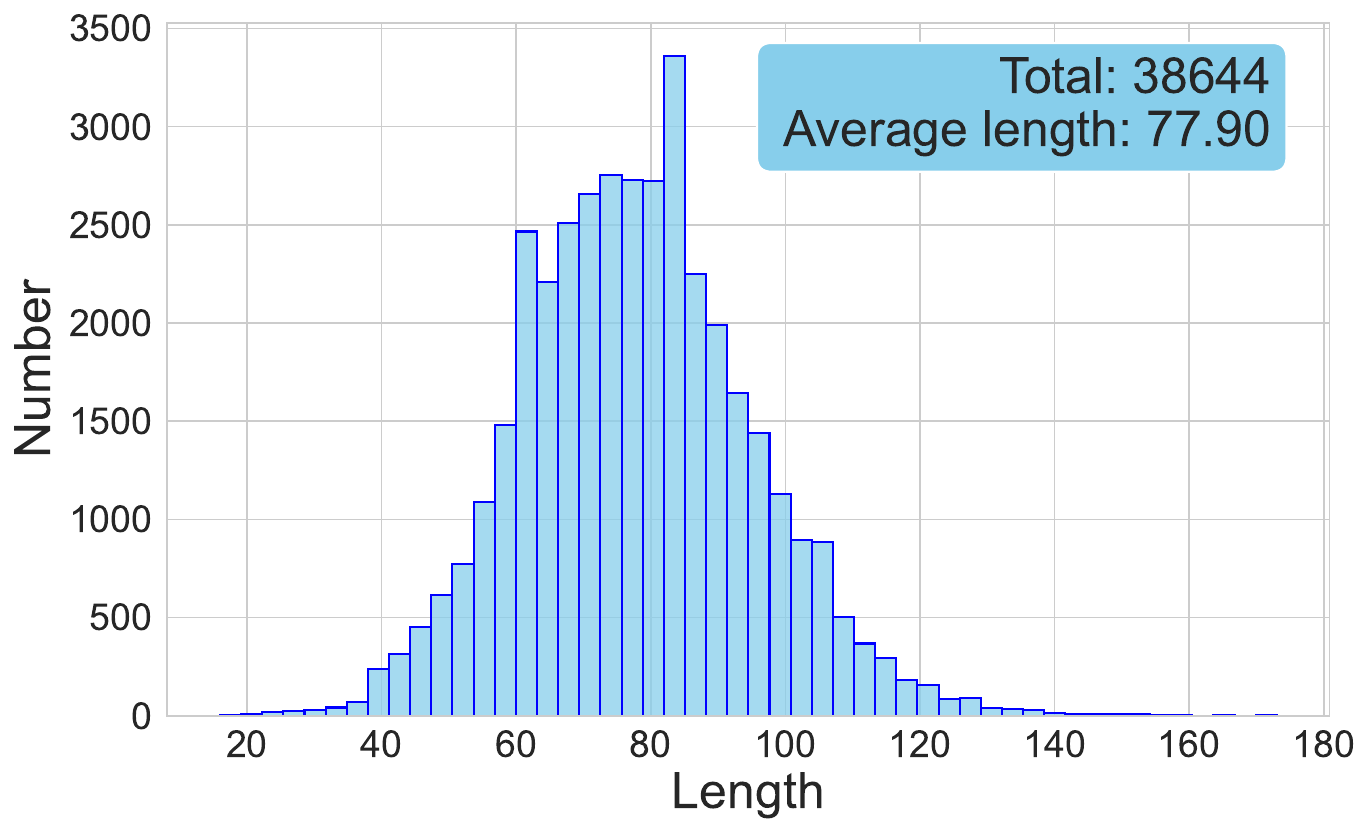} \\
    {\small Instructions} & {\small Responses} \\

    \end{tabular} 
    \end{tabular}
    \label{tab:dataset_stats_cap}
\end{figure}

\begin{figure}[t!]
    \centering
    \caption{\textbf{ShapeNeRF-Text statistics for ground-truth single-round and multi-round Q\&A.}}

    \setlength{\tabcolsep}{1pt}
    \begin{tabular}{c}
    \begin{tabular}{cc}
    \multicolumn{2}{c}{\small Single-round Q\&A - Word clouds} \\
    \includegraphics[width=0.5\linewidth]{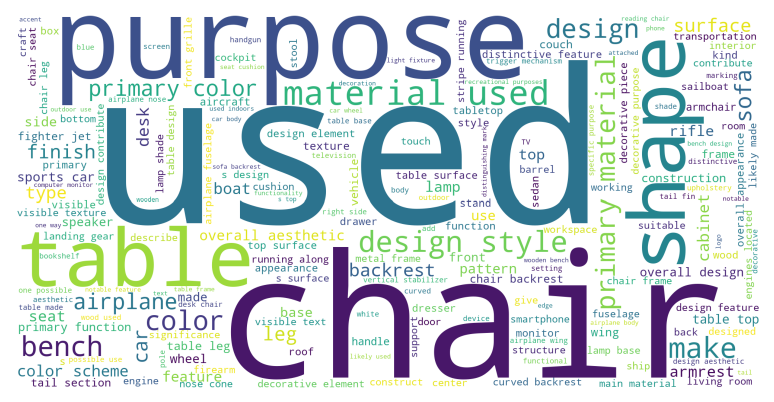} & \includegraphics[width=0.5\linewidth]{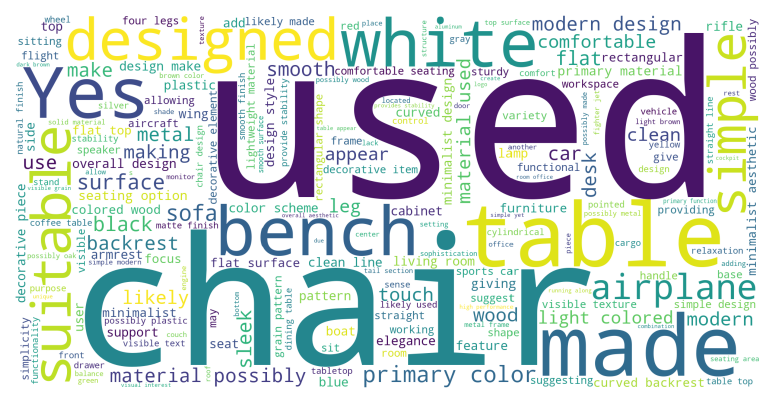} \\
    {\small Instructions} & {\small Responses} \\
    
    \multicolumn{2}{c}{\small Single-round Q\&A - Lengths (Words)} \\
    \includegraphics[width=0.5\linewidth]{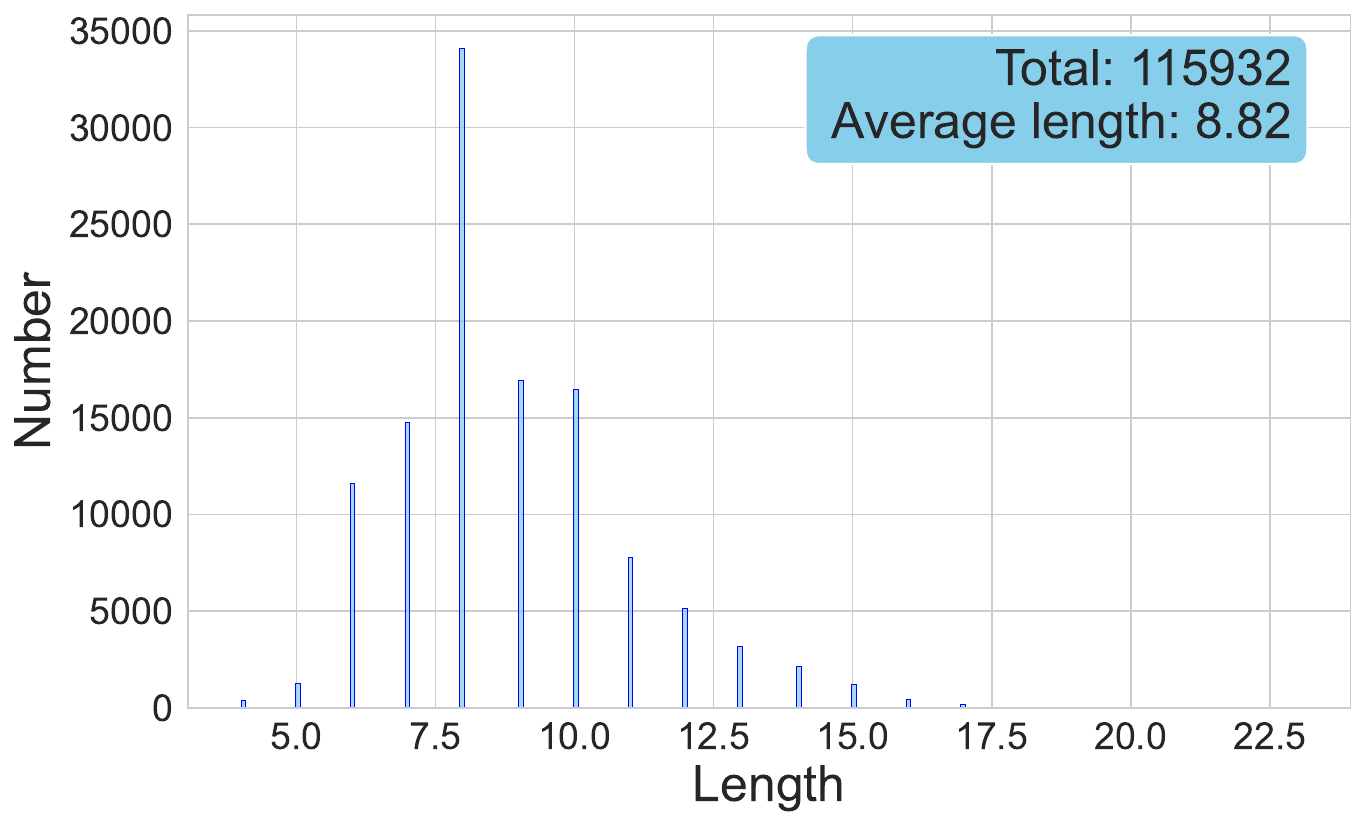} & \includegraphics[width=0.5\linewidth]{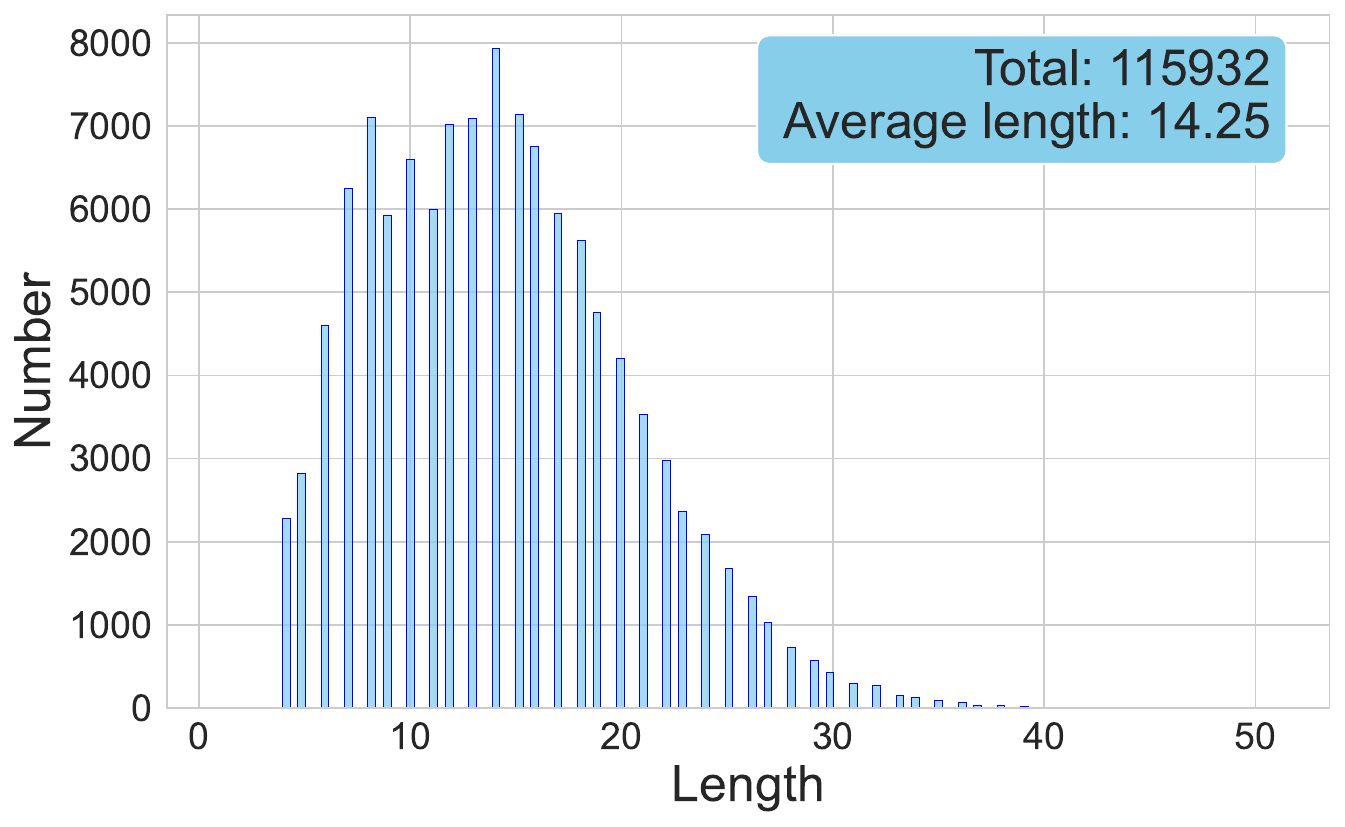} \\
    {\small Instructions} & {\small Responses} \\

    \multicolumn{2}{c}{\small Multi-round Q\&A - Word clouds} \\
    \includegraphics[width=0.5\linewidth]{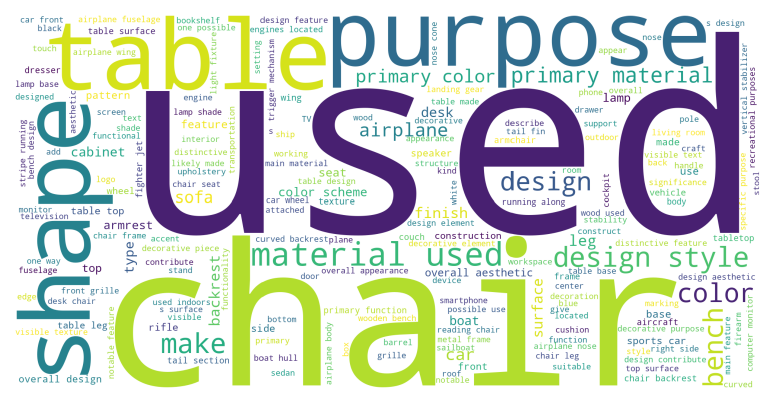} & \includegraphics[width=0.5\linewidth]{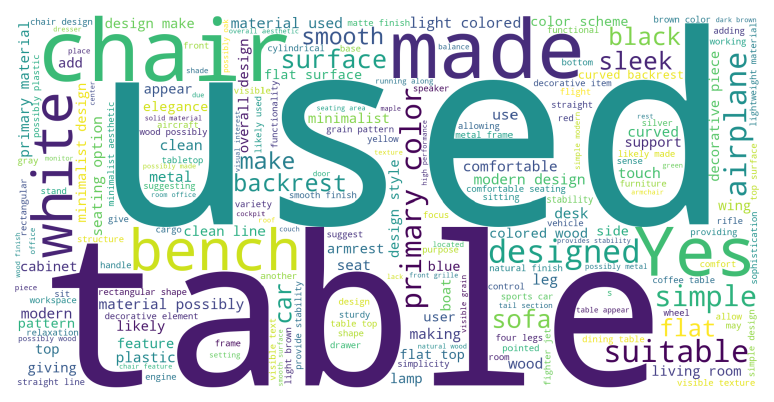} \\
    {\small Instructions} & {\small Responses} \\
    
    \multicolumn{2}{c}{\small Multi-round Q\&A - Lengths (Words)} \\
    \includegraphics[width=0.5\linewidth]{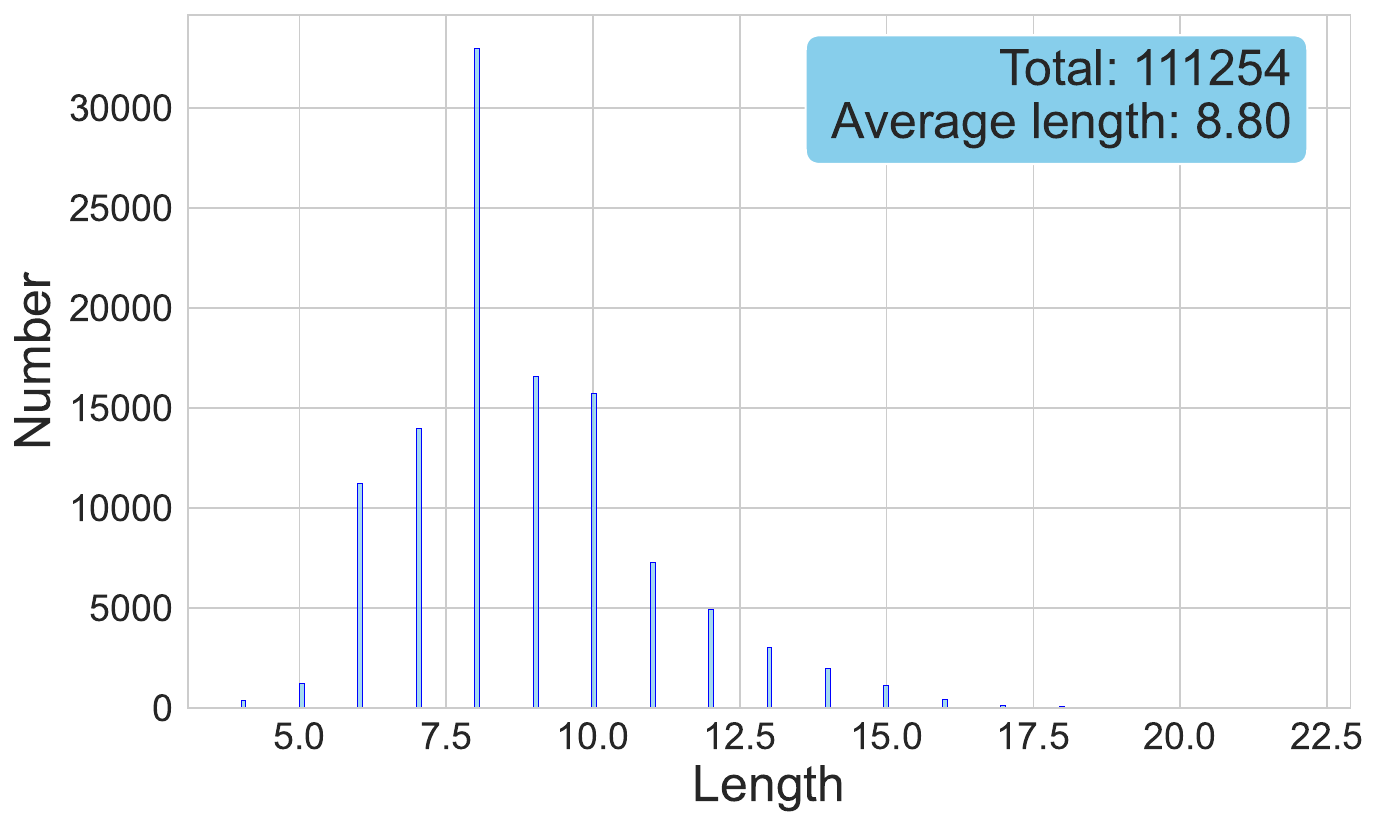} & \includegraphics[width=0.5\linewidth]{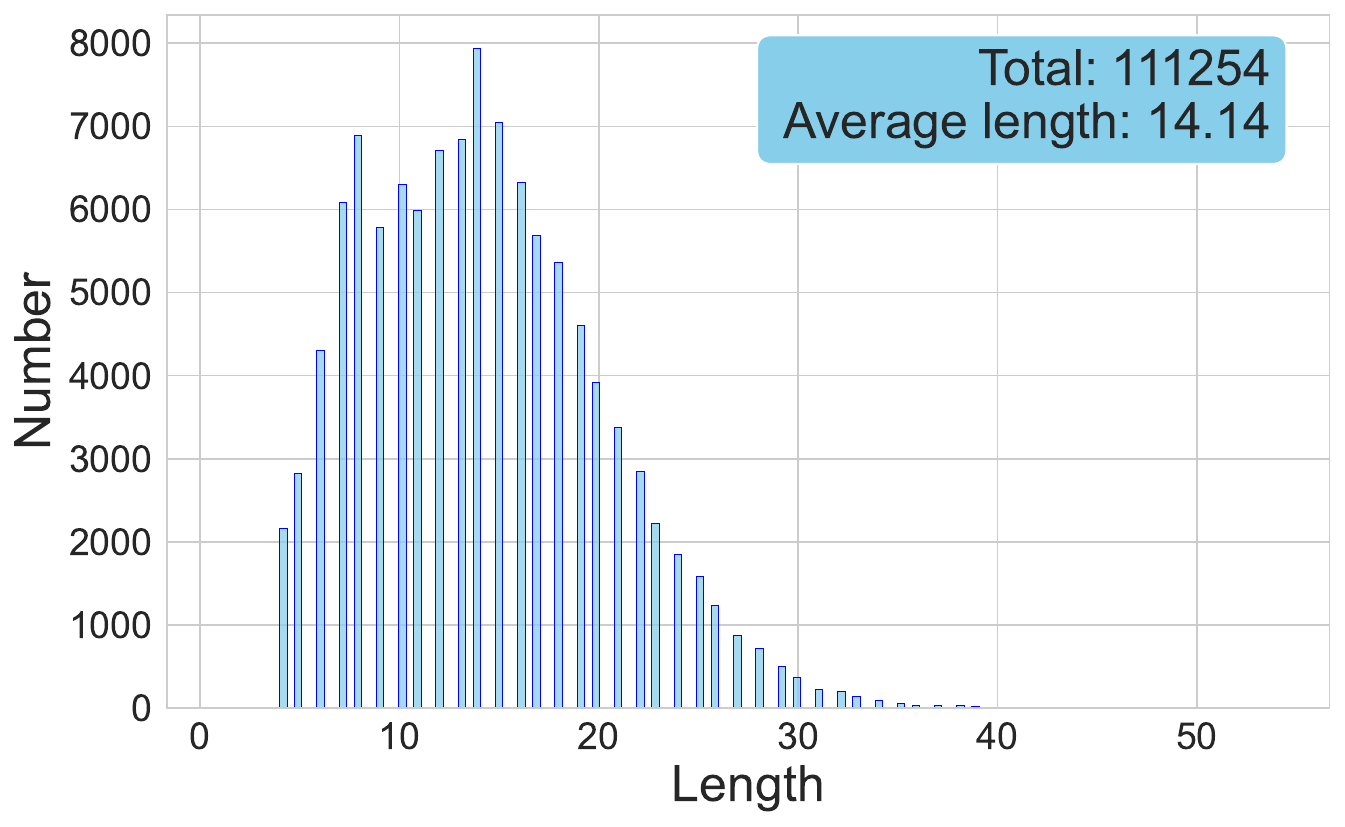} \\
    {\small Instructions} & {\small Responses} \\

    \end{tabular} 
    \end{tabular}
    \label{tab:dataset_stats_qa}
\end{figure}

The average lengths in words of the instructions/responses are $8.51$/$22.76$ for brief descriptions, $7.82$/$77.90$ for detailed descriptions, $8.81$/$14.25$ for single-round QAs and $8.80$/$14.14$ (per round) for multi-round QAs. \cref{tab:dataset_stats_cap} and \cref{tab:dataset_stats_qa} report instruction/response length histograms and the word clouds obtained after removing generic words like “model”, “object” and “NeRF”, emphasizing frequent words in the ground-truth instructions and responses.

\clearpage
\clearpage

\clearpage
\clearpage

\newpage
\subsection{ShapeNeRF--Text quality analysis}
\label{sec:data_quality_analysis}
We have carried out several experiments to assess the quality of the questions in ShapeNeRF--Text. More specifically, the purpose of this analysis was to understand how many questions referred to a detail that is visible only from a specific viewpoint of the object.
First, we evaluated our dataset questions with a language-only model, LLaMA3. For each question $Q$, we asked LLaMA3: \\
\emph{Is a random viewpoint of the object enough to answer this question? \\ 
<$Q$> \\
If so, reply "YES"; if a specific viewpoint is needed, answer "NO".}\\

By doing so, we obtained 5163 “YES” and 5847 “NO”, highlighting that most questions refer to some details which are visible only from a point of view.

\begin{figure}[ht]
    \centering
    \includegraphics[width=\linewidth]{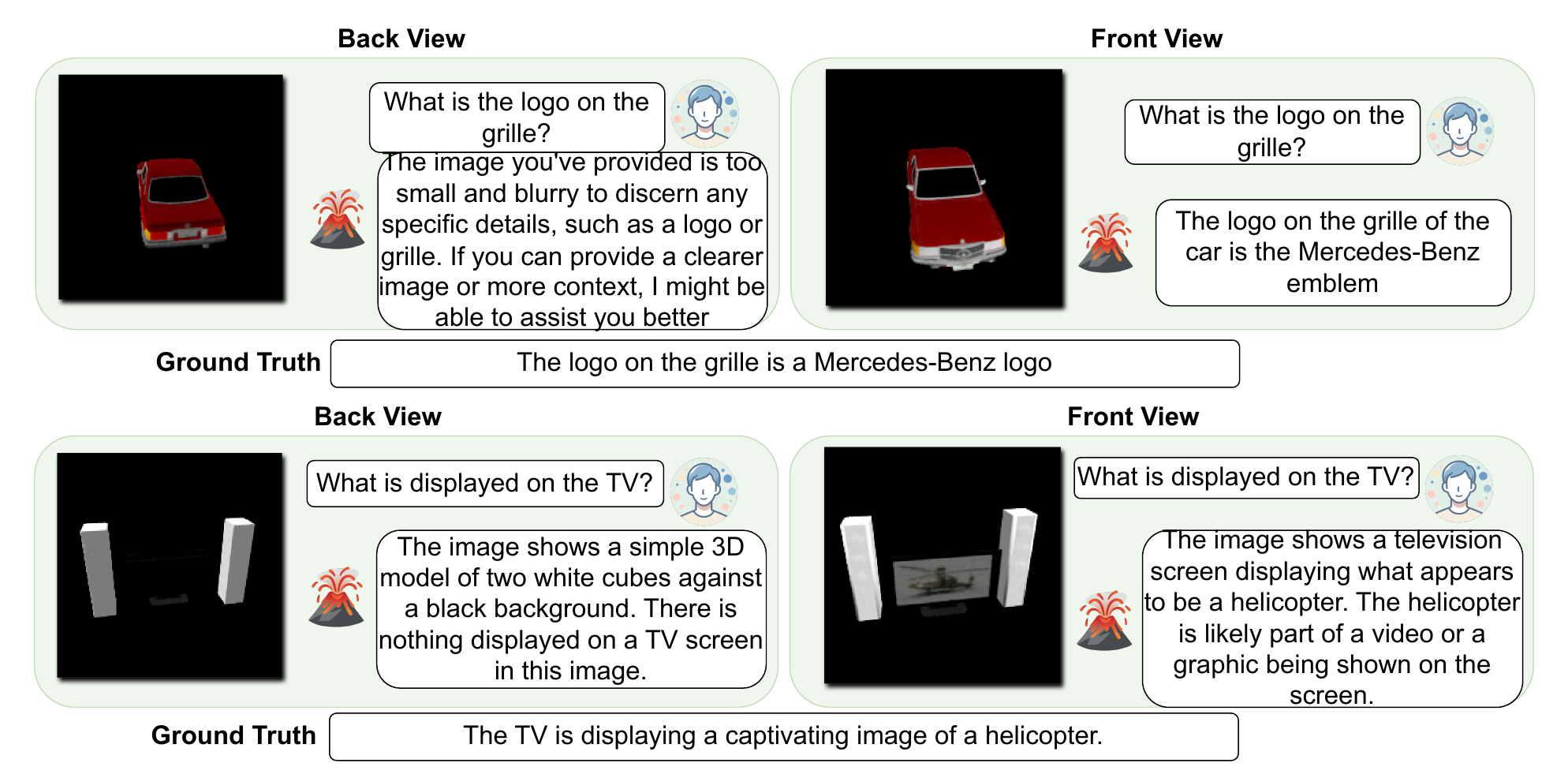}
    \caption{\textbf{Front vs back results with LLaVA.} The dataset contains many view-dependent questions.}
    \label{fig:qual_data_stats}
\end{figure}

Second, we ran a Vision-Language model, LLaVA-1.6-13b, on each question of the single-round Q\&A dataset, on the front and back views of objects. Then, we selected only the LLaVA responses where the answer for the front or back view achieves a SimCSE score higher than 80\%, i.e., likely correct answers, which selects approximately 45\% of the answers. Among these correct responses, we calculated the percentage of those where the front and back answers are extremely different (i.e., a difference in SimCSE scores > 10). Remarkably, 26\% of such answers are correct from one point of view but wrong from the other: these questions would have required multi-view information to be answered correctly. We report two qualitative examples in \cref{fig:qual_data_stats}. In the first row, the Mercedes-Benz logo cannot be recognized from the back view. In the second row, from the rear viewpoint, the monitor seems turned off, and thus it is not possible to identify correctly the helicopter displayed on the screen. Similarly, \cref{fig:single-round_supp} of the Appendix shows other examples of this kind of cases.

\section{Additional baselines results and details}

\subsection{Language-only baseline}
\label{sec:language_baseline_supp}
To assess potential spurious patterns in the question-answer relationships, we evaluate the performance of LLaMA 2, the LLM on which \model{} relies, fine-tuned on ShapeNeRF-Text. In this training and evaluation protocol, the LLM is provided with questions belonging to the dataset and must return the correct answers without having access to the NeRF data. Therefore, the predicted answers may be generated only based on the textual patterns present in the training set. Results are shown in \cref{tab:language_baseline}. A significant performance gap exists between LLaMA 2 and LLaNA, highlighting that our dataset consists of questions that can only be answered with access to information about 3D objects.

\begin{table}[ht]
    \centering
    \caption{\textbf{Language-only baseline.}}
    \resizebox{0.8\linewidth}{!}{
    \begin{tabular}{clcccccc}
    \toprule
    % \multicolumn{7}{c}{\textbf{\nerf{} captioning - brief description}} \\
    % \midrule
    & \textbf{Model} & \textbf{Modality} & \textbf{Sentence-BERT} & \textbf{SimCSE} & \textbf{BLEU-1} & \textbf{ROUGE-L} & \textbf{METEOR} \\
      \cmidrule(lr){2-2} \cmidrule(lr){3-3} \cmidrule(lr){4-8}
    
    \multirow{2}{*}{\rotatebox{90}{\scriptsize Brief}} &
    LLaMA2 & Text-only & 42.62 & 40.70 & \textbf{25.14} & 24.53 & 25.53 \\
      
      \cmidrule(lr){2-2} \cmidrule(lr){3-3} \cmidrule(lr){4-8}
    & \model{}-7b & \nerf{} & \textbf{68.63} & \textbf{70.54} & 20.64 & \textbf{28.33} & \textbf{31.76} \\
    \midrule
    
    \multirow{2}{*}{\rotatebox{90}{\scriptsize Detailed}} & LLaMA2 & Text-only & 49.73 & 47.68 & 15.15 & 23.27 & 14.78 \\
    
      \cmidrule(lr){2-2} \cmidrule(lr){3-3} \cmidrule(lr){4-8}
    & \model{}-7b & \nerf{} & \textbf{77.43} & \textbf{79.81} & \textbf{41.32} & \textbf{36.18} & \textbf{32.39} \\
      
    \midrule
    % \multicolumn{7}{c}{\textbf{\nerf{} Q\&A - single round conversation}} \\
    % \midrule
    % \textbf{Model} & \textbf{Modality} & \textbf{Sentence-BERT} & \textbf{SimCSE} & \textbf{BLEU-1} & \textbf{ROUGE-L} & \textbf{METEOR} \\
    %   \cmidrule(lr){1-1} \cmidrule(lr){2-2} \cmidrule(lr){3-7}
    \multirow{2}{*}{\rotatebox{90}{\scriptsize SingleQA}} & 
    LLaMA2 & Text-only & 68.37 & 68.46 & 44.07 & 51.15 & 48.00 \\
    
      \cmidrule(lr){2-2} \cmidrule(lr){3-3} \cmidrule(lr){4-8}
        
    & \model{}-7b & \nerf{} & \textbf{81.03} & \textbf{81.56} & \textbf{46.16} & \textbf{53.17} & \textbf{50.15} \\
    \bottomrule
    \end{tabular}}
    \label{tab:language_baseline}
\end{table}

%\subsection{Multi-view VLM baseline details}
%\label{sec:multiview_llava}
%\textcolor{red}{We have realized a multi-view baseline for LLaVA by rendering images from $N$ viewpoints randomly sampled between the set of camera poses used to train each NeRF. Then, we concatenated tokens from the N images and fed them into LLaVA alongside text instructions. We set $N$=$3$ because the model cannot process a higher number of images correctly. Results are reported in \cref{tab:brief_frozen}, \cref{tab:brief_frozen_hst}, \cref{tab:detailed_frozen}, \cref{tab:qa_frozen} and \cref{tab:classification_frozen}.}

\subsection{Zero-Shot NeRF classification of trained baselines}
We report in \cref{tab:classification_trained} the results obtained on zero-shot NeRF classification task by the baselines trained on ShapeNeRF-Text. Results follow the same trend as the other language tasks, reported in the main paper.

%%% AGGREGATED FROZEN AND TRAINED TABLES

\begin{table}[h]
\caption{\textbf{Zero-shot NeRF classification on ShapeNeRF-Text. Trained baselines.}\\Best results are in \textbf{bold}, runner-up is \underline{underlined}. (FV: front-view)}
\centering
\resizebox{0.5\linewidth}{!}{%
\begin{tabular}{lcc}
    \midrule
    \textbf{Model} & \textbf{Modality} & \textbf{Accuracy (\%)} \\
    \cmidrule(lr){1-1} \cmidrule(lr){2-2} \cmidrule(lr){3-3}
    \llava{}-vicuna-13b & Image (FV) & 36.49 \\
    \cmidrule(lr){1-1} \cmidrule(lr){2-2} \cmidrule(lr){3-3}
    PointLLM-7b & Point cloud &  \underline{49.69}\\
    GPT4Point-Opt-2.7b & Point cloud & 26.30 \\
    \cmidrule(lr){1-1} \cmidrule(lr){2-2} \cmidrule(lr){3-3}
    \model{}-7b & \nerf{} & \textbf{67.14} \\
    \bottomrule
    \end{tabular}}
    \label{tab:classification_trained}
\end{table}

\section{Ground-truth images and point clouds}
\label{sec:gt_mesh_supp}
This section presents the results of an experiment in which the baseline 2D and 3D MLLMs have been provided with ground-truth input images and point clouds extracted from the original 3D meshes in the dataset rather than from the NeRFs.
%, as shown previously. 
%
This scenario estimates an upper bound for the performance of such approaches when used as NeRF assistants, by simulating perfect extraction of images or point clouds from the NeRFs. In other words, it simulates the ideal scenario in which the encoding of information inside a NeRF is lossless, a non-realisitc situation in which the baselines can achieve their best performance. 
\cref{tab:brief_performance_comparison_mesh}, \cref{tab:det_performance_comparison_mesh}, and \cref{tab:single_round_performance_comparison_mesh} show the results of this experiments on the tasks of brief description, detailed description, and single-round Q\&A, respectively.
For brevity, the best-performing 2D model, i.e., \llava{}~\cite{llava} (on front views) and the best-performing 3D model, i.e., PointLLM~\cite{pointllm}, have been tested in this scenario.
The results demonstrate that, even in this idealized and most favorable scenario for the baselines, \model{} outperforms them.

\begin{table}[ht]
\centering
\resizebox{\linewidth}{!}{
\begin{tabular}{lcccccc}
\toprule
\textbf{Model} & \textbf{Modality} & \textbf{Sentence-BERT} & \textbf{SimCSE} & \textbf{BLEU-1} & \textbf{ROUGE-L} & \textbf{METEOR} \\
\cmidrule(lr){1-1} \cmidrule(lr){2-2} \cmidrule(lr){3-7}
\llava{}-vicuna-13b& Image (FV) & 68.61 & 67.99 & 17.48 & 23.08 & 27.03 \\
\cmidrule(lr){1-1} \cmidrule(lr){2-2} \cmidrule(lr){3-7}
PointLLM-7b & Point cloud                      & 51.99 & 51.70 & 17.19 & 18.63 &  15.03 \\
\cmidrule(lr){1-1} \cmidrule(lr){2-2} \cmidrule(lr){3-7}
\model{}-7b & \nerf{}                      & \textbf{68.63} & \textbf{70.54} &  \textbf{20.64} & \textbf{28.33} & \textbf{31.76} \\
\bottomrule
\end{tabular}}
\caption{\textbf{\nerf{} brief captioning on ShapeNeRF--Text dataset.} Frozen baseline results obtained on data extracted from ShapeNet mesh data. Best results in \textbf{bold}. Runner-up \underline{underlined}. \\(FV: front-view)}
\label{tab:brief_performance_comparison_mesh}
\end{table}

\begin{table}[ht]
\centering
\resizebox{\linewidth}{!}{
\begin{tabular}{lcccccc}
\toprule
\textbf{Model} & \textbf{Modality} & \textbf{Sentence-BERT} & \textbf{SimCSE} & \textbf{BLEU-1} & \textbf{ROUGE-L} & \textbf{METEOR} \\
\cmidrule(lr){1-1} \cmidrule(lr){2-2} \cmidrule(lr){3-7}
\llava{}-vicuna-13b& Image (FV) & 68.32 & 67.35 &  27.46 & 26.62 & 24.40 \\
\cmidrule(lr){1-1} \cmidrule(lr){2-2} \cmidrule(lr){3-7}
PointLLM-7b & Point cloud                      & 61.87 & 61.77 &  10.65 & 19.90 &  10.93 \\
\cmidrule(lr){1-1} \cmidrule(lr){2-2} \cmidrule(lr){3-7}
\model{}-7b & \nerf{}                      & \textbf{77.43} & \textbf{79.81} &  \textbf{41.32} & \textbf{36.18} & \textbf{32.39} \\
\bottomrule
\end{tabular}}
\caption{\textbf{\nerf{} detailed captioning on ShapeNeRF--Text dataset.} Frozen baseline results obtained on data extracted from ShapeNet mesh data. Best results in \textbf{bold}. Runner-up \underline{underlined}.}
\label{tab:det_performance_comparison_mesh}
\end{table}

\begin{table}[ht]
\centering
\resizebox{\linewidth}{!}{
\begin{tabular}{lcccccc}
\toprule
\textbf{Model} & \textbf{Modality} & \textbf{Sentence-BERT} & \textbf{SimCSE} & \textbf{BLEU-1} & \textbf{ROUGE-L} & \textbf{METEOR} \\
\cmidrule(lr){1-1} \cmidrule(lr){2-2} \cmidrule(lr){3-7}
\llava{}-vicuna-13b& Image (FV) & 78.40 & 75.68 & 22.65 & 33.04 & 35.70 \\
\cmidrule(lr){1-1} \cmidrule(lr){2-2} \cmidrule(lr){3-7}
PointLLM-7b & Point cloud                      & 74.98 & 74.90 &  36.93 & 44.60 &  39.87 \\
\cmidrule(lr){1-1} \cmidrule(lr){2-2} \cmidrule(lr){3-7}
\model{}-7b & \nerf{}                      & \textbf{81.03} & \textbf{81.56} & \textbf{46.16} & \textbf{53.17} & \textbf{50.15} \\
\bottomrule
\end{tabular}}
\caption{\textbf{\nerf{} single-round Q\&A on ShapeNeRF--Text dataset.} Frozen baseline results obtained on data extracted from ShapeNet mesh data. Best results in \textbf{bold}. Runner-up \underline{underlined}.}
\label{tab:single_round_performance_comparison_mesh}
\end{table}

\section{Generalization experiments}
\label{sec:generalize_supp}
We conducted an experiment to probe the generalization capabilities of \model{} against the trained baselines. We evaluate the models on the subset of 200 Objaverse~\cite{deitke2023objaverse} objects with human-annotated captions used as a test set by PointLLM~\cite{pointllm}. This evaluation protocol sets forth a challenging out-of-domain and open-set experiment (164 out of 200 Objaverse objects belong to categories not present in ShapeNeRF-Text). To test \model{}, we fit NeRFs for all the objects of the test set. Then, we extracted colored point clouds and rendered front views from NeRFs, in order to test the baselines. In~\cref{tab:generalization_objaverse} we can observe that the scores of all models are significantly lower compared to \cref{tab:brief_trained}, which hints at all models struggling when evaluated on objects very different from those belonging to the training domain. \model{} achieves the second-best generalization performance after PointLLM. Yet, it is worth highlighting that the frozen modality-specific encoder of PointLLM (and GPT4Point) is PointBERT, which was pre-trained on Objaverse. In contrast, LLaNA meta-encoder, nf2vec, has been trained only on ShapeNet, meaning it has never encountered objects outside the ShapeNet categories.

\begin{table}[ht]
\centering
\caption{\textbf{Generalization results on Objaverse.}}
\resizebox{\linewidth}{!}{
    \begin{tabular}{c}

        \begin{tabular}{lcccccc}
        \toprule
        \multicolumn{7}{c}{\textbf{\nerf{} captioning}} \\
        \midrule
        \textbf{Model} & \textbf{Modality} & \textbf{Sentence-BERT} & \textbf{SimCSE} & \textbf{BLEU-1} & \textbf{ROUGE-L} & \textbf{METEOR} \\
          \cmidrule(lr){1-1} \cmidrule(lr){2-2} \cmidrule(lr){3-7}
        
        \llava{}-vicuna-13b & Image (FV) & 27.07 & 26.82 & 4.41 & 6.81 & 9.77\\
          \cmidrule(lr){1-1} \cmidrule(lr){2-2} \cmidrule(lr){3-7}
        
        PointLLM-7b & Point cloud &  \textbf{33.88} & \textbf{33.04} & \textbf{5.37} & \textbf{8.14} & \textbf{12.28} \\
        
        GPT4Point-Opt-2.7b & Point cloud & 25.94 & 29.04& 4.25& 7.99& 10.42\\
        
          \cmidrule(lr){1-1} \cmidrule(lr){2-2} \cmidrule(lr){3-7}
        \model{}-7b & \nerf{} & 30.07 & 28.17 & 4.93 & 7.47 & 11.15  \\
        \bottomrule
        \end{tabular}
    \end{tabular}}
\label{tab:generalization_objaverse}
\end{table}

\section{Additional qualitative examples}
\label{sec:qualitatives_supp}
\begin{figure}
    \centering
    \includegraphics[width=\linewidth]{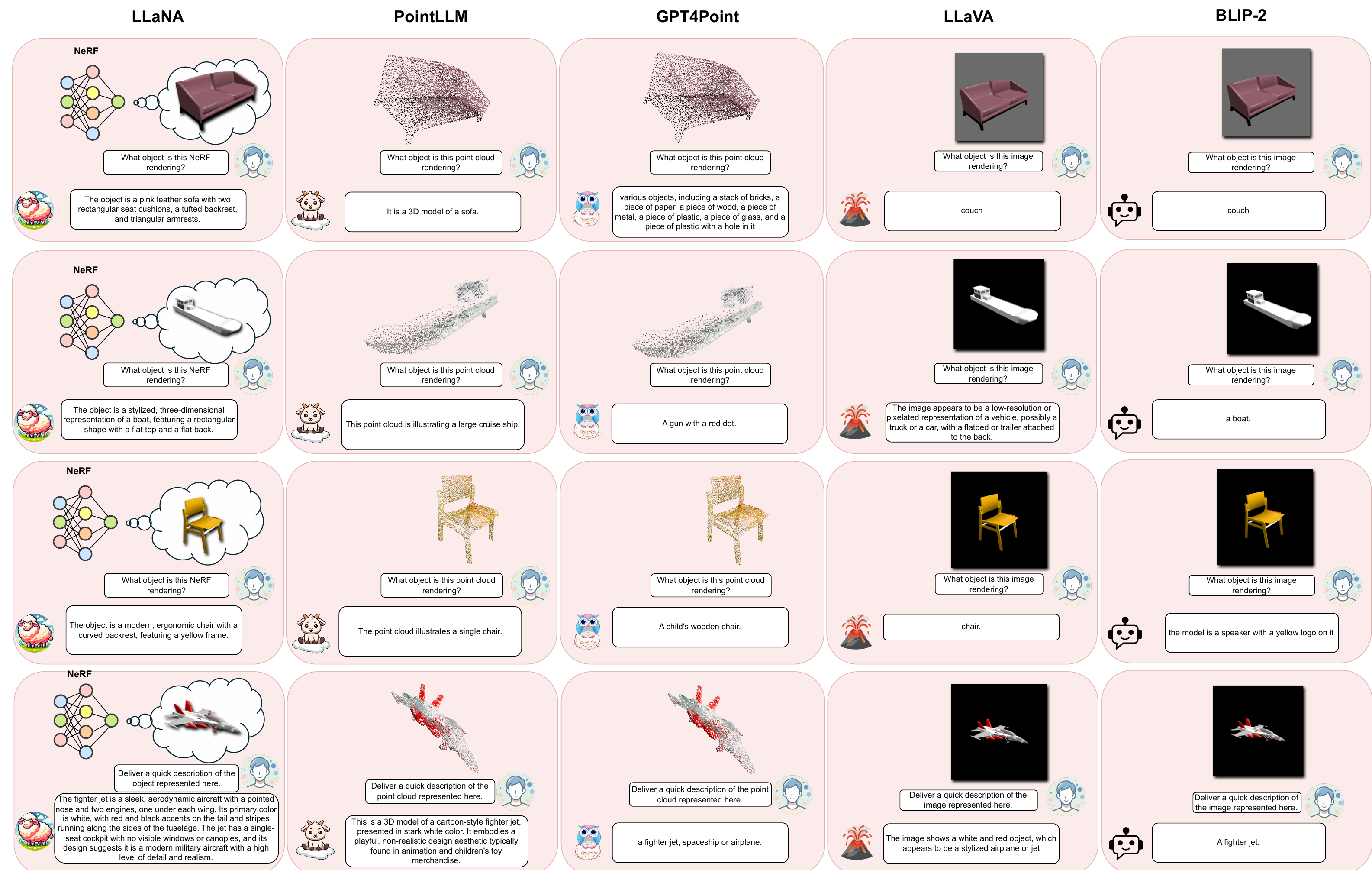}
    \caption{\textbf{Additional qualitative examples for the brief description task.}}
    \label{fig:brief_supp}
\end{figure}

\begin{figure}
    \centering
    \includegraphics[width=\linewidth]{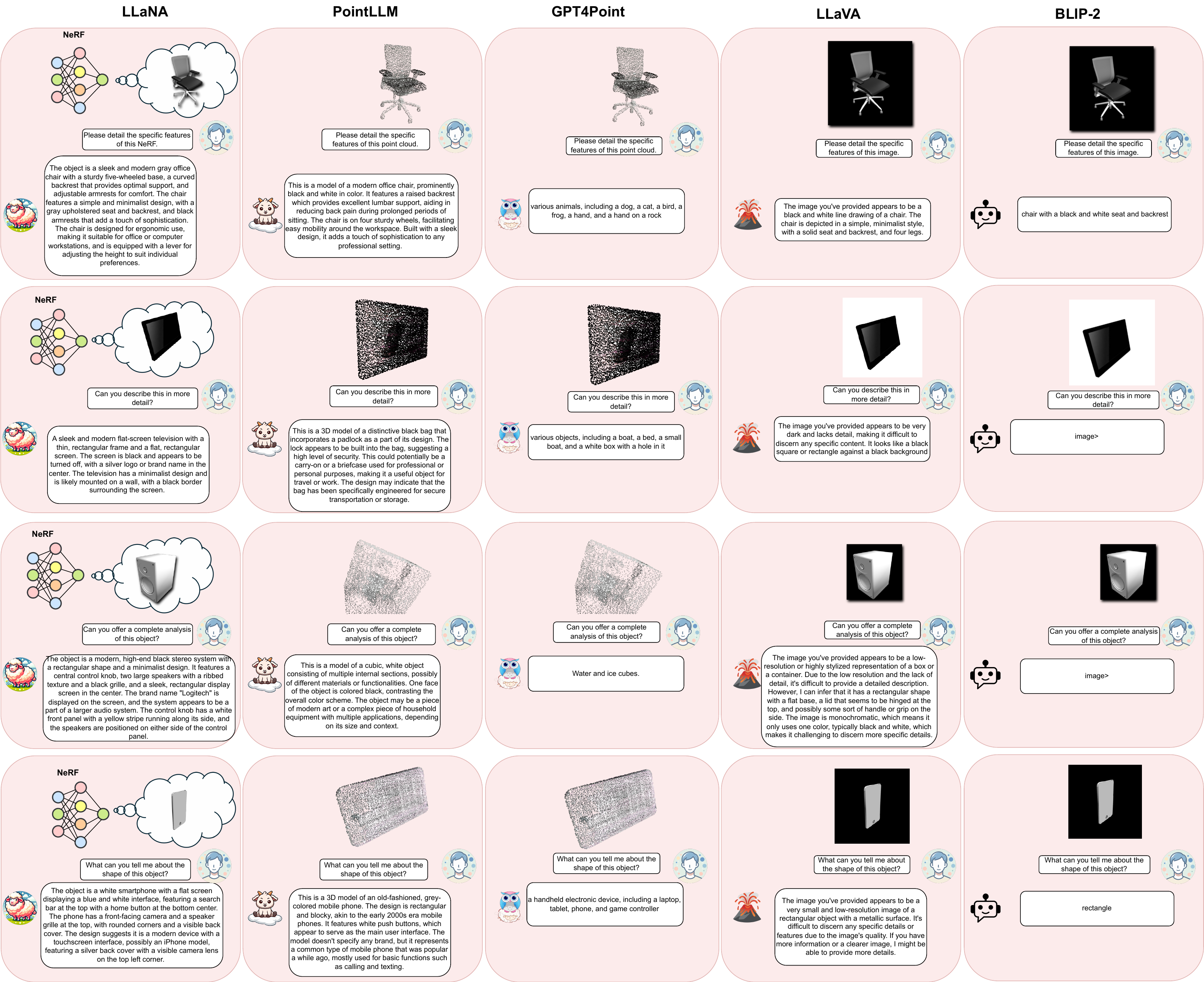}
    \caption{\textbf{Additional qualitative examples for the detailed description task.}}
    \label{fig:detailed_supp}
\end{figure}

\begin{figure}
    \centering
    \includegraphics[width=\linewidth]{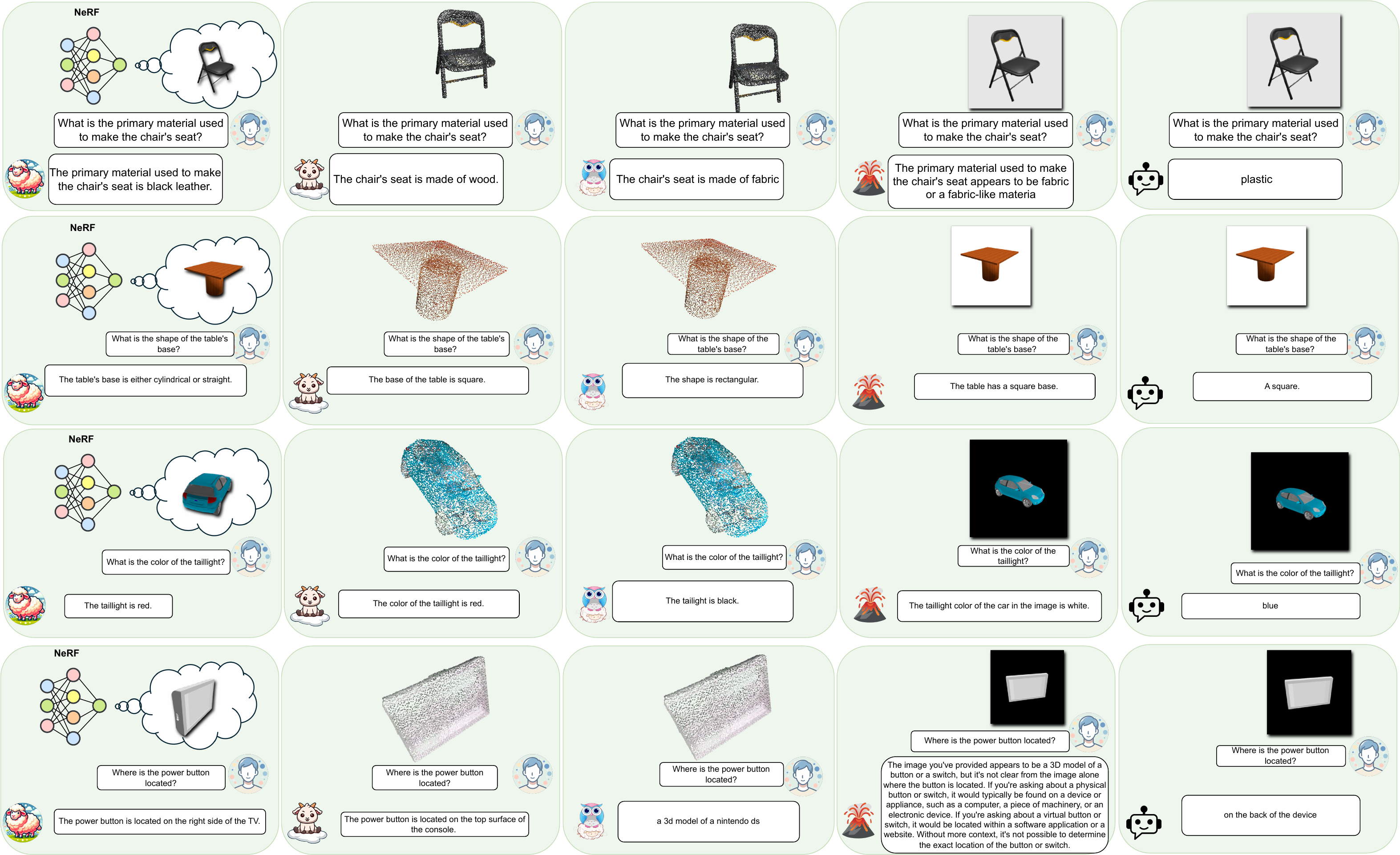}
    \caption{\textbf{Additional qualitative examples for the single-round Q\&A task.}}
    \label{fig:single-round_supp}
\end{figure}

This section provides additional qualitative comparisons between the proposed method, i.e. \model{} which directly processes \nerf{}, and the baselines that take as input images~\cite{llava,blip2} or 3D representations~\cite{pointllm,gpt4point}. 
In particular, \cref{fig:brief_supp,fig:detailed_supp,fig:single-round_supp} show additional brief descriptions, detailed descriptions, and single-round Q\&A provided as output by the different methods.
Many examples, such as the white speaker in the third row of \cref{fig:detailed_supp}, are not described properly by MLLMs operating on point clouds. Indeed, due to the input point cloud containing only $8192$ points, these methods cannot perceive the object details, such as the curved surface of the speaker, therefore they predict that the object is a ``cubic white object'' or an ``ice cube''. 
In other examples, such as the white screen sample in the last row of \cref{fig:single-round_supp}, the LLM operating on images cannot give the right answer to the question on the button location as it is not visible from the given viewpoint. 
Contrarily, by operating directly on the holistic representation provided by \nerf{}s, \model{} provides the right answer in these situations.


% \section{Failure cases}
% Failure Cases
%TODO: aggiungi failure cases

\section{Information about datasets, models, and source code licenses}
\label{sec:licenses}

This section provides details about the datasets, models, and source code licenses used in the paper, ensuring proper credit to the creators or original owners, and adherence to license terms.

\textbf{Datasets:}~the datasets employed in our work and the relative licenses are listed below:
\begin{itemize}
    \item \textbf{ShapeNet}: licensed under GNU Affero General Public License v3.0.
    \item \textbf{GPT2Shape HST}: licensed under Creative Commons Attribution-NonCommercial-ShareAlike 4.0 International License.
\end{itemize}

\textbf{Models:}~the models used in all our experiments and their relative licenses are detailed in the following:
\begin{itemize}
    \item \textbf{nf2vec}: licensed under MIT License.
    \item \textbf{PointLLM}: licensed under Creative Commons Attribution-NonCommercial-ShareAlike 4.0 International License.
    \item \textbf{GPT4Point}: licensed under Creative Commons Attribution-NonCommercial-ShareAlike 4.0 International License.
    \item \textbf{LLAMA-2}: licensed under META LLAMA 2 COMMUNITY LICENSE AGREEMENT\footnote{\url{https://ai.meta.com/llama/license/}}.
    \item \textbf{LLAMA-3}: licensed under META LLAMA 3 COMMUNITY LICENSE AGREEMENT\footnote{\url{https://ai.meta.com/llama/license/}}.
    \item \textbf{LLAVA}: licensed under Apache License 2.0.
\end{itemize}

Proper care has been taken to ensure that all licenses and terms of use are explicitly mentioned and respected throughout this paper.

%%%%%%%%%%%%%%%%%%%%%%%%%%%%%%%%%%%%%%%%%%%%%%%%%%%%%%%%%%%%
%\input{sections_camera_ready/checklist}
%%%%%%%%%%%%%%%%%%%%%%%%%%%%%%%%%%%%%%%%%%%%%%%%%%%%%%%%%%%%

\end{document}